\documentclass{article}

\usepackage{microtype}
\usepackage{graphicx}
\usepackage{subfigure}
\usepackage{booktabs} 

\usepackage{hyperref}

\usepackage[accepted]{icml2025}

\usepackage[textsize=tiny]{todonotes}
\definecolor{babypink}{rgb}{0.96, 0.76, 0.76} 
\definecolor{burntsienna}{rgb}{0.91, 0.45, 0.32}     
\definecolor{crimson}{rgb}{0.86, 0.08, 0.24}
\definecolor{darkspringgreen}{rgb}{0.09, 0.45, 0.27}
\definecolor{deepcarrotorange}{rgb}{0.91, 0.41, 0.17}
\usepackage[utf8]{inputenc} 
\usepackage[T1]{fontenc}    
\usepackage{hyperref}       
\usepackage{nccmath}
\hypersetup{
     colorlinks=true,
     linkcolor=blue,
     filecolor=blue,
     citecolor=blue,
     urlcolor=blue,
    }
\usepackage{url}
\usepackage{wrapfig}
\usepackage{colortbl}
\usepackage{booktabs}       
\usepackage{amsfonts}       
\usepackage{nicefrac}       
\usepackage{microtype}      
\usepackage{xcolor}
\usepackage{multicol}
\usepackage{graphicx}
\usepackage{transparent}
\usepackage{amsthm}
\usepackage{bbm}
\usepackage{comment}
\usepackage{enumitem}
\usepackage{bm}
\usepackage{subfigure}
\usepackage{mathtools}
\usepackage[export]{adjustbox}
\usepackage{stmaryrd}
\usepackage{xifthen}
\usepackage{xargs}
\usepackage{amsmath,amssymb}
\usepackage{algorithm,algorithmic}
\usepackage{hyperref}
\usepackage{caption}
\usepackage{titlesec}
\usepackage{multirow}
\usepackage[capitalize,noabbrev]{cleveref}

\def\rset{\mathbb{R}}
\def\nset{\mathbb{N}}
\def\rmd{\mathrm{d}}
\def\rme{\mathrm{e}}
\def\normpdf{\mathrm{N}}

\def\eqsp{\,}
\def\Id{\mathbf{I}}
\def\zero{0}

\def\pE{\mathbb{E}}

\def\std{\sigma}
\def\a{\alpha}

\def\eqdef{\vcentcolon=}

\def\wrt{w.r.t.}
\def\diag{\mbox{diag}}

\def\smbs{{\scriptscriptstyle{\backslash}}}

\def\txts{\textstyle}
\def\rhs{r.h.s.}

\newcommandx\probmeas[4][4=]{
    \ifthenelse{\equal{#4}{}}{
    \ifthenelse{\equal{#2}{}}{
        \ifthenelse{\equal{#3}{}}{
            \mu _{#1}
        }{
            \mu _{#1}(#3)
        }
    }{
        \ifthenelse{\equal{#3}{}}{
            \mu _{#1}(\cdot|#2)
        }{
            \mu _{#1}(#3|#2)
        }
    }}{
    \ifthenelse{\equal{#2}{}}{
        \ifthenelse{\equal{#3}{}}{
            \mu^{#4} _{#1}
        }{
            \mu^{#4} _{#1}(#3)
        }
    }{
        \ifthenelse{\equal{#3}{}}{
            \mu^{#4} _{#1}(\cdot|#2)
        }{
            \mu^{#4} _{#1}(#3|#2)
        }
    }
    }
}

\newcommandx\post[4][4=\obs]{
    \ifthenelse{\equal{#2}{}}{
        \ifthenelse{\equal{#3}{}}{
            \pi^{#4} _{#1}
        }{
            \pi^{#4} _{#1}(#3)
        }
    }{
        \ifthenelse{\equal{#3}{}}{
            \pi^{#4} _{#1}(\cdot|#2)
        }{
            \pi^{#4} _{#1}(#3|#2)
        }
    }
    }

\newcommandx\vi[3][3=\vparam]{
    \ifthenelse{\equal{#2}{}}{
            \lambda^{#3} _{#1}
        }{
            \lambda^{#3} _{#1}(#2)
        }
    }

\newcommandx\hpost[4][4=\obs]{
    \ifthenelse{\equal{#2}{}}{
        \ifthenelse{\equal{#3}{}}{
            \hat\pi^{#4} _{#1}
        }{
            \hat\pi^{#4} _{#1}(#3)
        }
    }{
        \ifthenelse{\equal{#3}{}}{
            \hat\pi^{#4} _{#1}(\cdot|#2)
        }{
            \hat\pi^{#4} _{#1}(#3|#2)
        }
    }}

\newcommand\epost[3]{
    \ifthenelse{\equal{#2}{}}{
        \ifthenelse{\equal{#3}{}}{
            \overline\pi^{\obs}_{#1}
        }{
            \overline\pi^{\obs}_{#1}(#3) 
        }
    }{
        \ifthenelse{\equal{#3}{}}{
            \overline\pi^{\obs}_{#1}(\cdot|#2)
        }{
            \overline\pi^{\obs}_{#1}(#3|#2) 
        }
    }}

\newcommand\pot[2]{
    \ifthenelse{\equal{#2}{}}{
        g_{#1}(\obs|\cdot)
    }{
        g_{#1}(\obs|#2)
    }}
\newcommandx\hpot[3][3=]{
    \ifthenelse{\equal{#2}{}}{
        \hat{g}^{#3} _{#1}(\obs|\cdot)
    }{
        \hat{g}^{#3} _{#1}(\obs|#2)
    }}

\newcommandx\pdata[4][4=]{
    \ifthenelse{\equal{#4}{}}{
    \ifthenelse{\equal{#2}{}}{
        \ifthenelse{\equal{#3}{}}{
            p _{#1}
        }{
            p _{#1}(#3)
        }
    }{
        \ifthenelse{\equal{#3}{}}{
            p _{#1}(\cdot|#2)
        }{
            p _{#1}(#3|#2)
        }
    }}{
    \ifthenelse{\equal{#2}{}}{
        \ifthenelse{\equal{#3}{}}{
            p^{#4} _{#1}
        }{
            p^{#4} _{#1}(#3)
        }
    }{
        \ifthenelse{\equal{#3}{}}{
            p^{#4} _{#1}(\cdot|#2)
        }{
            p^{#4} _{#1}(#3|#2)
        }
    }
    }
}
\newcommand\fw[3]{
        \ifthenelse{\equal{#3}{}}{
            q _{#1}(\cdot|#2)
        }{
            q _{#1}(#3|#2)
        }
    }
\newcommandx\denoiser[4][4=]{
    \ifthenelse{\equal{#2}{}}{
        \ifthenelse{\equal{#3}{}}{
            D^{#4}_{#1}
        }{
            D^{#4} _{#1}(#3)
        }
    }{
        \ifthenelse{\equal{#3}{}}{
            D^{#4} _{#1}(\cdot|#2)
        }{
            D^{#4} _{#1}(#3|#2)
        }
    }}
\def\wght{\omega}
\newcommand{\intset}[2]{\llbracket #1, #2 \rrbracket}
\newcommand{\kldivergence}[2]{\mathsf{KL}(#1 \parallel #2)}

\def\algoname{{\sc{Mixture-Guided Diffusion Model}}}
\def\ddimfn{\small{\mathsf{DDPM}}}
\def\vifn{\small{\mathsf{Gauss}\_\mathsf{VI}}}

\def\algo{{\sc{MGDM}}}
\def\pgdm{{\sc{PGDM}}}
\def\dps{{\sc{DPS}}}
\def\diffpir{{\sc{DiffPir}}}

\def\ddnm{{\sc{DDNM}}}
\def\resample{{\sc{ReSample}}}
\def\psld{{\sc{PSLD}}}

\def\reddiff{{\sc{RedDiff}}}

\def\daps{{\sc{DAPS}}}
\def\pnpdm{{\sc{PNP-DM}}}
\def\isdm{{\sc{ISDM}}}
\def\msdm{{\sc{MSDM}}}
\def\demucs{{\textsc{{Demucs}\textsubscript{512}}}}

\def\ffhq{{\texttt{FFHQ}}}
\def\imagenet{{\texttt{ImageNet}}}
\def\slakh{{\texttt{slakh2100}}}
\def\sisdri{{SI-SDR\textsubscript{I}}}

\def\gauss{\mathcal{N}}
\def\bfA{\mathbf{A}}
\def\stdobs{\sigma_\obs}
\def\obs{{\mathbf{y}}}
\def\postet{\pi}
\def\param{\theta}

\def\dimobs{{d_\obs}}
\def\dimx{{d}}

\newcommandx{\acp}[2][2=]{\ifthenelse{\equal{#2}{}}{\alpha_{#1}}{\alpha_{#1:#2}}}
\newcommandx{\hpredx}[3][2=0,3=\param]{\smash{\boldsymbol{m}^{#3} _{#2|#1}}}
\newcommandx{\predx}[2][2=0]{\smash{\boldsymbol{m} _{#2|#1}}}
\newcommandx{\prednoise}[2][2=\param]{\smash{\boldsymbol{\epsilon}^{#2} _{#1}}}
\newcommandx{\fwmarg}[3][3=]{\ifthenelse{\equal{#2}{}}{q^{#3} _{#1}}{q^{#3} _{#1}(#2)}}
\newcommandx{\bw}[4][4=]{\ifthenelse{\equal{#3}{}}{q^{#4} _{#1}}{q^{#4} _{#1}(#3|#2)}}
\newcommandx{\bwp}[4][4=\param]{\ifthenelse{\equal{#3}{}}{p^{#4} _{#1}}{p^{#4} _{#1}(#3|#2)}}
\newcommand{\hpotn}[2]{\ifthenelse{\equal{#2}{}}{\hat{g}^\param _{#1}}{\hat{g}^\param _{#1}(#2)}}
\newcommand{\potn}[2]{\ifthenelse{\equal{#2}{}}{g_{#1}}{g_{#1}(#2)}}
\newcommandx{\pibw}[4][4=]{\ifthenelse{\equal{#3}{}}{\smash{\postet}^{#4} _{#1}}{\smash{\postet}^{#4} _{#1}(#3|#2)}}
\newcommandx{\fwtrans}[4][4=]{\ifthenelse{\equal{#2}{}}{q_{#1}}{q _{#1}(#3|#2)}}
\newcommandx{\mgibbs}[4][4=]{\ifthenelse{\equal{#2}{}}{\tilde{\pi}_{#1}}{\tilde{\pi} _{#1}(#3|#2)}}

\def\vparam{{\bm\varphi}}
\def\vmu{{\bm\mu}}
\def\vlstd{{\bm\rho}}

\def\bx{\mathbf{x}}
\def\bX{X}

\def\bXy{\bar{X}}
\def\bZy{\bar{Z}}
\def\vX{\hat{X}}
\def\bY{Y}
\def\bz{\mathbf{z}}

\newcommand{\first}[1]{\colorbox{deepcarrotorange!80}{#1}}
\newcommand{\second}[1]{\colorbox{deepcarrotorange!40}{#1}}
\newcommand{\third}[1]{\colorbox{deepcarrotorange!15}{#1}}

\def\gibbsReps{{R}}

\theoremstyle{plain}
\newtheorem{theorem}{Theorem}[section]

\theoremstyle{definition}

\theoremstyle{remark}
\newtheorem{remark}[theorem]{Remark}

\begin{document}

\twocolumn[
\icmltitle{A Mixture-Based Framework for Guiding Diffusion Models}

\icmlsetsymbol{equal}{*}

\begin{icmlauthorlist}
    \icmlauthor{Yazid Janati}{equal,yyy}
    \icmlauthor{Badr Moufad}{equal,yyy}
    \icmlauthor{Mehdi Abou El Qassime}{yyy}\\
    \icmlauthor{Alain Durmus}{yyy}
    \icmlauthor{Eric Moulines}{yyy}
    \icmlauthor{Jimmy Olsson}{sch}
\end{icmlauthorlist}

\icmlaffiliation{yyy}{Ecole polytechnique}
\icmlaffiliation{sch}{KTH University}

\icmlcorrespondingauthor{Yazid Janati, Badr Moufad}{first.last@polytechnique.edu}

\icmlkeywords{Machine Learning, ICML}

\vskip 0.3in
]



\printAffiliationsAndNotice{\icmlEqualContribution} 

\begin{abstract}
  Denoising diffusion models have driven significant progress in the field of Bayesian inverse problems. Recent approaches use pre-trained diffusion models as priors to solve a wide range of such problems, only leveraging inference-time compute and thereby eliminating the need to retrain task-specific models on the same dataset. To approximate the posterior of a Bayesian inverse problem, a diffusion model samples from a sequence of intermediate posterior distributions, each with an intractable likelihood function. This work proposes a novel mixture approximation of these intermediate distributions. Since direct gradient-based sampling of these mixtures is infeasible due to intractable terms, we propose a practical method based on Gibbs sampling. We validate our approach through extensive experiments on image inverse problems, utilizing both pixel- and latent-space diffusion priors, as well as on source separation with an audio diffusion model. The code is available at \url{https://www.github.com/badr-moufad/mgdm}.
\end{abstract}

\section{Introduction}
Inverse problems occur when a signal $X$ of interest must be inferred from an incomplete and noisy observation $Y$, a challenge frequently encountered in diverse fields such as weather forecasting, image reconstruction (\emph{e.g.}, tomography or black-hole imaging), and speech processing.  Such problems are typically ill-posed, 
making it essential to incorporate additional constraints, regularization techniques, or prior knowledge to arrive at meaningful and realistic solutions. 

The Bayesian framework, in conjunction with generative modeling, offers a systematic approach to the challenges associated with inverse problems. Prior knowledge about the signal of interest, often represented through samples from its underlying distribution $\pdata{0}{}{}$, can be leveraged to train a generative model $\pdata{0}{}{}[\param]$ that acts as a prior. 
By combining it with the conditional density $\pot{0}{\bx}$ of the observation given the signal, deduced from the form of the inverse problem at hand, we can compute the posterior distribution. 
Samples drawn from this posterior encapsulate plausible solutions that harmonize prior knowledge with the observed data.
One straightforward approach to approximate sampling from the posterior distribution involves constructing a paired dataset of i.i.d. signals and observations, $(\bX_i, Y_i)_{i = 1}^N$, where $\bX_i \sim \pdata{0}{}{}$ and $Y_i \sim g_0(\cdot | \bX_i)$, and learning a direct mapping \cite{dong2015image} or generative model \cite{ledig2017photo,isola2017image}. The latter, when queried with multiple independent noise samples alongside an observation, 
generates a diverse set of potential reconstructions. However, this approach is inherently \emph{task-specific}, delivering reliable reconstructions only when the conditional distribution of the observation remains unchanged at test time. As a result, it cannot straightforwardly adapt to unseen tasks with the same prior. Adaptation to a new task can only be achieved by retraining a new generative model. 

An increasingly popular approach consists in learning a generative model only for the prior $\pdata{0}{}{}$, and then leveraging inference-time compute to solve any inverse problem for which the likelihood function $\bx \mapsto \pot{0}{\bx}$ is provided in a closed form. This strategy eliminates the need for expensive and inefficient task-specific training. Initially explored with generative models such as variational autoencoders and generative adversarial networks \cite{xia2022gan}, this framework has recently been extended to denoising diffusion models (DDMs) \cite{song2021score,kadkhodaie2020solving,kawar2021snips,kawar2022denoising,chung2023diffusion,song2022pseudoinverse,daras2024survey}, which are the focus of the present paper.

DDMs \cite{sohl2015deep,song2019generative,ho2020denoising} achieve state-of-the-art generative performance across a wide range of domains. At their core is a forward noising process that transforms the data distribution $\pdata{0}{}{}$ into a Gaussian distribution. A generative model is then learned by  reversing this noising process. With a specific parameterization of the backward process, which converts noise into data samples, training the generative model reduces to approximating denoisers for each noise level introduced during the forward process. Recent methods for training-free posterior sampling aim to approximate the denoisers for the posterior distribution, enabling the use of diffusion models for sampling \cite{ho2022video,chung2023diffusion,song2022pseudoinverse}. A posterior distribution denoiser can be decomposed into two terms: the prior denoiser at the same noise level (provided by a pre-trained diffusion model) and the gradient of the log-likelihood of the observation conditioned on the current noisy sample. The latter term, which is intractable, is what guides the samples during the denoising process towards the posterior distribution. Various approximations for this gradient term have been proposed. However, they are often crude and require significant adjustments and heuristics to ensure stability and satisfactory performance. When applied to latent diffusion models, they often demand additional, model-specific adjustments 
\cite{rout2024solving}.\\

\textbf{Our contribution.}\, In this paper, we present a principled method that circumvents these issues by introducing a new approximation of the likelihood term, paired with a sampling scheme based on Gibbs sampling \cite{geman1984stochastic}. 
Our key observation is that multiple approximations can be derived for each likelihood term at a fixed noise level using a simple identity that it satisfies.
However, the scores of these new likelihood approximations are not available in closed form, preventing us from deriving a direct posterior denoiser approximation by combining, through a mixture, the different likelihood approximations. We overcome this limitation by constructing a mixture approximation of the intermediate posterior distributions defined by the diffusion model for the original posterior. 
Our algorithm,  \algoname\ (\algo), proceeds by sequentially sampling from these mixtures using Gibbs sampling. This is enabled by a carefully designed data augmentation scheme that ensures straightforward Gibbs updates. A key advantage of our approach is its adaptability to available computational resources. Specifically, the number of Gibbs iterations acts as a tunable parameter, allowing substantial improvements with increased inference-time compute. \algo\ demonstrates strong empirical performance across 10 image-restoration tasks involving both pixel-space and latent-space diffusion models, as well as in musical source separation, even matching the performance of supervised methods. 

\section{Background}
\subsection{Diffusion models} 
\label{sec:diffusion}
DDMs define a generative process for a data distribution $\pdata{0}{}{}$ on $\rset^d$ by sequentially sampling from a series of progressively less smoothed distributions $(\pdata{t}{}{})_{t=T}^0$, starting from a highly smoothed prior $\pdata{T}{}{}$ and ending at the data distribution $\pdata{0}{}{}$. For all $s, t \in \intset{0}{T}$ with $s < t$, define the noising Markov transition kernels 
\begin{equation}
\fw{t|s}{\bx_s}{\bx_t} = \normpdf(\bx_t; (\a_t / \a_s) \bx_s, \std^2 _{t|s} \Id_\dimx) \eqsp,
\label{eq:def_gauss_transition}
\end{equation} 
where $(\a_t)_{t = 0}^T$ is monotonically decreasing with $\a_0 = 1$, $\a_T \approx 0$, and $\std^2 _{\smash{t|s}} = 1 - (\a_t / \a_s)^2$. 
Each smoothed distribution is a noised version of $\pdata{0}{}{}$ and has density  
$\pdata{t}{}{\bx_t} \eqdef \int \fw{t|0}{\bx_0}{\bx_t} \pdata{0}{}{\bx_0} \, \rmd \bx_0$. The final distribution $\pdata{T}{}{}$ is close to $\gauss(\zero_\dimx, \Id_\dimx)$.   Moreover, define the backward Markov transition kernels $\pdata{s|t}{\bx_t}{\bx_s} \propto \pdata{s}{}{\bx_s} \fw{t|s}{\bx_s}{\bx_t}$ with $s < t$.
Note that for all $\ell < s$, the backward transitions satisfy 
\begin{equation}
\label{eq:back_chapman}
\pdata{\ell|t}{\bx_t}{\bx_\ell} = \int \pdata{\ell|s}{\bx_s}{\bx_\ell} \pdata{s|t}{\bx_t}{\bx_s} \, \rmd \bx_s \eqsp.
\end{equation}
Consecutive distributions $\pdata{t}{}{}$ and $\pdata{t+1}{}{}$ are linked through the identity $\pdata{t+1}{}{\bx_{t+1}} = \int \fw{t+1|t}{\bx_t}{\bx_{t+1}} \pdata{t}{}{\bx_t} \, \rmd \bx_t$. 
Hence, given a sample $\bX_{t+1} \sim \pdata{t+1}{}{}$, $\bX_t \sim \pdata{t|t+1}{\bX_{t+1}}{\cdot}$ is an exact sample from $\pdata{t}{}{}$. 
This procedure defines a generative model, in the sense that the last state $X_0$ of the Markov chain $(\bX_{t})_{t = T} ^0$,  where the initial state $X_T$ is sampled from $\pdata{T}{}{}$, is a sample from $\pdata{0}{}{}$.

However, simulating the backward transitions is impracticable in most applications, so the following Gaussian approximation is used in practice. 
First, for $s \in \intset{1}{t-1}$, define the conditional density of $\bX_s$ given $\bX_0$ and $\bX_{t}$: 
\begin{align}
    \label{eq:bridge}
    & \fw{s|0, t}{\bx_0, \bx_{t}}{\bx_s} \\
    & \hspace{.2cm} = \normpdf(\bx_s; \gamma_{t|s} \a_{s|0} \bx_0 + (1 - \gamma_{t|s}) \a^{-1} _{t|s} \bx_{t}, \std^2 _{s|0,t} \Id_\dimx) \eqsp,\nonumber 
\end{align}
where $\gamma_{t|s} \eqdef \std^2 _{t|s} / \std^2 _{t|0}$ and $\std^2 _{s|0,t} \eqdef \std^2 _{t|s}  \std^2 _{s|0} / \std^2 _{t|0}$. Next, define by $\denoiser{t+1}{}{\bx_{t+1}} \eqdef \int \bx_0 \, \pdata{0|t+1}{\bx_{t+1}}{\bx_0} \, \rmd \bx_0$ the conditional expectation of $\bX_0$ given $\bX_{t+1} = \bx_{t+1}$ (referred to as the denoiser). Denote by $\denoiser{t+1}{}{}[\param]$  a parametric approximation of $\denoiser{t+1}{}{}$. 
Following \citet{ho2020denoising} and given an approximate sample $\smash{\vX_{t+1}}$ from  $\pdata{t+1}{}{}$, sampling from 
the bridge kernel $\fw{t|0, t+1}{\denoiser{t+1}{}{\vX_{t+1}}[\param], \smash{\vX_{t+1}}}{}$, where $\bx_0$ is replaced by the estimate $\denoiser{t+1}{}{\vX_{t+1}}[\param]$, 
yields an approximate sample from $\pdata{t}{}{}$. The complete sampling process proceeds as follows: first, $\hat\bX_T \sim \gauss(\zero_\dimx, \Id_\dimx)$; then, recursively, for every $t \geq 1$, $\smash{\hat\bX_t \sim \pdata{t|t+1}{\hat\bX_{t+1}}{\cdot}[\param]}$, where for all $s < t$, 
 \begin{equation}
    \label{eq:ddpm-transition}
    \pdata{s|t}{\bx_{t}}{\bx_s}[\param] \eqdef \fw{s|0, t}{\denoiser{t}{}{\bx_{t}}[\param], \bx_{t}}{\bx_s} \eqsp.
\end{equation} 
The final sample is defined as $\hat\bX_0 \eqdef \denoiser{1}{}{\hat\bX_1}[\param]$ and serves as an approximate sample from $\pdata{0}{}{}$. The parametric approximations 
of the denoisers are trained by minimizing, with respect to the parameter $\param$, an $L_2$ denoising loss across all time steps. Finally, using the Tweedie formula 
\cite{robbins1956empirical}, we obtain the identity $\denoiser{t}{}{\bx_t} = \a^{-1}_t \big(\bx_t + \std^2_t \nabla \log \pdata{t}{}{\bx_t}\big)$. Consequently, the trained denoisers not only serve as generative models but also provide parametric approximations of the score functions $\nabla \log \pdata{t}{}{\bx_t}$.

\subsection{Training-free guidance.}
After training a diffusion model for the data distribution $\pdata{0}{}{}$, it can be leveraged through \emph{guidance} to address various downstream tasks without the need for additional fine-tuning. This line of research was pioneered in the seminal works of \citet{song2019generative}, \citet{kadkhodaie2020solving}, \citet{song2021score}, and \citet{kawar2021snips}, where the sampling process described in the previous section is adapted on-the-fly to address Bayesian inverse problems. In this setting, the user observes a realization $\obs$ of a random variable $\bY \in \rset^\dimobs$, assumed to be drawn from the distribution with density 
$p_{\bY}(\obs) \eqdef \int \pot{0}{\bx}  \pdata{0}{}{\bx} \, \rmd \bx$, where $\pot{0}{\bx}$ is a likelihood term that encapsulates the knowledge of the \emph{forward model}. A typical example is inverse problems with Gaussian noise, \emph{i.e.} $\pot{0}{\bx} = \normpdf(\obs; \bfA(\bx), \bm\Sigma_\obs)$, where $\bfA : \rset^\dimx \to \rset^\dimobs$ and $\bm\Sigma_\obs$ is a covariance matrix. 
The objective is to recover plausible underlying signals $\bx$, for which prior information is encoded in $\pdata{0}{}{}$. This recovery is achieved by sampling from the posterior distribution
\[
\post{0}{}{\bx_0} \propto \pot{0}{\bx_0}  \pdata{0}{}{\bx_0}\eqsp.
\]
A common approach to constructing a sampler for this posterior distribution is to adopt the diffusion model framework by sequentially sampling from the smoothed distributions $\post{T}{}{}, \dotsc, \post{1}{}{}$, which are defined analogously to those introduced in the previous section:
\begin{equation}
    \label{eq:smoothed-posterior}
    \post{t}{}{\bx_t} \eqdef \int \fw{t|0}{\bx_0}{\bx_t}  \post{0}{}{\bx_0} \, \rmd \bx_0 \eqsp.
\end{equation}
Following the derivations above, sampling these distributions backwards in time is feasible provided that the conditional denoisers $(\denoiser{t}{}{}[\obs])_{t=1}^T$ are accessible. Each conditional denoiser is  defined by
\[
\denoiser{t}{}{\bx_t}[\obs] \eqdef \int \bx_0 \, \post{0|t}{\bx_t}{\bx_0} \, \rmd \bx_0,
\]
where the conditional posterior $\post{0|t}{\bx_t}{\bx_0}$ is given by
$\post{0|t}{\bx_t}{\bx_0} \propto \post{0}{}{\bx_0} \fw{t|0}{\bx_0}{\bx_t}$.
By analogy with the smoothed distributions defined for the prior, we obtain that
\begin{align}
    \label{eq:posterior-prior}
\post{t}{}{\bx_t} & \propto \int  \pot{0}{\bx_0} \fw{t|0}{\bx_0}{\bx_t} \pdata{0}{}{\bx_0} \, \rmd \bx_0 \nonumber \\
& \propto \pot{t}{\bx_t} \pdata{t}{}{\bx_t}, 
\end{align}
where 
\begin{equation}
    \label{eq:pot-defn}
    \pot{t}{\bx_t} \eqdef \int \pot{0}{\bx_0} \pdata{0|t}{\bx_t}{\bx_0} \, \rmd \bx_0, 
\end{equation} and we used that $\pdata{0}{}{\bx_0} \fw{t|0}{\bx_0}{\bx_t} =  \pdata{0|t}{\bx_t}{\bx_0} \pdata{t}{}{\bx_t}$.
Next, using the Tweedie formula, the posterior and prior denoisers can be related as 
\begin{equation} 
    \label{eq:posterior-denoiser}
    \denoiser{t}{}{\bx_t}[\obs] = \denoiser{t}{}{\bx_t} + \a^{-1} _t \std^2 _t \nabla \log \pot{t}{\bx_t} \eqsp. 
\end{equation}
This shows that in order to estimate $\denoiser{t}{}{}[\obs]$ we only need to estimate $\nabla \log \pot{t}{}$, as we already have access to a pre-trained parametric approximation of $\denoiser{t}{}{}$. A widely used approximation of this likelihood term \cite{ho2022video, chung2023diffusion}, which we will also use in the next section, is 
\begin{equation}
    \label{eq:dps}
    \hpot{t}{\bx_t}[\param] \eqdef \pot{}{\denoiser{t}{}{\bx_t}[\param]}, 
\end{equation} 
and amounts to approximating the posterior distribution $\pdata{0|t}{\bx_t}{\cdot}$ with a Dirac mass at $\denoiser{t}{}{\bx_t}[\theta]$, which we express as $\pdata{0|t}{\bx_t}{\cdot} \approx \delta_{\denoiser{t}{}{\bx_t}[\theta]}$ with a slight abuse of notation. To improve the quality of the sample, $\nabla \log \hpot{t}{\bx_t}[\param]$ is rescaled with a suitable weight (possibly depending on $\bx_t$); see \citep[Equation 8]{ho2022video} and \citep[Algorithm 1]{chung2023diffusion}. 
    Compared to previous works, methods that perform guidance using the approximation \eqref{eq:dps} incur additional computational overhead due to the calculation of a vector-Jacobian product when evaluating $\nabla \log \hpot{t}{\bx_t}[\param]$. Nevertheless, subsequent works using this approximation have shown remarkable improvements in performance across various applications; see for example \cite{song2022pseudoinverse, rozet2023score, yu2023freedom, wu2023practical,jiang2023motiondiffuser,rozet2024learning, moufad2024variational}. 

\section{Guidance with mixtures}
We now present our main contribution: a novel density approximation of the smoothed posteriors $\post{t}{}{}$. Since their scores are intractable, gradient-based samplers cannot be directly applied. Thus, we develop a Gibbs sampling scheme targeting a data augmentation of our smoothed posterior approximation, marking our second key contribution. 

In the next two sections we develop an algorithm for the ideal generative model, \emph{i.e.}, we assume that we have at hand the true marginals $\pdata{t}{}{}$ and backward transitions $\pdata{s|t}{}{}$; then, in \Cref{sec:implementation}, we provide a practical implementation involving the learned model. 

\subsection{Guidance approximation} 
We begin by extending the likelihood approximation in \eqref{eq:dps} introduced by \citet{ho2022video, chung2023diffusion}. 
First, note that by combining \eqref{eq:back_chapman} with \eqref{eq:pot-defn}, we find that $\pot{t}{}$ satisfies
\[
\pot{t}{\bx_t} = \int \pot{s}{\bx_s} \pdata{s|t}{\bx_t}{\bx_s} \, \rmd \bx_s, 
\]
for all $t\in \intset{1}{T}$ and $s \in \intset{0}{t-1}$. Thus, we obtain $t-1$ different approximations of $\pot{t}{}$ by simply setting,  for $s \in \intset{1}{t-1}$,
\begin{equation}
    \label{eq:hpot-def}
\hpot{t}{\bx_t}[s] \eqdef \int \hpot{s}{\bx_s}  \pdata{s|t}{\bx_t}{\bx_s} \, \rmd \bx_s \eqsp, \quad t \geq 2 \eqsp,
\end{equation}
where $\hpot{s}{}$ denotes the counterpart of \eqref{eq:dps}, with the learned denoiser $\denoiser{s}{}{}[\param]$ replaced by the true denoiser $\denoiser{s}{}{}$. 
In contrast to $\hpot{t}{\cdot}[\param]$ in \eqref{eq:dps}, the scores of these approximations remain intractable even when the approximate model is used, as they involve an intractable integral.
Instead, we take a different approach and use $ \hpot{t}{\cdot}[s] $ to define density approximations 
\begin{equation}
    \label{eq:mixture-component}
\hpost{t}{}{\bx_t}[s] \eqdef \frac{\hpot{t}{\bx_t}[s] \pdata{t}{}{\bx_t}}{\int \hpot{t}{\bx^\prime_t}[s] \pdata{t}{}{\bx^\prime_t} \, \rmd \bx^\prime_t} 
\end{equation}
of the smoothed posteriors $\post{t}{}{}$. Since we have $t-1$ such approximations, we consider a weighted mixture approximation of $\post{t}{}{}$ defined, for $t \geq 1$, as 
\begin{equation}
    \label{eq:posterior-approximation}
    \hpost{t}{}{\bx_t}  \eqdef \sum_{s = 1}^{t-1} \wght^s _t \hpost{t}{}{\bx_t}[s] \eqsp, 
\end{equation}
where $(\wght^s _t)_{s = 1} ^{t-1}$ are time-dependent weights and $\sum_{s = 1}^{t-1} \wght^s _t = 1$ with $\wght^s _t \geq 0$. 
Then, to sample approximately from $\post{0}{}{}$ we can use a sequential sampling procedure that runs through the intermediate distributions $\hpost{T}{}{}, \dotsc, \hpost{1}{}{}$. Similar sequential sampling procedures from posterior sequences different from $(\post{t}{}{})_t$ have also been utilized in previous works. For instance, \citet{wu2023practical,rozet2023score} use $\hpost{t}{}{\bx_t} \propto \hpot{t}{\bx_t} \pdata{t}{}{\bx_t}$. Our approach differs from these prior works by employing a mixture-based formulation with non-standard approximations of $\post{t}{}{}$. 
However, sampling from $\post{t}{}$ remains a non-trivial challenge. Indeed, a naive procedure would consist in sampling an index $s \sim \mbox{Categorical}(\{ \wght^\ell _t \}_{\ell = 1}^{t-1})$ and then use an approximate sampler only for $\hpost{t}{}{}[s]$. However, we must address the intractability of both $\hpost{t}{}{\cdot}[s]$ and its score. In the next section, we propose a method that fully overcomes these challenges. The discussion on selecting the weight sequence $(\wght^s_t)_{s=1}^{t-1}$ is postponed until after presenting the algorithm, more specifically at the beginning of \Cref{sec:experiments}.

\subsection{Data augmentation and Gibbs sampling}

We first detail how to sample from a single component $\hpost{t}{}{}[s]$ of the mixture \eqref{eq:posterior-approximation} for given $t \in \intset{2}{T}$ and $s \in \intset{1}{t-1}$. 
Consider first the extended distribution
\begin{multline}
    \label{eq:extended}
    \epost{0, s, t}{}{\bx_0, \bx_s, \bx_t} \\ \propto \pdata{0|s}{\bx_s}{\bx_0} \hpot{s}{\bx_s} \pdata{s|t}{\bx_t}{\bx_s} \pdata{t}{}{\bx_t} \eqsp.
\end{multline}
From the definitions in \eqref{eq:posterior-approximation} and \eqref{eq:hpot-def} it follows that $\hpost{t}{}{}[s]$ is the $\bx_t$-marginal of \eqref{eq:extended}, \emph{i.e.}, 
\[
\hpost{t}{}{\bx_t}[s] = \int \epost{0, s, t}{}{\bx_0, \bx_s, \bx_t} \, \rmd \bx_0 \, \rmd \bx_s.
\]
To sample approximately from $\hpost{t}{}{}[s]$, we employ a sampler targeting $\epost{0, s, t}{}{}$ and retain only the $\bx_t$-coordinate of its output. 

Specifically, we use a Gibbs sampler (GS) \cite{geman1984stochastic, casella1992explaining,gelfand2000gibbs},  which, in this context, constructs a Markov chain $(\bXy^r _0, \bXy^r _s, \bXy^r _t)_{r\in\nset}$ having $\epost{0, s, t}{}{}$ as its stationary distribution. Denote by $\epost{\smash{s|0, t}}{}{}$, $\epost{t|0, s}{}{}$, and $\epost{0|s, t}{}{}$ its three full conditionals given by 
 \begin{equation*}
    \label{eq:conditionals}
    \begin{cases}
        \begin{aligned}
            \epost{\smash{s|0, t}}{\bx_0, \bx_t}{\bx_s} & \txts = \frac{\hpot{s}{\bx_s} \fw{s|0, t}{\bx_0, \bx_t}{\bx _s}}{\int \hpot{s}{\bx^\prime _s} \fw{s|0, t}{\bx_0, \bx_t}{\bx^\prime _s} \, \rmd \bx^\prime _s} \eqsp,\\
            \epost{t|0, s}{\bx_0, \bx_s}{\bx_t} & = \fw{t|s}{\bx_s}{\bx_t}, \\
            \epost{0|s, t}{\bx_s, \bx_t}{\bx_0} & = \pdata{0|s}{\bx_s}{\bx_0}.
        \end{aligned}
    \end{cases}
\end{equation*}
The proof of this fact is postponed to \Cref{apdx-sec:conditionals}.
Then, one step of the associated (deterministic scan) GS 
is described in
\Cref{algo:extended-gibbs}.

\begin{algorithm}[h]
    \caption{Gibbs sampler targeting \eqref{eq:extended}}
    \begin{algorithmic}[1]
        \STATE {\bfseries Input:} $(\bXy^r _0, \bXy^r _s, \bXy^r _t)$
        \STATE draw $\bXy^{r+1} _s \sim \epost{\smash{s|0, t}}{\bXy^r _0, \bXy^r _t}{}$ \label{step:bridge}
        \STATE draw $\bXy^{r+1} _t \sim \fw{t|s}{\bXy^{r+1} _s}{}$ \hfill {\small\texttt{//noising}}
        \STATE draw $\bXy^{r+1} _0 \sim \pdata{0|s}{\bXy^{r+1} _s}{}$ \label{step:denoising} \hfill {\small\texttt{//denoising}}  
    \end{algorithmic}
    \label{algo:extended-gibbs}
\end{algorithm}

Since \eqref{eq:extended} admits $\hpost{t}{}{}[s]$ as marginal, the process $(\bXy^r _t)_{r\in\nset}$ will, at stationarity of $(\bXy^r _0, \bXy^r _s, \bXy^r _t)_{r\in\nset}$, have $\hpost{t}{}{}$ as a marginal distribution. We provide basic background on Gibbs sampling in \Cref{apdx-sec:gibbs} and refer the reader to \cite{casella1992explaining}. 

It is clear from \Cref{algo:extended-gibbs} that only the update of $\bXy_s^r$ depends on the observation $\obs$, while the updates of the remaining components are sampled via (i) a noising step involving the forward transition \eqref{eq:def_gauss_transition}, which can be performed exactly, and (ii) a denoising step involving the prior diffusion model, which can be approximated using the pre-trained model. 

Finally, to target the mixture \eqref{eq:posterior-approximation}, we first sample the mixture index $s \sim \mbox{Categorical}\big(\{ \wght^\ell _t \}_{\ell=1}^{t-1}\big)$, which determines the component of the mixture $\hpost{t}{}{\bx_t}$. 
Next, we apply \Cref{algo:extended-gibbs} $R$ times to update the remaining coordinates, treating $s$ a fixed, and output the result $\bXy^R _t$. 
Note that an alternative to our method would be to consider a Gibbs sampler for which one of its marginal is directly the mixture 
\eqref{eq:posterior-approximation} incorporating also the mixture index $s$ as a state. However, this would then require sweeping over all states $(\bXy_0, \dotsc, \bXy_t)$, rendering it computationally expensive and impractical. We discuss other possible data augmentations and their limitations in \Cref{apdx-sec:data-aug}.

\begin{algorithm}[t]
    \caption{\algoname}
    \begin{algorithmic}[1]
        \STATE {\bfseries Input:} Timesteps $(t_i)_{i = 1} ^K$ with $t_1 > 1$ and $t_K = T$, Gibbs repetitions $\gibbsReps$, DDPM steps $M$, gradient steps $G$, probabilities $\{\wght^\ell _{t_i}\}^{2, t_{i-1}} _{i=K, \ell=1}$
        \STATE $\vX_{t_K} \sim \gauss(\zero_d, \Id_d)$
        \STATE $\vX_{0} \gets \denoiser{t_K}{}{\vX_{t_K}}[\param]$, $\, \vX^* _0 \gets \vX_0$

        \FOR{$i=K$ to $2$}
            \STATE $s \sim \mbox{Categorical}(\{\wght^\ell _{t_i}\}_{\ell = 1}^{t_{i-1}})$
            \STATE $\vX _0 \gets \vX^* _0$ \label{line:x0init}
            \STATE $\vX _{t_i} \sim \fw{t_i|0, t_{i+1}}{\vX^{*} _0, \vX _{t_{i+1}}}{}$ \label{line:xtinit}
            \FOR{$r = 1$ {\bfseries to} $R$}
                \STATE $\vX_s \gets \vifn\big(\vX_0, \vX_{t_i}, s, G\big)$ \label{line:gaussvi} \hfill {\small\texttt{//see \ref{apdx-sec:vi-approx}}}
                \STATE $\vX _0 \gets \ddimfn_{}(\vX _s, s, M)$ \label{line:ddpm}
                \STATE $\vX _{t_i} \sim \fw{t_i|s}{\vX _s}{}$
            \ENDFOR
            \STATE $\vX^* _{0} \gets \vX _0$
        \ENDFOR
        \STATE {\bfseries Output:} $\bX^* _0$
    \end{algorithmic}
    \label{algo:midpoint-gibbs}
\end{algorithm}

\begin{figure} 
    \centering
    \includegraphics[width=.45\textwidth]{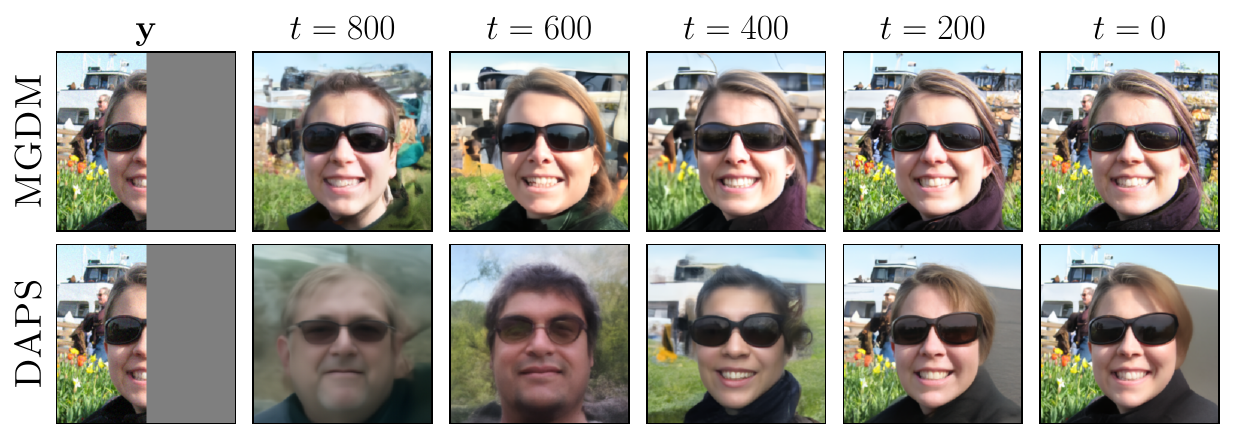}
    \captionsetup{font=small}
    \caption{Evolution of $\vX^* _0$ throughout the iterations for \algo\ and \daps\ \cite{zhang2024daps}. }
    \label{fig:running_x0}
\end{figure}
\label{sec:implementation}
\subsection{Practical implementation}
For simplicity, we present the algorithm in the case where we progressively sample from each $\hpost{t}{}{}$ for $t \in \intset{2}{T}$. In practice, however, we subsample a small number $K$ of timesteps $(t_i)_{i = K}^1$, with $t_1 > 1$ and $t_K = T$, and apply the algorithm only to $( \hpost{t_i}{}{} )_{i = K}^i$.

The denoising step in \Cref{algo:extended-gibbs} can be approximated by sampling from the learned diffusion model. To reduce runtime, we again subsample a small number of timesteps $\{s_i\}_{i = 0}^M \subset \intset{0}{s-1}$, ensuring that  $s_0 = 0$ and $s_M = s$. We then generate  $(X_{s_i})_{i = 0}^M$ by sampling iteratively $X_{s_i} \sim \pdata{s_i | s_{i+1}}{X_{s_{i+1}}}{\cdot}[\param]$ and retaining only $X_{s_0}$.
This operation is referred to as $\ddimfn(\cdot, s, M)$ on Line~\ref{line:ddpm} in \Cref{algo:midpoint-gibbs}.
As for the step involving $\hpost{s|0, t}{}{}$,
we follow \citet{moufad2024variational} and sample approximately by fitting a Gaussian variational approximation.
More specifically, given $(\bx_0, \bx_t)$, we draw from the Gaussian variational approximation $\smash{\vi{s|0, t}{} \eqdef \gauss\big(\vmu_{s|0, t}, \diag(\rme^{\vlstd_{s|0,t}})\big)}$ where the parameters $\vparam_{s|0, t} \eqdef (\vmu_{s|0, t}, \vlstd_{s|0,t}) \in \rset^{\dimx} \times \rset^{\dimx}$ are obtained by optimizing the right-hand side of 
\begin{multline*}
    \kldivergence{\vi{s|0,t}{}}{\post{s|0,t}{\bx_0, \bx_t}{}} \\ \approx - \pE \big[ \log \hpot{s}{\vX^{\vparam} _s}[\param] \big] + 
    \kldivergence{\vi{s|0,t}{}}{\fw{s|0,t}{\bx_0, \bx_t}{}},  
\end{multline*}
where $\vX^\vparam _s \sim \vi{s|0, t}{}$. The gradient of this quantity can be estimated straightforwardly using the reparameterization trick \cite{kingma2013auto}. The initial parameters $\vparam_{s|0, t}$ are set to the mean and covariance of $\fw{s|0,t}{\bx_0, \bx_t}{\cdot}$ defined in \eqref{eq:bridge}. 
This step corresponds to the $\vifn$ routine in \Cref{algo:midpoint-gibbs} and is detailed in \Cref{apdx-sec:vi-approx}. Regarding the initialization of the GS for $\hpost{t}{}{}$, we use the output of the previous GS targeting $\hpost{t+1}{}{}$; see Lines \ref{line:x0init} and \ref{line:xtinit} in \Cref{algo:midpoint-gibbs}. We maintain a running variable $\vX^* _0$, which is iteratively updated and serves as the initialization for the other variables at the beginning of each loop iteration. It is also the output of the algorithm. Indeed, note that the last distribution to which we apply the GS is $\epost{0, 1, 2}{}{\bx_0, \bx_1, \bx_2}$ of which the $\bx_0$-marginal is proportional to $\pdata{0}{}{\bx_0} \int \hpot{1}{\bx_1} \fw{1|0}{\bx_0}{\bx_1} \, \rmd \bx_1$. 
Since the Gaussian density $\fw{1|0}{\bx_0}{\cdot}$ has a very small variance and $\hpot{1}{} \approx \pot{0}{}$, we may assume that $ \int \hpot{1}{\bx_1} \, \fw{1|0}{\bx_0}{\bx_1} \rmd \bx_1 \approx \pot{0}{\bx_0}$ and hence that the posterior $\post{0}{}{}$ of interest is approximately the $\bx_0$-marginal of the last extended distribution. As a result, we can take the $\bx_0$-coordinate of the output of the last GS, which is $\vX_0$ and hence $\vX^* _0$, as an approximate sample from $\post{0}{}{}$. In the first row of \Cref{fig:running_x0} we display the evolution of $\vX^* _0$ throughout the iterations with a DDM pre-trained on the \ffhq\ dataset. It is seen that the algorithm reaches a plausible reconstruction of $\obs$ rather fast, at $t = 800$ with $T = 1000$, and then spends the remaining iterations refining the details.  As a comparison, the DAPS algorithm proposed by \citet{zhang2024daps}, which displayed in the second row, also maintains a running variable at time $0$ that serves as output to the algorithm.
\section{Related works}
\begin{figure*}[!t]
    \centering 
    \includegraphics[width=\textwidth]{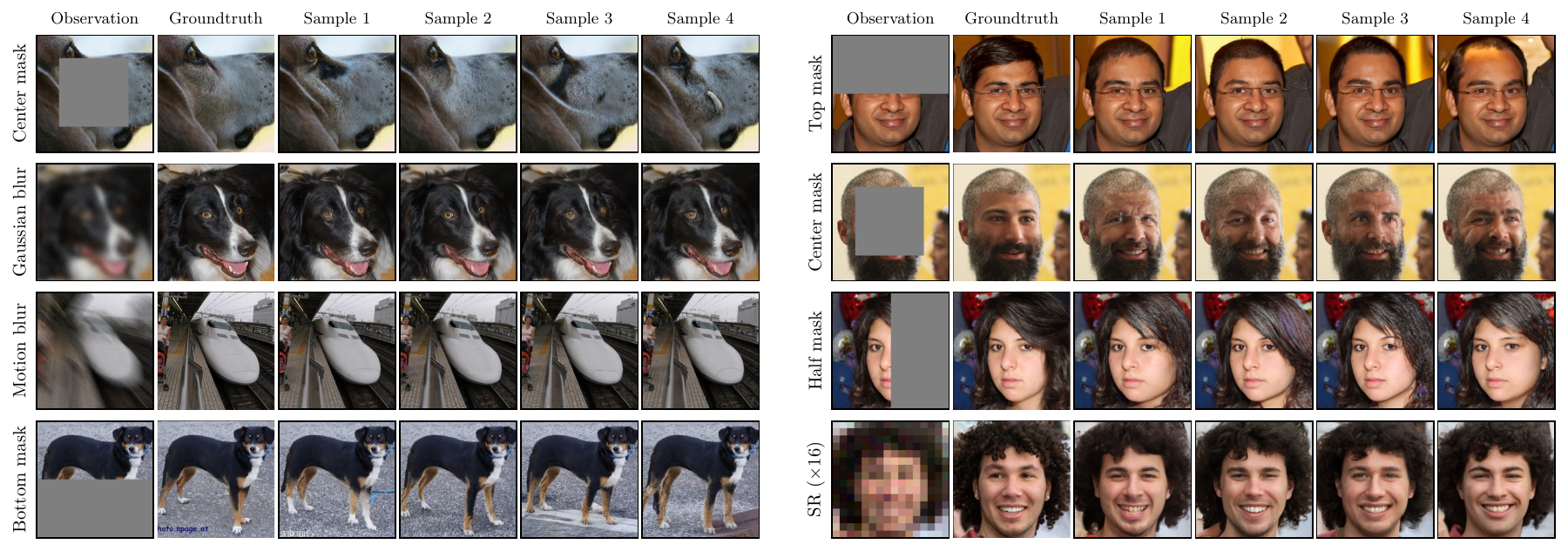}
    \captionsetup{font=small}
    \caption{\algo\ sample images for various tasks on \imagenet\ (left) and \ffhq\ (right) datasets.}
    \label{fig:main-reconstructions}
\end{figure*}
\paragraph{Alternative likelihood approximations.} In addition to this work, several other papers introduce alternative approximations of $\pot{t}{}$. \cite{song2022pseudoinverse} proposes a Gaussian approximation of $\pdata{0|t}{}{}$ with mean given by the denoiser $\denoiser{t}{}{}[\param]$ and covariance being left as a hyperparameter. For linear inverse problems with Gaussian noise, the likelihood $\pot{0}{}$ can be integrated exactly against this Gaussian approximation, providing an alternative approximation of $\pot{t}{}$. \citet{finzi2023user, stevens2023removing, boys2023tweedie} use that the covariance of $\pdata{0|t}{\bx_t}{\cdot}$ is proportional to the Jacobian of the denoiser \cite{meng2021estimating}. Computing the score of the resulting likelihood approximation, for linear inverse problems, is prohibitively expensive. To mitigate this, these works and subsequent ones assume that the Jacobian of the denoiser is constant with respect to $\bx_t$. Despite this simplification, the score approximation still involves an expensive matrix inversion. \citet{boys2023tweedie} use diagonal approximation of the covariance based on its row sums. \citet{rozet2024learning} use conjugate gradient to perform the matrix inversion efficiently. For general likelihoods $\pot{0}{}$, \citet{song2023loss} use Gaussian approximations of \citet{song2022pseudoinverse} to estimate $\pot{t}{}$ using a standard Monte Carlo approach. For latent diffusion models, \citet{rout2024solving} apply the approximation in \eqref{eq:dps} together with a regularization term that penalizes latent variables deviating from fixed points of the decoder-encoder composition. \citet{moufad2024variational} propose a general method for both vanilla and latent space diffusion models. At step $t$ of the diffusion process they first sample, at an intermediate timestep $s < t$, a state conditionally on $\obs$ with the approximation \eqref{eq:dps}, before returning back to the timestep $t$. In \Cref{apdx-sec:comparisons} we explain in more details how the present work differs from this method. 

\paragraph{Asymptotically exact methods.} \citet{trippe2023diffusion,wu2023practical, cardoso2024monte, dou2024diffusion, corenflos2024conditioning,li2024derivative} use the sequential Monte Carlo (SMC) framework to construct an empirical approximation of the posterior distribution represented by $N$ samples. The samples undergo transitions guided by user-defined updates, are reweighted using an appropriate importance weight, and are subsequently resampled to focus computational effort on the most promising candidates. The performance of these methods improves by scaling the number of samples $N$, which impacts both the memory requirement and compute time. As evidenced by the experiments in the next section, our method improves by increasing the number of Gibbs steps, which impacts only the runtime. 

\paragraph{Gibbs sampling approaches.} The recent works \cite{wu2024principled, xu2024provably} on \pnpdm\ also propose a Gibbs sampling-inspired algorithm. They consider (within the variance exploding framework) the distribution sequence $(\tilde{\pi}^\obs _t)_{t = 0} ^T$, where each distribution $\tilde{\pi}^\obs _t(\bx_t) \propto \pot{\textcolor{purple}{0}}{\bx_t} \pdata{t}{}{\bx_t}$ is the $\bx_t$-marginal of the extended distribution 
$$\tilde{\pi}^\obs _{0, t}(\bx_0, \bx_t) \propto \pot{0}{\bx_t} \pdata{0}{}{\bx_0} \fw{t|0}{\bx_0}{\bx_t}\eqsp.$$ 
As its full conditionals are $\smash{\tilde\pi^\obs _{0|t}(\bx_0|\bx_t)} = \pdata{0|t}{\bx_t}{\bx_0}$ and $\tilde\pi^\obs _{t|0}(\bx_t|\bx_0) \propto \pot{0}{\bx_t} \fw{t|0}{\bx_0}{\bx_t}$, the GS targeting this joint distribution also proceeds with a prior denoising step. On the other hand, sampling from $\tilde\pi^\obs _{t|0}(\cdot|\bx_0)$ can be performed exactly when $\pot{0}{}$ is the likelihood of a linear inverse problem with Gaussian noise, since $\fw{t|0}{\bx_0}{\cdot}$ is a Gaussian distribution. For more general problems, this step can be implemented using MCMC methods; see \emph{e.g.} \citep[Algorithms 3 \& 4]{xu2024provably}. Compared to our algorithm, \pnpdm\ has a lower memory footprint because it does not require a vector-Jacobian product as it uses the likelihood $\pot{0}{}$ instead of $\hpot{t}{}$.
However, as we show in the next section, this comes at the cost of performance, especially when using latent diffusion models. The \textsc{RePaint} algorithm \cite{lugmayr2022repaint}, which applies to noiseless linear inverse problems, uses noising and denoising steps repeatidly and can also be viewed as a variant of a specific Gibbs sampler. Finally, the recently proposed \daps\, \cite{zhang2024improving} can also be related to a Gibbs sampler targeting a specific sequence of distributions. Further details and comparisons to {\algo} are provided in \Cref{apdx-sec:comparisons}.

\section{Experiments}
\label{sec:experiments}
\begin{table*}[t]
    \centering
    \captionsetup{font=small}
    \caption{Mean LPIPS for linear/nonlinear imaging tasks on the \ffhq\ and \imagenet\ datasets with $\stdobs = 0.05$. Lower metrics are better.
    }
    \resizebox{\textwidth}{!}{
    \begin{tabular}{l cccccccc | cccccccc}
        \toprule
        \vspace{1mm}
        & \multicolumn{8}{c}{\bf{\ffhq}} & \multicolumn{8}{c}{\bf{\imagenet}} \\
        \textbf{Task} & {\bf \algo} & \dps & \pgdm & \ddnm & \diffpir & \reddiff & \daps & \pnpdm \ &\ {\bf \algo} & \dps & \pgdm & \ddnm & \diffpir & \reddiff & \daps & \pnpdm \\
        \midrule
        SR ($\times 4$)        & \first{0.09} & \first{0.09} & 0.30 & \third{0.15} & \second{0.10} & 0.39 & 0.16 & \second{0.10} \ &\ \second{0.26} & \first{0.25} & 0.56 & 0.34 & \third{0.31} & 0.57 & 0.37 & 0.66 \\
        SR ($\times 16$)       & \second{0.24} & \first{0.23} & 0.42 & 0.33 & \first{0.23} & 0.55 & 0.40 & \third{0.29} \ &\ \third{0.55} & \first{0.44} & 0.62 & 0.71 & \second{0.50} & 0.85 & 0.75 & 1.03 \\
        Box inpainting         & \first{0.10} & 0.17 & 0.17 & \second{0.12} & 0.14 & 0.19 & \third{0.13} & 0.18 \ &\ \first{0.23} & 0.35 & \third{0.29} & \second{0.28} & 0.30 & 0.36 & 0.30 & 0.42 \\
        Half mask              & \first{0.20} & \third{0.24} & \third{0.24} & \second{0.23} & 0.25 & 0.28 & \second{0.23} & 0.32 \ &\ \first{0.31} & 0.40 & \second{0.34} & \third{0.38} & 0.40 & 0.46 & 0.40 & 0.54 \\
        Gaussian Deblur        & \first{0.12} & \third{0.17} & 0.87 & 0.20 & \first{0.12} & 0.24 & 0.24 & \second{0.14} \ &\ \first{0.30} & \second{0.37} & 1.00 & \third{0.45} & \first{0.30} & 0.53 & 0.59 & 0.76 \\
        Motion Deblur          & \first{0.09} & \second{0.17} & $-$ & $-$ & $-$ & 0.22 & \third{0.19} & 0.21 \ &\ \first{0.22} & \third{0.40} & $-$ & $-$ & $-$ & \second{0.39} & 0.42 & 0.52 \\
        JPEG (QF = 2)          & \first{0.14} & 0.34 & 1.12 & $-$ & $-$ & 0.32 & \second{0.22} & \third{0.29} \ &\ \first{0.38} & 0.60 & 1.32 & $-$ & $-$ & \third{0.49} & \second{0.45} & 0.56 \\
        Phase retrieval        & \first{0.11} & 0.40  & $-$ & $-$ & $-$ & \third{0.26} & \second{0.14} & 0.34 \ &\ \second{0.55} & 0.62 & $-$ & $-$ & $-$ & \third{0.61} & \second{0.50} & 0.66 \\
        Nonlinear deblur       & \first{0.27} & 0.51 & $-$ & $-$ & $-$ & 0.68 & \second{0.28} & \third{0.31} \ &\ \first{0.41} & 0.82 & $-$ & $-$ & $-$ & \third{0.66} & \first{0.41} & \second{0.49} \\
        HDR    & \second{0.12} & 0.40 & $-$ & $-$ & $-$ & 0.20 & \second{0.10} & \third{0.19} \ &\ \third{0.21} & 0.84 & $-$ & $-$ & $-$ & \second{0.19} & \first{0.14} & 0.31 \\
        \bottomrule
    \end{tabular}
    }
    \label{table:lpips-ffhq-imagenet}
\end{table*}
We evaluate \algo\ on image inverse problems using both pixel-space and latent-space diffusion, as well as on musical source separation tasks. 
For the pixel-space diffusion and the audio diffusion model, we compare \algo\ against \emph{eight} competitors: \dps\ \cite{chung2023diffusion}, \pgdm\ \cite{song2022pseudoinverse}, \ddnm\ \cite{wang2023zeroshot}, \diffpir\ \cite{zhu2023denoising}, \reddiff\ \cite{mardani2024a}, \daps\ \cite{zhang2024daps}, and \pnpdm\ \cite{wu2024pnpdm}.
In the latent space setting, we benchmark against \emph{four} competitors: \psld\ \cite{rout2024solving}, \resample\ \cite{song2024solving}, \daps\ \cite{zhang2024daps}, and \pnpdm\ \cite{wu2024pnpdm}. In Appendixes~\ref{apdx-sec:hyperparameters}-\ref{apdx:competitors}, we provide a complete formal description of the parameters of our algorithm as well as the implementation details of each competitor and its hyperparameters. We emphasize that we have tuned the parameters of our algorithm per dataset and not per task. 

\emph{Index sampling and Gibbs steps.} During the first $75\%$ of the diffusion process, at timestep $t_i$, we sample the index $s$ from $\mbox{Uniform}[\tau, t_{i-1}]$ with $\tau = 10$ to mitigate instabilities. In the final $25\%$ of the steps we set $s = t_{i-1}$ as this yields slightly improved results. On the image inverse problems we use $100$ diffusion steps with $R=1$ Gibbs step. On the source separation task we use $20$ diffusion steps with $R=6$ Gibbs steps. The choice of weight sequence $\{\wght^\ell _t\}^{2, t-1} _{t=T, \ell=1}$ plays an important role for the algorithm's performance. Intuitively, it holds that that $\hpot{t}{}[s] \approx \pot{t}{}$ when $s \approx 0$, suggesting that for all $t \in \intset{2}{T}$, the weights should be set to $0$ beyond a certain threshold to ensure that $s$ is sampled near $0$. We found, however, that this strategy does not yield good performance for our algorithm. Instead, sampling the index uniformly leads to a faster mixing. On high-dimensional image datasets, we observe that when $s$ is consistently sampled near $0$ at all iterations, the algorithm struggles to overcome the errors that accumulate at initialization, leading to suboptimal reconstructions. We provide both quantitative and qualitative evidence in \Cref{apdx-sec:weight-seq}. 

\textbf{Images.} 
We evaluate our method on a diverse set of six linear inverse problems and four nonlinear inverse problems with three different image priors with $256\times256$ resolution: the pixel-space \ffhq\ model of \citet{choi2021ilvr}, the latent-space \ffhq\ of  \citet{rombach2022high}, and the \imagenet\ model of \citet{dhariwal2021diffusion}. We use the noise level $\stdobs = 0.05$ for all tasks. The linear problems include image inpainting with two masking configurations: a $150 \times 150$ central box mask and a half-mask covering the right side of the image; Super Resolution (SR) tasks with upscaling factors of $\times 4$ and $\times 16$;
Gaussian and motion deblurring, both using a kernel size of $61 \times 61$ following the experimental setup described by \citet[Section 4]{chung2023diffusion}.
For the nonlinear setting, we consider JPEG dequantization with a quality factor of $2\%$, implemented using the differentiable operator proposed by \citet{shin2017jpeg}; phase retrieval with an oversampling factor of $\times 2$; non-uniform deblurring using the operator introduced by \citet{tran2021non-uniform-deblurring}; High Dynamic Range (HDR) reconstruction following the setup detailed in \citet[Section 5.2]{mardani2024a}. The evaluation is done on a subset of 300 validation images per dataset. For \ffhq, we use the first 300 images, while for \imagenet, we randomly sample 300 images to avoid class bias. We report the LPIPS metric \cite{zhang2018lpips} in Tables~\ref{table:lpips-ffhq-imagenet} and \ref{table:lpips-ffhq-ldm} and defer the complete tables with PSNR and SSIM to Tables~\ref{table:extended-ffhq-imagenet} and \ref{table:extended-ffhq-ldm}. For the phase retrieval task specifically, we draw 4 samples for each algorithm and keep only the best scoring one in terms of LPIPS. A similar strategy is used in \cite{chung2023diffusion,zhang2024daps,wu2024principled}. 
Across table rows, we highlight the best value in \first{\transparent{0}{tx}}, the 2\textsuperscript{nd} best in \second{\transparent{0}{tx}} and 3\textsuperscript{rd} best in \third{\transparent{0}{tx}}. We provide a large gallery of exemplar reconstructions in \Cref{apdx-sec:visual-reconstructions}. 

\begin{figure}
    \centering
    \renewcommand{\arraystretch}{1.2} 
    \resizebox{0.482\textwidth}{!}{
    \begin{tabular}{c  r}
        \raisebox{112pt}[0pt][0pt]{
        \resizebox{0.82\textwidth}{!}{
        \begin{tabular}{l cccc | c}
            \toprule
                & $R=1$ & $R=2$ & $R=4$ & $R=6$ & $R=1,G\gg1$ \\
            \midrule
            Bass & 15.46 & 18.07 & \second{18.53}  & \third{ 18.49}  & \first{19.89} \\
            Drums & 16.28 & 17.93 & \second{18.19} & \third{ 18.07} & \first{18.95} \\
            Guitar & 12.58 & 14.73 & \third{16.26} & \first{ 16.68} & \second{16.07} \\
            Piano & 11.82 & 14.34 & \third{15.38}  & \second{16.17} & \first{16.50} \\
            \bottomrule
            All & 14.03 & 16.27 & \third{17.09} & \second{17.35} & \first{17.85} \\
            \bottomrule
        \end{tabular}
        }
        }
        &
        \includegraphics[width=0.5\textwidth]{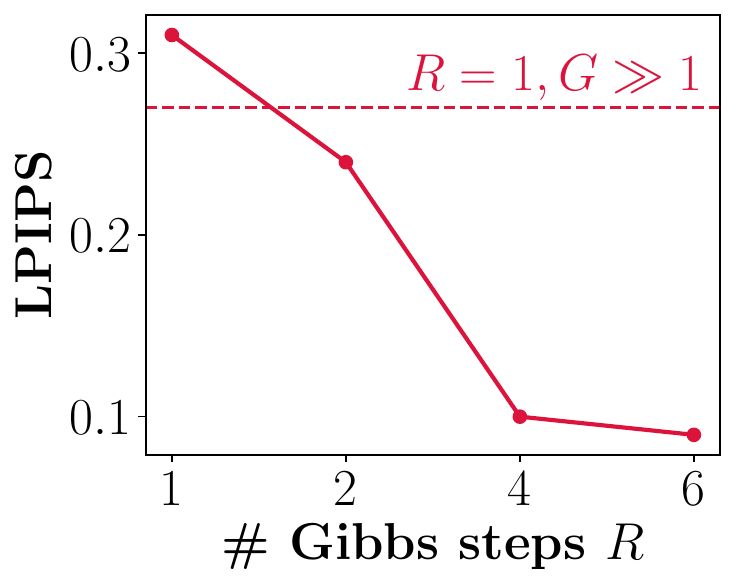}
    \end{tabular}
    }
    \vspace*{-1mm}
    \captionsetup{font=small}
    \caption{Performance of \algo\ as a function of the number of Gibbs steps $R$.
    The setup $R=1,G\gg1$ represents \algo\ with $R=1$ and a number of gradient steps resulting in a runtime equivalent to using $R=6$.
    Left: Mean \sisdri\ for multisource--audio separation task on \slakh\ test dataset.
    Right: Mean LPIPS for the phase retrieval task on \ffhq.}
    \vspace*{-3mm}
    \label{fig:scaling}
\end{figure}

\emph{Results.} Our method with a single Gibbs step consistently achieves competitive performance, ranking first on most tasks and standing out as the only approach to maintain robust performance across all tasks. On latent \ffhq, we outperform \resample\ and \psld, both of which are specifically designed for latent problems, while our method is applied seamlessly off-the-shelf without any adaptation to latent diffusion. Qualitative comparisons in \Cref{fig:main-reconstructions} and in \Cref{apdx-sec:visual-reconstructions} reveal that our method provides diverse, visually coherent and sharp reconstructions. In contrast, \daps, \ddnm\ and \diffpir, despite scoring higher in PSNR and SSIM on some tasks, provide less coherent reconstructions; see \Cref{apdx:extended-results} for a discussion and examples. 
Finally, a key strength of our algorithm is its ability to improve performance by increasing the number  $R$  of Gibbs steps. This is demonstrated for the most challenging task, phase retrieval, in \Cref{fig:scaling}. In this experiment, we compute the LPIPS using a single sample per image (instead of four) and achieve a threefold reduction in average LPIPS simply by increasing the compute time in the right direction. Indeed, increasing the number of gradient steps brings only marginal gains in this case whereas increasing the number of Gibbs steps leads to significant performance gains.
\begin{table}[t]
    \centering
    \captionsetup{font=small}
    \caption{Mean LPIPS for linear/nonlinear imaging tasks on \ffhq\ dataset with LDM prior and $\stdobs = 0.05$.
    Lower metrics are better.
    }
    \resizebox{0.45\textwidth}{!}{
    \begin{tabular}{l ccccc}
        \toprule
        Task & \algo\ & \resample & \psld & \daps & \pnpdm \\
        \midrule
        SR ($\times 4$) & \first{0.14} & \third{0.22} & \second{0.21} & 0.28 & 0.40 \\
        SR ($\times 16$) & \first{0.30} & \third{0.38} & \second{0.36} & 0.52 & 0.71 \\
        Box inpainting & \first{0.18} & \second{0.22} & \third{0.27} & 0.37 & 0.31 \\
        Half mask & \first{0.26} & \second{0.30} & \third{0.32} & 0.49 & 0.44 \\
        Gaussian Deblur & \second{0.18} & \first{0.16} & 0.59 & \third{0.32} & 0.32 \\
        Motion Deblur & \second{0.22} & \first{0.20} & 0.70 & \third{0.36} & 0.36 \\
        JPEG (QF = 2) & \first{0.23} & \second{0.26} & $-$ & \third{0.32} & 0.36 \\
        Phase retrieval & \second{0.29} & \third{0.39} & $-$ & \first{0.25} & 0.50 \\
        Nonlinear deblur & \first{0.29} & \second{0.33} & $-$ & \third{0.37} & 0.37 \\
        High dynamic range & \second{0.16} & \first{0.12} & $-$ & \third{0.24} & 0.24 \\
        \bottomrule
    \end{tabular}
    }
    \label{table:lpips-ffhq-ldm}
\end{table}

\textbf{Source separation.} 
We now consider a linear inverse problem with an audio diffusion prior that generates four dependent instrument soundtracks: bass, drums, guitar, and piano. The task involves separating the individual sources from a mixture $\obs$ of these four instruments; \emph{i.e.} denoting by $\dimx^\prime$ the dimension of one instrument soundtrack, the linear operator is $\bfA: \bx \in \rset^{4 \times \dimx^\prime} \mapsto \sum_{i = 1}^4 \bx_i \in \rset^{\dimx^\prime}$. We assume \emph{no noise} in the measurement and use the audio diffusion model of \citet{mariani2023multi}. The evaluation is conducted on the publicly available \slakh\ test dataset \cite{Manilow2019slakh} with the scale-invariant SDR improvement ($\text{SI-SDR}_\text{I}$) metric \cite{le2019sdr}. The $\text{SI-SDR}_\text{I}$ metric measures the improvement between the original audio source $\mathbf{x}_i$ and the generated source $\hat{\mathbf{x}}_i$, relative to the mixture baseline $\mathbf{y}$, \emph{i.e.} it computes the difference $\text{SI-SDR}(\mathbf{x}_i, \hat{\mathbf{x}}_i) - \text{SI-SDR}(\mathbf{x}_i, \mathbf{y})$ where 
\[
\text{SI-SDR}(\mathbf{x}_i, \hat{\mathbf{x}}_i) = 10 \log_{10} \frac{\| \alpha \mathbf{x}_i \|^2 + \epsilon}{\| \alpha \mathbf{x}_i - \hat{\mathbf{x}}_i \|^2 + \epsilon} \eqsp,
\] where $\alpha = \frac{\mathbf{x}_i ^\top \hat{\mathbf{x}}_i  + \epsilon}{\| \mathbf{x}_i \|^2 + \epsilon}$, and $\epsilon = 10^{-8}$. Following \citet[Section 5.2]{mariani2023multi}, tracks from the test dataset are evaluated using a sliding window approach with 4-second chunks and a 2-second overlap. We report the $\text{SI-SDR}_\text{I}$ metric in Table \ref{table:si-snri}. 
For this task we compare against three other competing algorithms. First, the best version of the MSDM algorithm in \cite{mariani2023multi} which uses the same pre-trained model and is directly comparable to our method. Then, the ISDM algorithm from the same paper and which relies on separate pre-trained models for each instrument, as well as the Demucs model \cite{defossez2019music}, trained with supervision to specifically solve source separation, augmented with 512 Gibbs sampling steps \cite{manilow2022improving} and is, to the best of our knowledge, considered to be state-of-the-art. We refer to it as \demucs. Finally, since the inverse problem is noiseless, we smooth it by using the likelihood $\pot{0}{\bx} = \normpdf(\obs; \bfA(\bx), \std^2 _\obs \Id_\dimobs)$ with $\std_\obs = 10^{-4}$. This smoothing is applied consistently across all competitors except the best-performing versions of MSDM and ISDM, which are tailored for noiseless problems, and \(\demucs\). The results are reported in \Cref{table:si-snri} \footnote{Due to space constraints we only show the best performing competitors and defer the complete table to \Cref{apdx:extended-results}}. 

\emph{Results.} We outperform, on average, the other training-free competitors that use the same pre-trained model by a substantial margin. In particular, we outperform the MSDM algorithm of \citet{mariani2023multi} as well as ISDM which uses a different model. With $R = 6$ Gibbs steps \algo\ falls short of matching the performance \demucs. We found instead that setting $R = 1$ and using a number of gradient steps ensuring equivalent runtime, as we did for the phase retrieval example, allows to achieve superior performance; see \Cref{fig:scaling}. It is also seen that the average \sisdri\ increases monotonically with the number of Gibbs steps. 
 
\begin{table}[t]
    \centering
    \captionsetup{font=small}
    \caption{Mean \sisdri\ on \slakh\ test dataset. The last row displays the mean over the four stems. Higher metrics are better.}
    \resizebox{0.48\textwidth}{!}{
    \begin{tabular}{l ccccc | cc}
        \toprule
        Stems  & \algo\         & \dps              & \pgdm             & \ddnm             &  \msdm            & \isdm             & \demucs \\
        \midrule
        Bass   & \first{18.49}  &  \third{16.50}    &  16.41            & 14.94             &  \second{17.12}   & 19.36     & 17.16   \\
        Drums  & 18.07          &  \third{18.29}    & 18.14             & \first{19.05}     &  \second{18.68}   & 20.90     & 19.61   \\
        Guitar & \first{16.68}  &   9.90            &  12.84            & \third{14.38}     &  \second{15.38}   & 14.70     & 17.82   \\
        Piano  & \first{16.17}  &  10.41            &  \third{12.31}    & 11.46             &  \second{14.73}   & 14.13     & 16.32   \\
        \midrule
        All    & \first{17.35}  &  13.77            &  14.92            & \third{14.96}     &  \second{16.48}   & 17.27     & 17.73   \\
        \bottomrule
    \end{tabular}
    }
    \label{table:si-snri}
\end{table}

\section{Conclusion}
We have developed a novel posterior sampling scheme for denoising diffusion priors. The proposed algorithm proceeds by sequentially sampling, using a Gibbs sampler, from a sequence of mixture approximations of the smoothed posteriors. Our experiments show that \algo\ not only matches but often surpasses state-of-the-art performance and reconstruction quality across various tasks.
Furthermore, we have demonstrated that the Gibbs sampling perspective allows favorable performance improvement with inference-time compute scaling.

This work has certain limitations that open avenues for further exploration. While we outperform the state-of-the-art on most tasks and remains competitive overall on latent diffusion,  we still fall short of what we achieve with pixel-space diffusion. We believe that bridging this gap requires a more careful selection of the weight sequence. More broadly, an observation-driven approach to sampling the index could further enhance \algo. A second limitation is that our methodology does not extend to ODE-based samplers or DDIM, and adapting related ideas to these methods is an interesting research direction. Finally, like all existing methods relying on \eqref{eq:dps}, our approach incurs a higher memory cost compared to unconditional diffusion. It remains an open question whether the vector-Jacobian product can be eliminated without compromising performance.

\nocite{langley00}

\clearpage
\newpage
\section*{Acknowledgements}
The work of Y.J. and B.M. has been supported by Technology Innovation Institute (TII), project Fed2Learn. The work of Eric Moulines has been partly funded by the European Union (ERC-2022-SYG-OCEAN-101071601). Views and opinions expressed are however those of the author(s) only and do not necessarily reflect those of the European Union or the European Research Council Executive Agency. Neither the European Union nor the granting authority can be held responsible for them. This work was granted access to the HPC resources of IDRIS under the allocation 2025-AD011015980 made by GENCI.
\bibliography{bibliography.bib}

\begin{thebibliography}{60}
\providecommand{\natexlab}[1]{#1}
\providecommand{\url}[1]{\texttt{#1}}
\expandafter\ifx\csname urlstyle\endcsname\relax
  \providecommand{\doi}[1]{doi: #1}\else
  \providecommand{\doi}{doi: \begingroup \urlstyle{rm}\Url}\fi

\bibitem[Boys et~al.(2023)Boys, Girolami, Pidstrigach, Reich, Mosca, and
  Akyildiz]{boys2023tweedie}
Boys, B., Girolami, M., Pidstrigach, J., Reich, S., Mosca, A., and Akyildiz,
  O.~D.
\newblock Tweedie moment projected diffusions for inverse problems.
\newblock \emph{arXiv preprint arXiv:2310.06721}, 2023.

\bibitem[Cardoso et~al.(2024)Cardoso, el~idrissi, Corff, and
  Moulines]{cardoso2024monte}
Cardoso, G., el~idrissi, Y.~J., Corff, S.~L., and Moulines, E.
\newblock Monte carlo guided denoising diffusion models for bayesian linear
  inverse problems.
\newblock In \emph{The Twelfth International Conference on Learning
  Representations}, 2024.
\newblock URL \url{https://openreview.net/forum?id=nHESwXvxWK}.

\bibitem[Casella \& George(1992)Casella and George]{casella1992explaining}
Casella, G. and George, E.~I.
\newblock Explaining the gibbs sampler.
\newblock \emph{The American Statistician}, 46\penalty0 (3):\penalty0 167--174,
  1992.

\bibitem[Choi et~al.(2021)Choi, Kim, Jeong, Gwon, and Yoon]{choi2021ilvr}
Choi, J., Kim, S., Jeong, Y., Gwon, Y., and Yoon, S.
\newblock Ilvr: Conditioning method for denoising diffusion probabilistic
  models. in 2021 ieee.
\newblock In \emph{CVF international conference on computer vision (ICCV)},
  volume~1, pp.\ ~2, 2021.

\bibitem[Chung et~al.(2023)Chung, Kim, Mccann, Klasky, and
  Ye]{chung2023diffusion}
Chung, H., Kim, J., Mccann, M.~T., Klasky, M.~L., and Ye, J.~C.
\newblock Diffusion posterior sampling for general noisy inverse problems.
\newblock In \emph{The Eleventh International Conference on Learning
  Representations}, 2023.
\newblock URL \url{https://openreview.net/forum?id=OnD9zGAGT0k}.

\bibitem[Corenflos et~al.(2024)Corenflos, Zhao, S{\"a}rkk{\"a}, Sj{\"o}lund,
  and Sch{\"o}n]{corenflos2024conditioning}
Corenflos, A., Zhao, Z., S{\"a}rkk{\"a}, S., Sj{\"o}lund, J., and Sch{\"o}n,
  T.~B.
\newblock Conditioning diffusion models by explicit forward-backward bridging.
\newblock \emph{arXiv preprint arXiv:2405.13794}, 2024.

\bibitem[Daras et~al.()Daras, Chung, Lai, Mitsufuji, Milanfar, Dimakis, Ye, and
  Delbracio]{daras2024survey}
Daras, G., Chung, H., Lai, C.-H., Mitsufuji, Y., Milanfar, P., Dimakis, A.~G.,
  Ye, C., and Delbracio, M.
\newblock A survey on diffusion models for inverse problems. 2024.
\newblock \emph{URL https://giannisdaras. github.
  io/publications/diffusion\_survey. pdf}.

\bibitem[D{\'e}fossez et~al.(2019)D{\'e}fossez, Usunier, Bottou, and
  Bach]{defossez2019music}
D{\'e}fossez, A., Usunier, N., Bottou, L., and Bach, F.
\newblock Music source separation in the waveform domain.
\newblock \emph{arXiv preprint arXiv:1911.13254}, 2019.

\bibitem[Dhariwal \& Nichol(2021)Dhariwal and Nichol]{dhariwal2021diffusion}
Dhariwal, P. and Nichol, A.
\newblock Diffusion models beat gans on image synthesis.
\newblock \emph{Advances in neural information processing systems},
  34:\penalty0 8780--8794, 2021.

\bibitem[Dong et~al.(2015)Dong, Loy, He, and Tang]{dong2015image}
Dong, C., Loy, C.~C., He, K., and Tang, X.
\newblock Image super-resolution using deep convolutional networks.
\newblock \emph{IEEE transactions on pattern analysis and machine
  intelligence}, 38\penalty0 (2):\penalty0 295--307, 2015.

\bibitem[Dou \& Song(2024)Dou and Song]{dou2024diffusion}
Dou, Z. and Song, Y.
\newblock Diffusion posterior sampling for linear inverse problem solving: A
  filtering perspective.
\newblock In \emph{The Twelfth International Conference on Learning
  Representations}, 2024.
\newblock URL \url{https://openreview.net/forum?id=tplXNcHZs1}.

\bibitem[Finzi et~al.(2023)Finzi, Boral, Wilson, Sha, and
  Zepeda-N{\'u}{\~n}ez]{finzi2023user}
Finzi, M.~A., Boral, A., Wilson, A.~G., Sha, F., and Zepeda-N{\'u}{\~n}ez, L.
\newblock User-defined event sampling and uncertainty quantification in
  diffusion models for physical dynamical systems.
\newblock In \emph{International Conference on Machine Learning}, pp.\
  10136--10152. PMLR, 2023.

\bibitem[Gelfand(2000)]{gelfand2000gibbs}
Gelfand, A.~E.
\newblock Gibbs sampling.
\newblock \emph{Journal of the American statistical Association}, 95\penalty0
  (452):\penalty0 1300--1304, 2000.

\bibitem[Geman \& Geman(1984)Geman and Geman]{geman1984stochastic}
Geman, S. and Geman, D.
\newblock Stochastic relaxation, gibbs distributions, and the bayesian
  restoration of images.
\newblock \emph{IEEE Transactions on pattern analysis and machine
  intelligence}, \penalty0 (6):\penalty0 721--741, 1984.

\bibitem[Ho et~al.(2020)Ho, Jain, and Abbeel]{ho2020denoising}
Ho, J., Jain, A., and Abbeel, P.
\newblock Denoising diffusion probabilistic models.
\newblock \emph{Advances in Neural Information Processing Systems},
  33:\penalty0 6840--6851, 2020.

\bibitem[Ho et~al.(2022)Ho, Salimans, Gritsenko, Chan, Norouzi, and
  Fleet]{ho2022video}
Ho, J., Salimans, T., Gritsenko, A., Chan, W., Norouzi, M., and Fleet, D.~J.
\newblock Video diffusion models.
\newblock \emph{Advances in Neural Information Processing Systems},
  35:\penalty0 8633--8646, 2022.

\bibitem[Isola et~al.(2017)Isola, Zhu, Zhou, and Efros]{isola2017image}
Isola, P., Zhu, J.-Y., Zhou, T., and Efros, A.~A.
\newblock Image-to-image translation with conditional adversarial networks.
\newblock In \emph{Proceedings of the IEEE conference on computer vision and
  pattern recognition}, pp.\  1125--1134, 2017.

\bibitem[Jiang et~al.(2023)Jiang, Cornman, Park, Sapp, Zhou, Anguelov,
  et~al.]{jiang2023motiondiffuser}
Jiang, C., Cornman, A., Park, C., Sapp, B., Zhou, Y., Anguelov, D., et~al.
\newblock Motiondiffuser: Controllable multi-agent motion prediction using
  diffusion.
\newblock In \emph{Proceedings of the IEEE/CVF Conference on Computer Vision
  and Pattern Recognition}, pp.\  9644--9653, 2023.

\bibitem[Kadkhodaie \& Simoncelli(2020)Kadkhodaie and
  Simoncelli]{kadkhodaie2020solving}
Kadkhodaie, Z. and Simoncelli, E.~P.
\newblock Solving linear inverse problems using the prior implicit in a
  denoiser.
\newblock \emph{arXiv preprint arXiv:2007.13640}, 2020.

\bibitem[Kawar et~al.(2021)Kawar, Vaksman, and Elad]{kawar2021snips}
Kawar, B., Vaksman, G., and Elad, M.
\newblock Snips: Solving noisy inverse problems stochastically.
\newblock \emph{Advances in Neural Information Processing Systems},
  34:\penalty0 21757--21769, 2021.

\bibitem[Kawar et~al.(2022)Kawar, Elad, Ermon, and Song]{kawar2022denoising}
Kawar, B., Elad, M., Ermon, S., and Song, J.
\newblock Denoising diffusion restoration models.
\newblock \emph{Advances in Neural Information Processing Systems},
  35:\penalty0 23593--23606, 2022.

\bibitem[Kingma \& Welling(2013)Kingma and Welling]{kingma2013auto}
Kingma, D.~P. and Welling, M.
\newblock Auto-encoding variational bayes.
\newblock \emph{arXiv preprint arXiv:1312.6114}, 2013.

\bibitem[Ledig et~al.(2017)Ledig, Theis, Husz{\'a}r, Caballero, Cunningham,
  Acosta, Aitken, Tejani, Totz, Wang, et~al.]{ledig2017photo}
Ledig, C., Theis, L., Husz{\'a}r, F., Caballero, J., Cunningham, A., Acosta,
  A., Aitken, A., Tejani, A., Totz, J., Wang, Z., et~al.
\newblock Photo-realistic single image super-resolution using a generative
  adversarial network.
\newblock In \emph{Proceedings of the IEEE conference on computer vision and
  pattern recognition}, pp.\  4681--4690, 2017.

\bibitem[Li et~al.(2024)Li, Zhao, Wang, Scalia, Eraslan, Nair, Biancalani, Ji,
  Regev, Levine, et~al.]{li2024derivative}
Li, X., Zhao, Y., Wang, C., Scalia, G., Eraslan, G., Nair, S., Biancalani, T.,
  Ji, S., Regev, A., Levine, S., et~al.
\newblock Derivative-free guidance in continuous and discrete diffusion models
  with soft value-based decoding.
\newblock \emph{arXiv preprint arXiv:2408.08252}, 2024.

\bibitem[Lugmayr et~al.(2022)Lugmayr, Danelljan, Romero, Yu, Timofte, and
  Van~Gool]{lugmayr2022repaint}
Lugmayr, A., Danelljan, M., Romero, A., Yu, F., Timofte, R., and Van~Gool, L.
\newblock Repaint: Inpainting using denoising diffusion probabilistic models.
\newblock In \emph{Proceedings of the IEEE/CVF Conference on Computer Vision
  and Pattern Recognition}, pp.\  11461--11471, 2022.

\bibitem[Manilow et~al.(2019)Manilow, Wichern, Seetharaman, and
  Le~Roux]{Manilow2019slakh}
Manilow, E., Wichern, G., Seetharaman, P., and Le~Roux, J.
\newblock Cutting music source separation some slakh: A dataset to study the
  impact of training data quality and quantity.
\newblock In \emph{2019 IEEE Workshop on Applications of Signal Processing to
  Audio and Acoustics (WASPAA)}, pp.\  45--49, 2019.
\newblock \doi{10.1109/WASPAA.2019.8937170}.

\bibitem[Manilow et~al.(2022)Manilow, Hawthorne, Huang, Pardo, and
  Engel]{manilow2022improving}
Manilow, E., Hawthorne, C., Huang, C.-Z.~A., Pardo, B., and Engel, J.
\newblock Improving source separation by explicitly modeling dependencies
  between sources.
\newblock In \emph{ICASSP 2022-2022 IEEE International Conference on Acoustics,
  Speech and Signal Processing (ICASSP)}, pp.\  291--295. IEEE, 2022.

\bibitem[Mardani et~al.(2024)Mardani, Song, Kautz, and Vahdat]{mardani2024a}
Mardani, M., Song, J., Kautz, J., and Vahdat, A.
\newblock A variational perspective on solving inverse problems with diffusion
  models.
\newblock In \emph{The Twelfth International Conference on Learning
  Representations}, 2024.
\newblock URL \url{https://openreview.net/forum?id=1YO4EE3SPB}.

\bibitem[Mariani et~al.(2023)Mariani, Tallini, Postolache, Mancusi, Cosmo, and
  Rodol{\`a}]{mariani2023multi}
Mariani, G., Tallini, I., Postolache, E., Mancusi, M., Cosmo, L., and
  Rodol{\`a}, E.
\newblock Multi-source diffusion models for simultaneous music generation and
  separation.
\newblock \emph{arXiv preprint arXiv:2302.02257}, 2023.

\bibitem[Meng et~al.(2021)Meng, Song, Li, and Ermon]{meng2021estimating}
Meng, C., Song, Y., Li, W., and Ermon, S.
\newblock Estimating high order gradients of the data distribution by
  denoising.
\newblock \emph{Advances in Neural Information Processing Systems},
  34:\penalty0 25359--25369, 2021.

\bibitem[Moufad et~al.(2024)Moufad, Janati, Bedin, Durmus, Douc, Moulines, and
  Olsson]{moufad2024variational}
Moufad, B., Janati, Y., Bedin, L., Durmus, A., Douc, R., Moulines, E., and
  Olsson, J.
\newblock Variational diffusion posterior sampling with midpoint guidance.
\newblock \emph{arXiv preprint arXiv:2410.09945}, 2024.

\bibitem[Robbins(1956)]{robbins1956empirical}
Robbins, H.~E.
\newblock An empirical bayes approach to statistics.
\newblock In \emph{Proceedings of the Third Berkeley Symposium on Mathematical
  Statistics and Probability, Volume 1: Contributions to the Theory of
  Statistics}, 1956.
\newblock URL \url{https://api.semanticscholar.org/CorpusID:26161481}.

\bibitem[Roberts \& Smith(1994)Roberts and Smith]{roberts1994simple}
Roberts, G.~O. and Smith, A.~F.
\newblock Simple conditions for the convergence of the gibbs sampler and
  metropolis-hastings algorithms.
\newblock \emph{Stochastic processes and their applications}, 49\penalty0
  (2):\penalty0 207--216, 1994.

\bibitem[Rombach et~al.(2022)Rombach, Blattmann, Lorenz, Esser, and
  Ommer]{rombach2022high}
Rombach, R., Blattmann, A., Lorenz, D., Esser, P., and Ommer, B.
\newblock High-resolution image synthesis with latent diffusion models.
\newblock In \emph{Proceedings of the IEEE/CVF Conference on Computer Vision
  and Pattern Recognition}, pp.\  10684--10695, 2022.

\bibitem[Rout et~al.(2024)Rout, Raoof, Daras, Caramanis, Dimakis, and
  Shakkottai]{rout2024solving}
Rout, L., Raoof, N., Daras, G., Caramanis, C., Dimakis, A., and Shakkottai, S.
\newblock Solving linear inverse problems provably via posterior sampling with
  latent diffusion models.
\newblock \emph{Advances in Neural Information Processing Systems}, 36, 2024.

\bibitem[Roux et~al.(2019)Roux, Wisdom, Erdogan, and Hershey]{le2019sdr}
Roux, J.~L., Wisdom, S., Erdogan, H., and Hershey, J.~R.
\newblock Sdr – half-baked or well done?
\newblock In \emph{ICASSP 2019 - 2019 IEEE International Conference on
  Acoustics, Speech and Signal Processing (ICASSP)}, pp.\  626--630, 2019.
\newblock \doi{10.1109/ICASSP.2019.8683855}.

\bibitem[Rozet \& Louppe(2023)Rozet and Louppe]{rozet2023score}
Rozet, F. and Louppe, G.
\newblock Score-based data assimilation.
\newblock \emph{Advances in Neural Information Processing Systems},
  36:\penalty0 40521--40541, 2023.

\bibitem[Rozet et~al.(2024)Rozet, Andry, Lanusse, and
  Louppe]{rozet2024learning}
Rozet, F., Andry, G., Lanusse, F., and Louppe, G.
\newblock Learning diffusion priors from observations by expectation
  maximization.
\newblock In \emph{The Thirty-eighth Annual Conference on Neural Information
  Processing Systems}, 2024.
\newblock URL \url{https://openreview.net/forum?id=7v88Fh6iSM}.

\bibitem[Schneider et~al.(2023)Schneider, Kamal, Jin, and
  Sch{\"o}lkopf]{schneider2023musai}
Schneider, F., Kamal, O., Jin, Z., and Sch{\"o}lkopf, B.
\newblock Musai: text-to-music generation with long-context latent diffusion.
  arxiv preprint.
\newblock \emph{arXiv preprint arXiv:2301.11757}, 2023.

\bibitem[Shin \& Song(2017)Shin and Song]{shin2017jpeg}
Shin, R. and Song, D.
\newblock Jpeg-resistant adversarial images.
\newblock In \emph{NIPS 2017 workshop on machine learning and computer
  security}, volume~1, pp.\ ~8, 2017.

\bibitem[Sohl-Dickstein et~al.(2015)Sohl-Dickstein, Weiss, Maheswaranathan, and
  Ganguli]{sohl2015deep}
Sohl-Dickstein, J., Weiss, E., Maheswaranathan, N., and Ganguli, S.
\newblock Deep unsupervised learning using nonequilibrium thermodynamics.
\newblock In \emph{International Conference on Machine Learning}, pp.\
  2256--2265. PMLR, 2015.

\bibitem[Song et~al.(2024)Song, Kwon, Zhang, Hu, Qu, and Shen]{song2024solving}
Song, B., Kwon, S.~M., Zhang, Z., Hu, X., Qu, Q., and Shen, L.
\newblock Solving inverse problems with latent diffusion models via hard data
  consistency.
\newblock In \emph{The Twelfth International Conference on Learning
  Representations}, 2024.
\newblock URL \url{https://openreview.net/forum?id=j8hdRqOUhN}.

\bibitem[Song et~al.(2023{\natexlab{a}})Song, Vahdat, Mardani, and
  Kautz]{song2022pseudoinverse}
Song, J., Vahdat, A., Mardani, M., and Kautz, J.
\newblock Pseudoinverse-guided diffusion models for inverse problems.
\newblock In \emph{International Conference on Learning Representations},
  2023{\natexlab{a}}.
\newblock URL \url{https://openreview.net/forum?id=9_gsMA8MRKQ}.

\bibitem[Song et~al.(2023{\natexlab{b}})Song, Zhang, Yin, Mardani, Liu, Kautz,
  Chen, and Vahdat]{song2023loss}
Song, J., Zhang, Q., Yin, H., Mardani, M., Liu, M.-Y., Kautz, J., Chen, Y., and
  Vahdat, A.
\newblock Loss-guided diffusion models for plug-and-play controllable
  generation.
\newblock In \emph{International Conference on Machine Learning}, pp.\
  32483--32498. PMLR, 2023{\natexlab{b}}.

\bibitem[Song \& Ermon(2019)Song and Ermon]{song2019generative}
Song, Y. and Ermon, S.
\newblock Generative modeling by estimating gradients of the data distribution.
\newblock \emph{Advances in neural information processing systems}, 32, 2019.

\bibitem[Song et~al.(2021)Song, Sohl-Dickstein, Kingma, Kumar, Ermon, and
  Poole]{song2021score}
Song, Y., Sohl-Dickstein, J., Kingma, D.~P., Kumar, A., Ermon, S., and Poole,
  B.
\newblock Score-based generative modeling through stochastic differential
  equations.
\newblock In \emph{International Conference on Learning Representations}, 2021.

\bibitem[Stevens et~al.(2023)Stevens, van Gorp, Meral, Shin, Yu, Robert, and
  van Sloun]{stevens2023removing}
Stevens, T.~S., van Gorp, H., Meral, F.~C., Shin, J., Yu, J., Robert, J.-L.,
  and van Sloun, R.~J.
\newblock Removing structured noise with diffusion models.
\newblock \emph{arXiv preprint arXiv:2302.05290}, 2023.

\bibitem[Tran et~al.(2021)Tran, Tran, Phung, and
  Hoai]{tran2021non-uniform-deblurring}
Tran, P., Tran, A.~T., Phung, Q., and Hoai, M.
\newblock Explore image deblurring via encoded blur kernel space.
\newblock In \emph{Proceedings of the IEEE/CVF conference on computer vision
  and pattern recognition}, pp.\  11956--11965, 2021.

\bibitem[Trippe et~al.(2023)Trippe, Yim, Tischer, Baker, Broderick, Barzilay,
  and Jaakkola]{trippe2023diffusion}
Trippe, B.~L., Yim, J., Tischer, D., Baker, D., Broderick, T., Barzilay, R.,
  and Jaakkola, T.~S.
\newblock Diffusion probabilistic modeling of protein backbones in 3d for the
  motif-scaffolding problem.
\newblock In \emph{The Eleventh International Conference on Learning
  Representations}, 2023.
\newblock URL \url{https://openreview.net/forum?id=6TxBxqNME1Y}.

\bibitem[Wang et~al.(2023)Wang, Yu, and Zhang]{wang2023zeroshot}
Wang, Y., Yu, J., and Zhang, J.
\newblock Zero-shot image restoration using denoising diffusion null-space
  model.
\newblock In \emph{The Eleventh International Conference on Learning
  Representations}, 2023.
\newblock URL \url{https://openreview.net/forum?id=mRieQgMtNTQ}.

\bibitem[Wu et~al.(2023)Wu, Trippe, Naesseth, Cunningham, and
  Blei]{wu2023practical}
Wu, L., Trippe, B.~L., Naesseth, C.~A., Cunningham, J.~P., and Blei, D.
\newblock Practical and asymptotically exact conditional sampling in diffusion
  models.
\newblock In \emph{Thirty-seventh Conference on Neural Information Processing
  Systems}, 2023.
\newblock URL \url{https://openreview.net/forum?id=eWKqr1zcRv}.

\bibitem[Wu et~al.(2024{\natexlab{a}})Wu, Sun, Chen, Zhang, Yue, and
  Bouman]{wu2024pnpdm}
Wu, Z., Sun, Y., Chen, Y., Zhang, B., Yue, Y., and Bouman, K.~L.
\newblock Principled probabilistic imaging using diffusion models as
  plug-and-play priors.
\newblock \emph{arXiv preprint arXiv:2405.18782}, 2024{\natexlab{a}}.

\bibitem[Wu et~al.(2024{\natexlab{b}})Wu, Sun, Chen, Zhang, Yue, and
  Bouman]{wu2024principled}
Wu, Z., Sun, Y., Chen, Y., Zhang, B., Yue, Y., and Bouman, K.~L.
\newblock Principled probabilistic imaging using diffusion models as
  plug-and-play priors.
\newblock \emph{arXiv preprint arXiv:2405.18782}, 2024{\natexlab{b}}.

\bibitem[Xia et~al.(2022)Xia, Zhang, Yang, Xue, Zhou, and Yang]{xia2022gan}
Xia, W., Zhang, Y., Yang, Y., Xue, J.-H., Zhou, B., and Yang, M.-H.
\newblock Gan inversion: A survey.
\newblock \emph{IEEE transactions on pattern analysis and machine
  intelligence}, 45\penalty0 (3):\penalty0 3121--3138, 2022.

\bibitem[Xu \& Chi(2024)Xu and Chi]{xu2024provably}
Xu, X. and Chi, Y.
\newblock Provably robust score-based diffusion posterior sampling for
  plug-and-play image reconstruction.
\newblock In \emph{The Thirty-eighth Annual Conference on Neural Information
  Processing Systems}, 2024.
\newblock URL \url{https://openreview.net/forum?id=SLnsoaY4u1}.

\bibitem[Yu et~al.(2023)Yu, Wang, Zhao, Ghanem, and Zhang]{yu2023freedom}
Yu, J., Wang, Y., Zhao, C., Ghanem, B., and Zhang, J.
\newblock Freedom: Training-free energy-guided conditional diffusion model.
\newblock In \emph{Proceedings of the IEEE/CVF International Conference on
  Computer Vision}, pp.\  23174--23184, 2023.

\bibitem[Zhang et~al.(2024{\natexlab{a}})Zhang, Chu, Berner, Meng, Anandkumar,
  and Song]{zhang2024daps}
Zhang, B., Chu, W., Berner, J., Meng, C., Anandkumar, A., and Song, Y.
\newblock Improving diffusion inverse problem solving with decoupled noise
  annealing.
\newblock \emph{arXiv preprint arXiv:2407.01521}, 2024{\natexlab{a}}.

\bibitem[Zhang et~al.(2024{\natexlab{b}})Zhang, Chu, Berner, Meng, Anandkumar,
  and Song]{zhang2024improving}
Zhang, B., Chu, W., Berner, J., Meng, C., Anandkumar, A., and Song, Y.
\newblock Improving diffusion inverse problem solving with decoupled noise
  annealing.
\newblock \emph{arXiv preprint arXiv:2407.01521}, 2024{\natexlab{b}}.

\bibitem[Zhang et~al.(2018)Zhang, Isola, Efros, Shechtman, and
  Wang]{zhang2018lpips}
Zhang, R., Isola, P., Efros, A.~A., Shechtman, E., and Wang, O.
\newblock The unreasonable effectiveness of deep features as a perceptual
  metric.
\newblock In \emph{Proceedings of the IEEE conference on computer vision and
  pattern recognition}, pp.\  586--595, 2018.

\bibitem[Zhu et~al.(2023)Zhu, Zhang, Liang, Cao, Wen, Timofte, and
  Van~Gool]{zhu2023denoising}
Zhu, Y., Zhang, K., Liang, J., Cao, J., Wen, B., Timofte, R., and Van~Gool, L.
\newblock Denoising diffusion models for plug-and-play image restoration.
\newblock In \emph{Proceedings of the IEEE/CVF Conference on Computer Vision
  and Pattern Recognition}, pp.\  1219--1229, 2023.

\end{thebibliography}
\bibliographystyle{icml2025}

\newpage
\appendix
\onecolumn
\section{Methodology details}
\subsection{Primer on Gibbs sampling}
\label{apdx-sec:gibbs}
In this section we lay out the basic properties of Gibbs sampling. We use measure-theoretic notation for conciseness. 

Let $\probmeas{0,1}{}{\rmd (\bx_0, \bx_1)}$ be a probability measure on $\rset^\dimx \times \rset^\dimx$. We denote by $\probmeas{0|1}{\bx_1}{\rmd \bx_0}$ and $\probmeas{1|0}{\bx_0}{\rmd \bx_1}$ the associated full conditionals and we write $\probmeas{0}{}{}$, $\probmeas{1}{}{}$ for its marginals. Define the 
transition kernels 
\begin{align*} 
    P_0(\rmd (\bx^\prime _0, \bx^\prime _1) | \bx_0, \bx_1) & \eqdef \probmeas{0|1}{\bx_1}{\rmd \bx^\prime _0} \delta_{\bx_1}(\rmd \bx^\prime _1),\\
    P_1(\rmd (\bx^\prime _0, \bx^\prime _1) | \bx_0, \bx_1) & \eqdef \probmeas{1|0}{\bx_0}{\rmd \bx^\prime _1} \delta_{\bx_0}(\rmd \bx^\prime _0).
\end{align*}
Each transition kernel updates only one coordinate at a time. A full update of the coordinates is obtained by composition of the kernels, \emph{i.e.}
$$ 
    P_0 P_1(\rmd (\bx^\prime _0, \bx^\prime _1) | \bx_0, \bx_1) \eqdef \int P_1(\rmd (\bx^\prime _0, \bx^\prime _1) | \tilde\bx_0, \tilde\bx_1) P_0(\rmd (\tilde\bx _0, \tilde\bx _1) | \bx_0, \bx_1) \eqsp.
$$  
 Each transition admits the joint distribution $\probmeas{0, 1}{}{}$ as stationary distribution, meaning that $\probmeas{0, 1}{}{\rmd (\bx_0, \bx_1)} = \int P_0(\rmd (\bx _0, \bx _1) | \bx^\prime _0, \bx^\prime _1)\, \probmeas{0, 1}{}{\rmd (\bx^\prime _0, \bx^\prime _1)}$. Indeed, this is seen by noting that 
\begin{align*}
    P_0(\rmd (\bx _0, \bx _1) | \bx^\prime _0, \bx^\prime _1) \probmeas{0, 1}{}{\rmd (\bx^\prime _0, \bx^\prime _1)}  & =  \probmeas{0|1}{\bx^\prime _1}{\rmd \bx _0} \delta_{\bx^\prime _1}(\rmd \bx _1) \probmeas{0, 1}{}{\rmd (\bx^\prime _0, \bx^\prime _1)}\\
    & =   \probmeas{0|1}{\bx^\prime _1}{\rmd \bx _0} \delta_{\bx^\prime _1}(\rmd \bx _1) \probmeas{0|1}{\bx^\prime _1}{\rmd \bx^\prime _0} \probmeas{1}{}{\rmd \bx^\prime _1}\\ 
    & = \probmeas{0|1}{\bx _1}{\rmd \bx _0} \probmeas{1}{}{\rmd \bx _1}  \probmeas{0|1}{\bx^\prime _1}{\rmd \bx^\prime _0}\delta_{\bx _1}(\rmd \bx^\prime _1),
\end{align*}
and then integrating both sides \wrt\ $(\bx^\prime _0, \bx^\prime _1)$. It then follows immediately that also $P_0 P_1$ admits $\probmeas{0,1}{}{}$ as stationary distribution. Letting $\big( (X^k _0, X^k _1) \big)_{k \in \nset}$ be a Markov chain with transition kernel $P_0 P_1$, the law of $(X^k _0, X^k _1)$ converges to $\probmeas{0,1}{}{}$ as $k \to \infty$ under mild conditions; see \cite{roberts1994simple}. 
\subsection{Full Gibbs conditionals}
\label{apdx-sec:conditionals}
In the main paper we consider the following data augmentation of the mixture $\hpost{t}{}{}$ \eqref{eq:posterior-approximation}
\begin{equation}
    \label{eq:extended-distr-normalized}
    \epost{0, s, t}{}{\bx_0, \bx_s, \bx_t} \\ = \pdata{0|s}{\bx_s}{\bx_0} \frac{\hpot{s}{\bx_s} \pdata{s|t}{\bx_t}{\bx_s} \pdata{t}{}{\bx_t}}{\int  \hpot{s}{\bx^\prime _s} \pdata{s|t}{\bx^\prime _t}{\bx^\prime _s} \pdata{t}{}{\bx^\prime _t} \, \rmd \bx_{s, t}}  \eqsp.
\end{equation}
From this definition it is straightforward to see that $\epost{0|s, t}{\bx_s, \bx_t}{\bx_0} = \pdata{0|s}{\bx_s}{\bx_0}$. In order to compute the full conditional $\epost{s|0, t}{\bx_0, \bx_t}{\bx_s}$ we use the identity 
\begin{equation} 
    \label{eq:bw-fw}
    \pdata{0|s}{\bx_s}{\bx_0} \pdata{s|t}{\bx_t}{\bx_s} \pdata{t}{}{\bx_t}= \pdata{0}{}{\bx_0} \fw{s|0}{\bx_0}{\bx_s} \fw{t|s}{\bx_s}{\bx_t},
\end{equation} 
from which it follows that 
\begin{align*} 
    \epost{s|0, t}{\bx_0, \bx_t}{\bx_s} & = \frac{\pdata{0|s}{\bx_s}{\bx_0} \hpot{s}{\bx_s} \pdata{s|t}{\bx_t}{\bx_s}}{\int \pdata{0|s}{\bx^\prime _s}{\bx_0} \hpot{s}{\bx^\prime _s} \pdata{s|t}{\bx_t}{\bx^\prime _s}\, \rmd \bx^\prime _s } \\
    & = \frac{ \fw{s|0}{\bx_0}{\bx_s} \hpot{s}{\bx_s} \fw{t|s}{\bx_s}{\bx_t}}{\int \fw{s|0}{\bx _0}{\bx^\prime _s} \hpot{s}{\bx^\prime _s} \fw{t|s}{\bx^\prime _s}{\bx _t}\, \rmd \bx^\prime _s } \\
    & = \frac{ \fw{s|0}{\bx_0}{\bx_s} \hpot{s}{\bx_s} \fw{t|s}{\bx_s}{\bx_t} \big/ \fw{t|0}{\bx_0}{\bx_t}}{\int \fw{s|0}{\bx _0}{\bx^\prime _s} \hpot{s}{\bx^\prime _s} \fw{t|s}{\bx^\prime _s}{\bx _t} \big/ \fw{t|0}{\bx_0}{\bx_t} \, \rmd \bx^\prime _s } \eqsp.
\end{align*}
Then, by noting that the bridge transition \eqref{eq:bridge} satisfies $\fw{s|0, t}{\bx_0, \bx_t}{\bx_s} = \fw{s|0}{\bx_0}{\bx_s} \fw{t|s}{\bx_s}{\bx_t} / \fw{t|0}{\bx_0}{\bx_t}$, we find that 
$$ 
\epost{s|0, t}{\bx_0, \bx_t}{\bx_s} = \frac{\hpot{s}{\bx_s} \fw{s|0, t}{\bx_0, \bx_t}{\bx_s}}{\int \hpot{s}{\bx^\prime _s} \fw{s|0, t}{\bx_0, \bx_t}{\bx^\prime _s} \, \rmd \bx^\prime _s}
$$
Finally, for the third conditional, using again the identity \eqref{eq:bw-fw}, we find that 
\begin{align*} 
    \epost{t|0, s}{\bx_0, \bx_s}{\bx_t} & = \frac{\pdata{0|s}{\bx_s}{\bx_0} \hpot{s}{\bx_s} \pdata{s|t}{\bx_t}{\bx_s} \pdata{t}{}{\bx_t}}{\int \pdata{0|s}{\bx _s}{\bx_0} \hpot{s}{\bx _s} \pdata{s|t}{\bx^\prime _t}{\bx _s} \pdata{t}{}{\bx^\prime _t}\, \rmd \bx^\prime _t } \\
    & = \fw{t|s}{\bx_s}{\bx_t} \eqsp.
\end{align*}
\subsection{Variational approximation}
\label{apdx-sec:vi-approx}
In this section we describe the variational approach of \citet{moufad2024variational}, which we use to fit a Gaussian variational approximation to $\post{s|0, t}{\bx_0, \bx_t}{}$ for fixed $(\bx_0, \bx_t)$. Similarly to the main paper we consider the variational approximation 
\begin{equation} 
    \smash{\vi{s|0, t}{} \eqdef \gauss\big(\vmu_{s|0, t}, \diag(\rme^{\vlstd_{s|0,t}})\big)}, 
\end{equation}
and let $\vparam_{s|0, t} \eqdef (\vmu_{s|0, t}, \vlstd_{s|0,t}) \in \rset^{\dimx} \times \rset^{\dimx}$ denote the variational parameters. 
The reverse KL divergence writes, following definition \eqref{eq:bridge}, 
\begin{align} 
    \lefteqn{\kldivergence{\vi{s|0, t}{}}{\post{s|0, t}{\bx_0, \bx_t}{}}} \nonumber \\
    & \hspace{.7cm} = \int \log \frac{\vi{s|0, t}{\bx_s}}{\hpot{s}{\bx_s}{} \fw{s|0, t}{\bx_0, \bx_t}{\bx_s}} \, \vi{s|0, t}{\bx_s} \, \rmd \bx_s + \mathrm{C} \nonumber \\
    & \hspace{.7cm} = \pE _{\vi{s|0, t}{}} \left[ - \log \hpot{s}{\vX^\vparam _s} + \frac{\| \vX^\vparam _s - \big( \gamma_{t|s} \a_{s|0} \bx_0 + (1 - \gamma_{t|s}) \a^{-1} _{t|s} \bx_t \big) \|^2}{2 \std^2 _{s|0, t}} \right] - \frac{1}{2} \vlstd_{s|0, t}^T \mathbf{1}_d + \mathrm{C}^\prime. \label{eq:gradient-estimator} 
\end{align}
Using the reparameterization trick \cite{kingma2013auto} and plugging-in the neural network approximation $\hpot{s}{}[\param]$ of $\hpot{s}{}$, we obtain the gradient estimator  
\begin{multline*}
    \nabla_\vparam \mathcal{L}^s _t(\vparam; \bx_0, \bx_t, Z) \eqdef - \nabla_\vparam \log \hpot{s}{\vmu_{s|0, t} + \diag(\rme^{\vlstd_{s|0, t} })^{1/2} Z}[\param] \\
     + \nabla_\vparam \bigg[ \frac{\|\vmu_{s|0, t} + \diag(\rme^{\vlstd_{s|0, t} })^{1/2} Z - \big( \gamma_{t|s} \a_{s|0} \bx_0 + (1 - \gamma_{t|s}) \a^{-1} _{t|s} \bx_t \big)\|^2 }{2 \std^2 _{s|0, t}} - \frac{1}{2} \vlstd_{s|0, t}^T \mathbf{1}_d \bigg], 
\end{multline*}
where $Z \sim \gauss(\zero_\dimx, \Id_\dimx)$. We initialize the variational parameters with the mean and covariance of the bridge kernel \eqref{eq:bridge},  \emph{i.e.}, at initialization, $\vmu^0 _{s|0, t} \eqdef \gamma_{t|s} \a_{s|0} \bx_0 + (1 - \gamma_{t|s}) \a^{-1} _{t|s} \bx_t$ and $\vlstd^0 _{s|0, t} = \log \std^2 _{s|0, t} \Id_\dimx$.
The $\vifn$ routine is summarized in \Cref{algo:gauss_vi}. 
\begin{algorithm}
    \caption{$\vifn$ routine}
    \begin{algorithmic}[1]
        \STATE {\bfseries Input:} vectors $(\bx_0, \bx_t)$, timesteps $(s, t)$, gradient steps $G$
        \STATE $\vmu \gets \gamma_{t|s} \a_{s|0} \bx_0 + (1 - \gamma_{t|s}) \a^{-1} _{t|s} \bx_t$
        \STATE $\vlstd \gets \log \std^2 _{s|0, t}$

        \FOR{$g=1$ to $G$}
            \STATE $Z \sim \gauss(\zero_\dimx, \Id_\dimx)$
            \STATE $(\vmu, \vlstd) \gets \mathsf{OptimizerStep}(\nabla _\vparam \mathcal{L}^s _t(\cdot, \bx_0, \bx_t, Z))$
        \ENDFOR
        \STATE $Z \sim \gauss(\zero_\dimx, \Id_\dimx)$
        \STATE {\bfseries Output:} $\vmu + \diag(\rme^{\vlstd / 2}) Z$
    \end{algorithmic}
    \label{algo:gauss_vi}
\end{algorithm}
\begin{remark} 
    While the expectation of the squared norm in \eqref{eq:gradient-estimator} can be computed exactly, we found that, in practice, doing so degraded the algorithm’s performance, producing blurrier images compared to simply using a Monte Carlo estimator for the full expectation.
\end{remark} 
\begin{remark} 
    \label{rem:metropolis}
    The fact that the density of our target distribution can be computed approximately by plugging the denoiser approximation allows us to add a Metropolis--Hastings (MH) correction with approximate acceptance ratio. Indeed, once we fit the Gaussian approximation, we can improve the accuracy of our sampler by simulating a Markov chain $(\vX^k _s)_k$ where, given $\vX^k _s$, 
    $$ 
    \vX^{k+1} _s \sim M_s(\rmd \bx_s | \vX^k _s) \eqdef \int \vi{s|0,t}{z} \bigg[ r_s(\vX^k _s, z) \delta_{z}(\rmd \bx_s) + (1 - r_s(\vX^k _s, z) ) \delta_{\vX^k _s}(\rmd \bx_s)\bigg]\, \rmd z \eqsp,
    $$ 
    with
    $$ 
        r_s(\bx_s, \bx^* _s) = \mbox{min}\left(1, \frac{\hpot{s}{\bx^* _s} \fw{s|0, t}{\bx_0, \bx_t}{\bx^* _s} \vi{s|0,t}{\bx_s}}{\hpot{s}{\bx _s} \fw{s|0, t}{\bx_0, \bx_t}{\bx _s} \vi{s|0,t}{\bx^* _s}} \right) \eqsp.
    $$ 
\end{remark}
\subsection{Alternative data augmentation and sequence}
\label{apdx-sec:data-aug}
\paragraph{Data augmentation.} Our algorithm is based on one data-augmentation approach, but alternative augmentations could also be considered. 
Let $s \in \intset{1}{t-1}$. Then the most obvious and natural data augmentation involves simply marginalizing out the $\bx_0$ variable in \eqref{eq:extended-distr-normalized}, yielding 
$$ 
    \epost{s, t}{}{\bx_s, \bx_t} \propto \hpot{s}{\bx_s} \pdata{s|t}{\bx_t}{\bx_s} \pdata{t}{}{\bx_t} \eqsp.
$$ 
Its full conditionals are $\epost{s|t}{\bx_t}{\bx_s} \propto \hpot{s}{\bx_s} \pdata{s|t}{\bx_t}{\bx_s}$ and $\epost{t|s}{\bx_s}{\bx_t} = \fw{t|s}{\bx_s}{\bx_t}$. The first conditional is intractable for sampling, and we could approximate it with a Gaussian variational distribution, similar to our approach for $\epost{s|0, t}{\bx_0, \bx_t}{}$. Indeed, this is possible since $\nabla_{\bx_s} \log \epost{s|t}{\bx_t}{\bx_s} = \nabla_{\bx_s} \log \hpot{s}{\bx_s} + \nabla_{\bx_s} \log \pdata{s}{}{\bx_s} + \nabla_{\bx_s} \log \fw{t|s}{\bx_s}{\bx_t}$, which can then be approximated using the parametric approximations $\nabla \log \hpot{s}{\bx_s}[\param]$ and $\nabla \log \pdata{s}{}{\bx_s} \approx (- \bx_s + \a_s \denoiser{s}{}{\bx_s}[\param]) / (1 - \a^2 _s)$. 

The first drawback of this approach is that, in practice, it tends to degrade reconstruction quality---\emph{e.g.}, introducing blurriness---as $t$ tends to $0$, due to the poor approximation of the score near the data distribution. Additionally, beyond the loss of quality, we observe that it produces more incoherent reconstructions with noticeable artifacts. We hypothesize that this issue arises because the distribution we aim to approximately sample involves the prior transition $\pdata{s|t}{}{}$,  which can be highly multi-modal when $s \ll t$. This multi-modality may make the posterior $\epost{s|t}{\bx_t}{}$ more challenging to approximately sample from. On the other hand, when further conditioning on $\bx_0$, the sampling problem becomes more well-behaved, as we then target the posterior of a Gaussian distribution. Finally, while the score of $\epost{\smash{s|t}}{\bx_t}{\bx_s}$ can be easily approximated, its density cannot, preventing the use of a Metropolis--Hastings correction, unless we use the independent proposal $\pdata{s|t}{\bx_t}{}$. However, this approach is suboptimal, as it does not incorporate any information from the observation. This is not the case of the data-augmentation approach we use in \algo\ as we highlight in \Cref{rem:metropolis}. 
\paragraph{Alternative sequence.} An alternative to the mixture of posterior approximations \eqref{eq:posterior-approximation}, on which \algo\ is based, is the posterior formed as a mixture of likelihoods: 
$$ 
    \hpost{t}{}{\bx_t} = \frac{ \sum_{s = 1}^{t-1} \wght^s _t \hpot{t}{\bx_t}[s] \pdata{t}{}{\bx_t} }{\int \sum_{s = 1}^{t-1} \wght^s _t \hpot{t}{\bx^\prime _t}[s] \pdata{t}{}{\bx^\prime _t} \, \rmd \bx^\prime _t} \eqsp, 
$$ 
being the $\bx_t$-marginal of the extended distribution 
\begin{equation}
    \label{eq:mixture-pot-extended}
    \epost{0, \smbs, t}{}{s, \bx_0, \bz, \bx_t} \propto  \wght^s _t \pdata{0|s}{\bz}{\bx_0} \hpot{s}{\bz} \pdata{s|t}{\bx_t}{\bz} \pdata{t}{}{\bx_t} \eqsp.
\end{equation}
Now, let $(s, \bXy_0, \bZy, \bXy_t) \sim \epost{0, \smbs, t}{}{}$; then, conditionally on $s$, the distribution of $(\bXy_0, \bZy, \bXy_t)$ is $\epost{0, s, t}{}{}$, whereas 
$$
s | \bXy_0, \bZy, \bXy_t \, \sim \mbox{Categorical}\left( \left\{ \frac{\wght^\ell _t \hpot{\ell}{\bZy} \fw{\ell|0, t}{\bXy_0, \bXy_t}{\bZy}}{\sum_{k = 1}^{t-1} \wght^k _t \hpot{k}{\bZy} \fw{k|0, t}{\bXy_0, \bXy_t}{\bZy}}\right\}_{\ell = 1} ^{t-1} \right) \eqsp.
$$

A Gibbs sampler targeting \eqref{eq:mixture-pot-extended} is described in \Cref{algo:mixturepot-extended-gibbs}. It allows updating the index $s$ in an observation-driven fashion, but is unfortunately computationally expensive as we need to evaluate the denoiser at $\bZy$ in parallel for $t-1$ timesteps. A cheaper alternative could be to block the variables $(s, \bZy)$ and use an independent MH step to target their joint conditional distribution. Denoting by $\lambda$ the joint proposal distribution on $\intset{1}{t-1} \times \rset^\dimx$ used in this independent MH step, the probability of accepting a candidate $(s^*, \bz^*)$ is 
$$ 
    r_t\big( (s, \bz), (s^*, \bz^*)\big) = \mbox{min}\left( 1, \frac{\wght^{s^*} _{t} \hpot{s^*}{\bz^*}\fw{s^* | 0, t}{\bx_0, \bx_t}{\bz^*} \lambda(s, \bz)}{\wght^{s} _{t} \hpot{s}{\bz}\fw{s | 0, t}{\bx_0, \bx_t}{\bz} \lambda(s^{*}, \bz^{*})} \right) \eqsp.
$$ 
\begin{algorithm}[h]
    \caption{Gibbs sampler targeting \eqref{eq:extended}}
    \begin{algorithmic}[1]
        \STATE {\bfseries Input:} $(s^r, \bXy^r _0, \bZy^r, \bXy^r _t)$
        \STATE draw $s^{r+1} \sim \mbox{Categorical}\left( \left\{ \frac{\wght^\ell _t \hpot{\ell}{\bZy^r} \fw{\ell|0, t}{\bXy^r _0, \bXy^r _t}{\bZy^r}}{\sum_{k = 1}^{t-1} \wght^k _t \hpot{k}{\bZy^r} \fw{k|0, t}{\bXy^r _0, \bXy^r _t}{\bZy^r}}\right\}_{k = 1} ^{t-1} \right)$ 
        \STATE draw $\bZy^{r+1} \sim \epost{s^{r+1} |0, t}{\bXy^r _0, \bXy^r _t}{}$
        \STATE draw $\bXy^{r+1} _t \sim \fw{t|s^{r+1}}{\bZy^{r+1}}{}$ 
        \STATE draw $\bXy^{r+1} _0 \sim \pdata{0|s^{r+1}}{\bZy^{r+1}}{}$
    \end{algorithmic}
    \label{algo:mixturepot-extended-gibbs}
\end{algorithm}
\begin{remark} 
    \label{rem:mgdm-weight}
    Note that we could have used a similar data augmentation \eqref{eq:mixture-pot-extended} for the mixture used in \algo. This would yield the full conditional 
$$
    s | \bXy_0, \bZy, \bXy_t \, \sim \mbox{Categorical}\left( \left\{ \frac{\wght^\ell _t \epost{\ell|0, t}{\bXy_0, \bXy_t}{\bZy}}{\sum_{k = 1}^{t-1} \wght^k _t \epost{\ell|0, t}{\bXy_0, \bXy_t}{\bZy}}\right\}_{k = 1} ^{t-1} \right) \eqsp, 
$$
which is, however, intractable due to the normalizing constant involved in each $\epost{\ell|0, t}{}{}$. 
\end{remark}
\subsection{Related algorithms}
\label{apdx-sec:comparisons}
\paragraph{Comparison with \citet{zhang2024daps}} In this section we clarify the difference between \algo\ and the \daps\ algorithm \cite{zhang2024daps}, which shares some similarities with our approach. The sampling procedure in \daps\ relies on sequential approximate sampling from the joint distribution 
$$ 
    \tilde\pi^\obs _{0:T}(\bx_{0:T}) \eqdef \post{T}{}{\bx_T} \prod_{t = 0}^{T-1} \tilde{\pi}_{t|t+1}(\bx_t | \bx_{t+1}),  
$$ 
where 
\begin{equation}
    \label{eq:daps-bw}
    \tilde\pi^\obs _{t|t+1}(\bx_t | \bx_{t+1}) \eqdef \int \fw{t|0}{\bx_0}{\bx_t} \post{0|t+1}{\bx_{t+1}}{\bx_0} \, \rmd \bx_0
\end{equation}
and $\post{0|t+1}{\bx_{t+1}}{\bx_0} = \post{0}{}{\bx_0} \fw{t+1|0}{\bx_0}{\bx_{t+1}} \big/ \post{t+1}{}{\bx_{t+1}}$. From this definition it follows that 
$$ 
 \post{t}{}{\bx_t} = \int \tilde\pi^\obs _{t|t+1}(\bx_t | \bx_{t+1}) \post{t+1}{}{\bx_{t+1}} \, \rmd \bx_{t+1} \eqsp, 
$$ 
and hence that the marginals of the joint distribution $\tilde\pi^\obs _{0:T}$ are $(\post{t}{}{})_{t = 0}^T$. The canonical backward transition $\post{t|t+1}{\bx_{t+1}}{\bx_t} \propto \post{t}{}{\bx_t} \fw{t+1|t}{\bx_t}{\bx_{t+1}}$ has the alternative form 
$$ 
    \post{t|t+1}{\bx_{t+1}}{\bx_t} = \int \fw{t|0, t+1}{\bx_0, \bx_{t+1}}{\bx_t} \post{0|t+1}{\bx_{t+1}}{\bx_t} \, \rmd \bx_0 \eqsp,
$$
which differs from \eqref{eq:daps-bw} in the use of the bridge transition $q_{t|0, t+1}$ instead of the forward transition $q_{t|0}$. 

In order to sample from $\tilde\pi_{t|t+1}(\cdot | \bx_{t+1})$, one needs to first sample $X_0 \sim \post{0|t+1}{\bx_{t+1}}{}$ and then $X_t \sim \fw{t|0}{X_0}{}$. \daps\ performs the former step using Langevin dynamics on an approximation of $\post{0|t+1}{\bx_{t+1}}{}$. More specifically, the authors use the approximation 
$$ 
\post{0|t+1}{\bx_{t+1}}{\bx_0} \approx \frac{\pot{0}{\bx_0} \normpdf(\bx_0; \denoiser{t+1}{}{\bx_{t+1}}, r^2 _{t+1} \Id_\dimx)}{\int \pot{0}{\bx^\prime _0} \normpdf(\bx^\prime _0; \denoiser{t+1}{}{\bx_{t+1}}, r^2 _{t+1} \Id_\dimx) \, \rmd \bx^\prime _0} \eqsp,
$$ 
where $r^2 _{t+1}$ is a hyperparameter. This approximation follows by noting that $\post{0|t+1}{\bx_{t+1}}{\bx_0} \propto \pot{0}{\bx_0} \pdata{0|t+1}{\bx_{t+1}}{\bx_0}$ and using the Gaussian approximation of $\pdata{0|t+1}{\bx_{t+1}}{}$ proposed by \citet{song2022pseudoinverse}. The Langevin step is initialized with a sample obtained by discretizing the probability flow ODE \cite{song2021score} between $t+1$ and $0$. 

Both \algo\ and \daps\ perform full noising and denoising steps and operate in a similar manner in this respect (with the distinction that we use DDPM instead of the probability flow ODE). The first fundamental difference is that we sample, conditionally on \(\obs\) and at a random timestep $s$, by drawing from \(\epost{s|0, t}{\bx_0, \bx_t}{} \propto \hpot{s}{\bx_s} \fw{s|0, t}{\bx_0, \bx_t}{\bx_s}\). Unlike \daps, our method does not rely on a density approximation prior to applying an approximate sampler. The second main difference is the fact that within each denoising step, we can increase the number of Gibbs iterations to improve the overall performance, as demonstrated in \Cref{fig:scaling}. This is on top of the number of gradient steps that we use to fit the variational approximation and which enhance the performance when we increase them.  

On the other hand, \daps\ does not require the computation of vector-Jacobian products of the denoiser and is thus more efficient in terms of memory. However it requires many calls to the likelihood function, which can substantially increase the runtime if it is expensive to evaluate. For example, with a latent diffusion model, the runtime of DAPS is at least three times larger than that of \algo, \resample, and \psld. 
\paragraph{Comparison with \citet{moufad2024variational}}
The more recent {\sc{MGPS}} algorithm of \citet{moufad2024variational} is also related to \algo. Similarly to DAPS \cite{zhang2024daps}, their methodology relies on sampling approximately from the posterior transition $\post{t|t+1}{\bx_{t+1}}{}$ at each step of the backward denoising process. It builds on the following decomposition, which holds for all $s \in \intset{0}{t-1}$:
$$ 
    \post{t|t+1}{\bx_{t+1}}{\bx_t} = \int \fw{t|s, t+1}{\bx_s, \bx_{t+1}}{\bx_t} \post{s|t+1}{\bx_{t+1}}{\bx_s} \, \rmd \bx_s \eqsp.
$$ 
One step of {\sc{MGPS}} proceeds by first sampling from an approximation of the posterior transition $\post{s|t+1}{\bx_{t+1}}{}$ and then sampling from the bridge transition to return back to time $t$. The approximation of the posterior transition used in the {\sc{MGPS}} is 
\begin{equation} 
    \label{eq:mgps-approx}
    \post{s|t+1}{\bx_{t+1}}{\bx_s} \approx \frac{\hpot{s}{\bx_s}[\param] \pdata{s|t+1}{\bx_{t+1}}{\bx_s}[\param]}{\int \hpot{s}{\bx^\prime _s} \pdata{s|t+1}{\bx_{t+1}}{\bx^\prime _s}[\param] \, \rmd \bx^\prime _s} \eqsp.
\end{equation}
Here one can then choose $s$ to be sufficiently small to enhance the likelihood approximation, while still having an accurate Gaussian approximation of the transition $\pdata{s|t+1}{\bx_{t+1}}{}$. The authors demonstrate, using a solvable toy example, that this trade-off indeed exists; see \citep[Example 3.2]{moufad2024variational}. The approximate sampling step is then performed by fitting a Gaussian variational approximation to the approximation on the \rhs\ of \eqref{eq:mgps-approx}, similarly to what we do in \Cref{algo:midpoint-gibbs}.  

Both \algo\ and {\sc{MGPS}} leverage the same idea of using, at step time $t$, likelihood approximations at earlier steps $s < t$. While in {\sc{MGPS}} the time $s$ is set deterministically as a function of $t$, we sample it randomly. However, the main difference lies in the step where we sample conditionally on the observation $\obs$. Once the index $s$ is sampled we proceed with $R$ rounds of reverse KL minimization \wrt\ to a \emph{different} target distribution. Indeed, following \Cref{algo:midpoint-gibbs}, in the first round we seek to fit a distribution with density proportional to $\bx_s \mapsto \hpot{s}{\bx_s}[\param] \fw{s|0, t}{\textcolor{purple}{\vX^* _0}, \vX_t}{\bx_s}$, where $\vX^* _0$ is an output from the previous step of the algorithm. 
At step $r$, we fit $\bx_s \mapsto \hpot{s}{\bx_s}[\param] \fw{s|0, t}{\textcolor{purple}{\vX^{r-1} _0}, \vX^{r-1} _t}{\bx_s}$, where $\vX^{r-1} _0$ is sampled using a few DDPM steps starting from $\vX^{r-1} _s$ at time $s$ and $\vX^{r-1} _t \sim \fw{t|s}{\vX^{r-1} _s}{}$. On the other hand, {\sc{MGPS}} fits in a single round the distribution with density proportional to $\bx_s \mapsto \hpot{s}{\bx_s}[\param] \fw{s|0, t+1}{\textcolor{purple}{\denoiser{t+1}{}{\vX_{t+1}}[\param]}, \vX_{t+1}}{\bx_s}$, where $\vX_{t+1}$ is the output of the previous step. Finally, the authors report that the performance of {\sc{MGPS}} improves when the number of gradient steps is increased. In our case, we have two axes, Gibbs iterations $R$ and gradient steps, that allow us to improve the performance when more compute is available. 

\section{Experiments details}
\subsection{Choice of weight sequence}
\label{apdx-sec:weight-seq}
In all our experiments we draw the index $s$, at time $t_i$, from $\mbox{Uniform}\intset{\tau}{t_{i-1}}$ with $\tau = 10$. The main motivation behind setting $\tau = 10$ and not $\tau = 1$, which is more natural, is that we have found that otherwise it may lead to instabilities. This arises typically when an index $s$ is sampled very close to $0$ when $t \approx T$. To avoid such behavior we use a smaller learning rate in \Cref{algo:gauss_vi} for the first few iterations and set $\tau > 1$. For the last $25\%$ diffusion steps we set $s$ deterministically to $t_{i-1}$ as we have found that this slightly improves the reconstructions quality. We also ramp up the number of gradient steps as this significantly sharpens the details in the images. 

\begin{figure}
\centering 
\includegraphics[width=\textwidth]{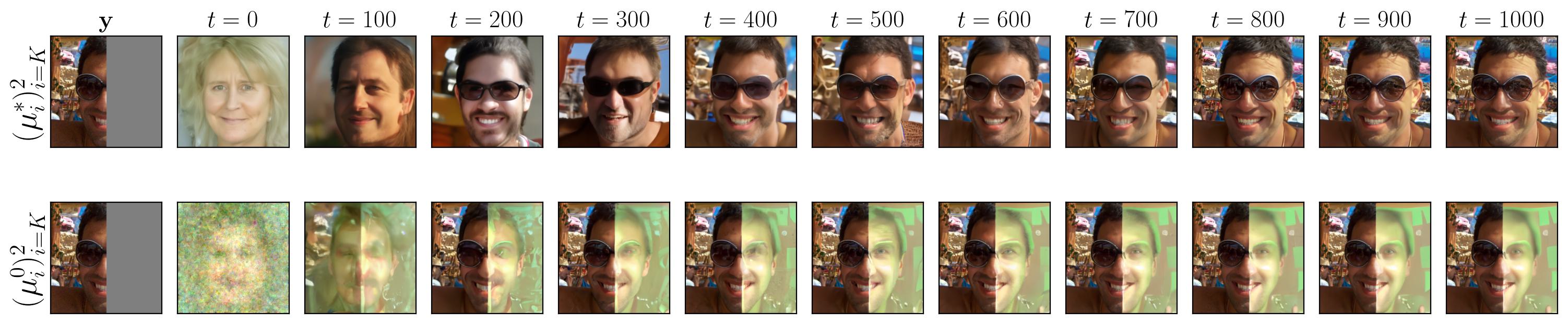}
\caption{Evolution of the running state $\vX^* _0$ in \Cref{algo:midpoint-gibbs} for the two time-sampling distributions given in \eqref{eq:sampling-dist-mix} and \eqref{eq:sampling-dist-zero}. }
\label{fig:sampling-comparison}
\end{figure}

While it is more intuitive to sample $s$ close to $0$ as it provides the best approximation error for the likelihood, we have found that this can significantly slow the mixing of the Gibbs sampler in very large dimensions and provides rather poor results when used with a small number of Gibbs steps. Practically speaking, significant artifacts arise during the initial iterations of the algorithm due to the optimization procedure, and they tend to persist in subsequent iterations when $s$ is sampled close to 0. To see why this is the case consider the following empirical discussion on a simplified scenario. We write $\bx = [\bar\bx, \underline\bx]$  where $\bar\bx \in \rset^\dimobs$ and $\underline\bx \in \rset^{\dimx - \dimobs}$. 
We assume that $\pot{0}{\bx} = \normpdf(\obs; \bar\bx, \stdobs^2 \Id_\dimobs)$, \emph{i.e.}, we observe only the first $\dimobs$ coordinates of the hidden state. Since $s$ is sampled near $0$ we may assume that $\hpot{s}{} = \pot{0}{}$. Then, sampling $Z \sim \epost{s|0, t}{\bx_0, \bx_t}{}$  is equivalent to sampling 
\begin{align*}
    \bar{Z} & \sim \gauss\left(\frac{\std^2 _{s|0, t}}{\stdobs^2 + \std^2 _{s|0,t}} \obs + \frac{\stdobs^2}{\stdobs^2 + \std^2 _{s|0, t}} \big[ \gamma_{t|s} \a_{s|0} \bar\bx_0 + (1 - \gamma_{t|s}) \a^{-1} _{t|s} \bar\bx_t\big], \frac{\stdobs^2 \std^2 _{s|0,t}}{\stdobs^2 + \std^2 _{s|0,t}} \Id_\dimobs\right) \eqsp, \\
    \underline{Z} & \sim \gauss(\gamma_{t|s} \a_{s|0} \underline\bx_0 + (1 - \gamma_{t|s}) \a^{-1} _{t|s} \underline\bx_t, \sigma^2 _{s|0,t} \Id_{\dimx - \dimobs}) \eqsp,
\end{align*}
setting $Z = [\bar{Z}, \underline{Z}]$ and then concatenating both vectors. It is thus seen that the observed part of the state is updated with the observation whereas the bottom part is simply drawn from the prior. Moreover, if $\std^2 _{\smash{s|0, t}} \approx 0$ then $\gamma_{t|s} \a_{s|0} \approx 1$ and $\underline{Z}$ is almost the same as $\bx_0$.  In \Cref{algo:midpoint-gibbs}, once we have sampled $\vX_s \sim \epost{\smash{s|0,t}}{\vX_0, \vX_t}{}$, we first denoise it to obtain the new $\vX_0$ and then noise it to obtain the new $\vX_t$. As $s$ is sampled near 0, the denoising step will merely modify $\vX_s$ whereas the noising step will add significant noise to $\vX_s$ and may help with removing the artifacts. This noised sampled has however only a small impact on the next samples $\vX_0, \vX_s$ since $(1 - \gamma_{t|s}) \a^{-1} _{t|s} \approx 0$. In short, the first $\dimobs$ coordinates of the running state $\vX^* _0$ will be quickly replaced by the observation whereas the last $\dimx - \dimobs$ coordinates will be stuck at their initialization and will evolve only by a small amount throughout the iterations of the algorithm. We illustrate this situation on a concrete example in \Cref{fig:sampling-comparison} where we consider a half mask inpainting task. The first and second rows show the evolution of the running state $\vX^* _0$ with the time-sampling distributions  
\begin{align}
    \label{eq:sampling-dist-mix}
  & \mu^* _{i} = \begin{cases} \mbox{Uniform}\intset{\tau}{t_{i-1}} \, & \quad \text{if} \quad i > \lfloor  K / 4 \rfloor \\
                    t_{i-1} \, & \quad \text{else} 
\end{cases},\\
\label{eq:sampling-dist-zero}
& \mu^0 _i = \mbox{Uniform}\intset{1}{\lfloor t_i / 5 \rfloor} \eqsp, 
\end{align}
\emph{i.e.}, the time-sampling distribution we use in all our experiments, where $K$ is the number of diffusion steps, and the one that we use to sample only close to $0$, respectively. In \Cref{table:sampling-comparison} we compute the LPIPS for both distributions on a subset of the tasks we consider in the main paper. It is clear that $\mu^* _i$ outperforms $\mu^0 _i$, even when we increase the number of Gibbs steps (see phase retrieval task). 


\begin{table} 
    \centering 
    \caption{LPIPS on the \ffhq\ dataset for the two time-sampling distributions given in \eqref{eq:sampling-dist-mix} and \eqref{eq:sampling-dist-zero}. We use $R = 4$ Gibbs steps for the phase retrieval task.}
    \resizebox{0.60\textwidth}{!}{
    \begin{tabular}{l cccc}
        \toprule
        Distribution & Phase retrieval ($R = 4$) & JPEG2 & Gaussian deblurring & Motion deblurring \\
        \midrule
        $\mu^* _t$  & \textbf{0.10} & \textbf{0.14} & \textbf{0.12} & \textbf{0.09} \\
        $\mu^0 _t$ & 0.53 & 0.19 & 0.16 & 0.19\\
        \bottomrule
    \end{tabular} 
    }
    \label{table:sampling-comparison}
\end{table}

\subsection{Hyperparameters setup of \algo}
\label{apdx-sec:hyperparameters}
The details about the hyperparameters of \algo\ are reported in \Cref{table:hyperparams-algo}.
We adjust the optimization of the Gaussian Variational approximation in \Cref{algo:gauss_vi} during the first and last diffusion steps.
We ramp up the number of gradient steps during the final diffusion steps.
This allows us to substantially improve the fine grained details of the reconstructions. 
Similarly, we reduce the learning rate in the early step to alleviate potential instabilities.

\begin{table}[ht]
    \centering
    \caption{The hyperparameters used in \algo\ for the considered datasets. The index $i$ of the timesteps $\{t_i\}_{i=K}^0$ is taken in reverse order. The symbol \# stands for \emph{``number of''}.}
    \vspace{-0.2cm}
    \renewcommand{\arraystretch}{1.3} 
    \resizebox{\textwidth}{!}{
    \begin{tabular}{l cccccc}
        \toprule
        & \# Gibbs repetitions $R$ & \# Diffusion steps $K$ & \# Denoising steps $M$ & Time-sampling distribution & Learning rate $\eta$ & \# Gradient steps $G$ \\
        \midrule
        \ffhq & $R=1$ & $K=100$ & $M=20$ & $\mu^* _{i}$ as in \eqref{eq:sampling-dist-mix} & $\eta=\begin{cases}
            0.01  & \text{ if } i \geq \lfloor 3K/4 \rfloor \\
            0.03  & \text{ otherwise} \\
            \end{cases}$ 
            & 
            $ G = \begin{cases}
            20 & \text{ if } i \leq \lfloor K/4 \rfloor \\
            5  & \text{ otherwise} \\
            \end{cases}$
        \\
        \midrule
        \ffhq\ LDM & $R=1$ & $K=100$ & $M=20$ & $\mu^* _{i}$ as in \eqref{eq:sampling-dist-mix}  & $\eta=\begin{cases}
            0.01  & \text{ if } i \geq \lfloor 3K/4 \rfloor \\
            0.03  & \text{ otherwise} \\
            \end{cases}$ 
            &
            $ G = \begin{cases}
            20 & \text{ if } i \leq \lfloor K/4 \rfloor \\
            20 & \text{ if } i \mod 10 = 0   \\
            3  & \text{ otherwise} \\
            \end{cases}$
        \\
        \midrule
        \imagenet & $R=1$ & $K=100$ & $M=20$ & $\mu^* _{i}$ as in \eqref{eq:sampling-dist-mix} & $\eta=\begin{cases}
            0.01  & \text{ if } i \geq \lfloor 3K/4 \rfloor \\
            0.03  & \text{ otherwise} \\
            \end{cases}$ 
            &
            $ G = \begin{cases}
            20 & \text{ if } i \leq \lfloor K/4 \rfloor \\
            5  & \text{ otherwise} \\
            \end{cases}$
        \\
        \midrule
        Audio-source separation & $R=6$ & $K=20$ & $M=15$ & $\mu^* _{i}$ as in \eqref{eq:sampling-dist-mix} & $\eta=0.005$ &  $ G = \begin{cases}
            20 & \text{ if } i \leq \lfloor K/4 \rfloor \\
            3  & \text{ otherwise} \\
            \end{cases}$
        \\
        \midrule
        \parbox[m]{10em}{Audio-source separation\\ (Best result in \Cref{table:si-snri})}
        & $R=1$ & $K=20$ & $M=15$ & $\mu^* _{i}$ as in \eqref{eq:sampling-dist-mix} & $\eta=0.005$ &  $ G = 90$
        \\
        \bottomrule
    \end{tabular}
    \label{table:hyperparams-algo}
    }
\end{table}

\subsection{Audio source separation}


In our experiment, the diffusion model employed provided by \cite{mariani2023multi} is trained on the \slakh\ training dataset\footnote{\url{http://www.slakh.com/}},  using only the four abundant instruments (bass, drums, guitar and piano) downsampled to 22 kHz. The denoiser network is based on a non-latent, time-domain unconditional variant of \citep{schneider2023musai}.

Its architecture follows a U-Net design, comprising an encoder, bottleneck, and decoder. The encoder consists of six layers with channel numbers $[256, 512, 1024, 1024, 1024, 1024]$, where each layer includes two convolutional ResNet blocks, and multihead attention is applied in the last three layers. The decoder mirrors the encoder structure in reverse. The bottleneck contains a ResNet block, followed by a self-attention mechanism, and then another ResNet block. Training is performed on the four stacked instruments using the publicly available trainer from repository\footnote{\url{https://github.com/archinetai/audio-diffusion-pytorch-trainer}}.

\subsection{Implementation of the competitors}
\label{apdx:competitors}
In this section, we provide implementation details of the competitors.
We adopt the hyperparameters recommended by the authors tune them on each dataset if they are not provided.
The complete set of hyperparameters and there values for both image experiments and audio-sound separation can be found in the supplementary material under the folders \texttt{configs\slash experiments/sampler} and \texttt{configs\slash exp\_sound/sampler}.

\paragraph*{DPS.}
We implemented \citet[Algorithm~1]{chung2023diffusion} and selected the hyperparameters of each considered task  based on \citet[App.~D]{chung2023diffusion}.
We tuned the algorithm for the other tasks, namely, we use $\gamma = 0.2$ for JPEG $2\%$, $\gamma = 0.07$ for High Dynamic Range tasks, and $\gamma = 1$ for audio-source separation.

\paragraph*{DiffPIR.}
We implemented \citet[Algorithm 1]{zhu2023denoising} to make it compatible with our existing code base.
We adopt the hyperparameters recommended in the official, released version\footnote{\url{https://github.com/yuanzhi-zhu/DiffPIR}}.
We followed the guidelines in \citep[Eqn. (13)]{zhu2023denoising} to extend the algorithm to nonlinear problems.
However, we noticed that the algorithm diverges in these cases and we could not follow up as the paper and the released code lack examples of nonlinear problems.
\citet{zhu2023denoising} provides an FFT-based solution for the motion blur tasks which is only valid in the case of circular convolution.
Hence, and since we adapted the experimental setup of \citet{chung2023diffusion}, we do not run the algorithm on motion blur task as it uses convolution with reflect padding. For audio-source separation, we found that $\lambda = \mu = 1$ works best.

\paragraph*{DDNM.}
We adapted the implementation provided in the released code\footnote{\url{https://github.com/wyhuai/DDNM}}.
Namely, the authors provide classes, in the module \texttt{functions\slash svd\_operators.py} that implement the logic of the algorithm on each degradation operator separately.
The adaptation includes factorizing these classes to a single class to support all SVD linear degradation operators.
On the other hand, we notice \ddnm\ is unstable for operators whose SVD decomposition is prone to numerical errors, such as Gaussian Blur with wide convolution kernel. This results from the algorithm using the pseudo-inverse of the operator.

\paragraph*{RedDiff.}
We used the implementation of \reddiff\ available in the released code\footnote{\url{https://github.com/NVlabs/RED-diff}}.
For linear problems, we use the pseudo-inverse of the observation as an initialization of the variational optimization problem. 
On nonlinear problems, for which the pseudo-inverse of the observation is not available, we initialized the optimization with a sample from the standard Gaussian distribution. 

\paragraph*{PGDM.}
We opted for the implementation available in the \reddiff\'s repository as some the authors are co-authors of \pgdm as well.
Notably, the implementation introduces a subtle deviation from \citet[Algorithm 1]{song2022pseudoinverse}: in the algorithm's final step, the guidance term $g$ is scaled by $\a_t$ ($\sqrt{\a_t}$ in their notation) whereas the implementation scales it by $\a_{t-1}\a_t$.
This adjustment improves the algorithm for most tasks except for JPEG dequantization. We found that the original scaling by $\a_t$ is better in this case.

\paragraph*{PSLD.}
We implemented the \psld\ algorithm provided in \citet[Algorithm 2]{rout2024solving} and referred to the publicly available implementation\footnote{\url{https://github.com/LituRout/PSLD}} to set the hyperparameters of the algorithm for the different tasks.

\paragraph*{ReSample.}
We modified the original code\footnote{\url{https://github.com/soominkwon/resample}} provided by the authors to make its hyperparameters directly adjustable, namely, the tolerance $\varepsilon$ and the maximum number of iterations $N$ for solving the optimization problems related to hard data consistency, and the scaling factor for the variance of the stochastic resampling distribution $\gamma$.
We found the algorithm to be sensitive to $\varepsilon$ and that setting it to the noise level of the inverse problem yields the best reconstructions across tasks and noise levels.
On the other hand, we noticed that $\gamma$ has less impact on the quality of the reconstructions.
Finally, we set a threshold $N=200$ on the maximum number of gradient iterations to make the algorithm less computationally intensive.

\paragraph*{DAPS.}
We have the official codebase \footnote{\url{https://github.com/zhangbingliang2019/DAPS}}.
We referred to \citet[Table.~7]{zhang2024daps} to set the hyperparameters.
For audio-source separation, we set $\sigma_{\max}$ and $\sigma_{\min}$ to match those of the sound model and adapted the Langevin stepsize \texttt{lr} and the standard deviation \texttt{tau} to the audio-separation task.

\begin{figure}[tb]
    \centering
    \subfigure{
        \includegraphics[width=.49\textwidth]{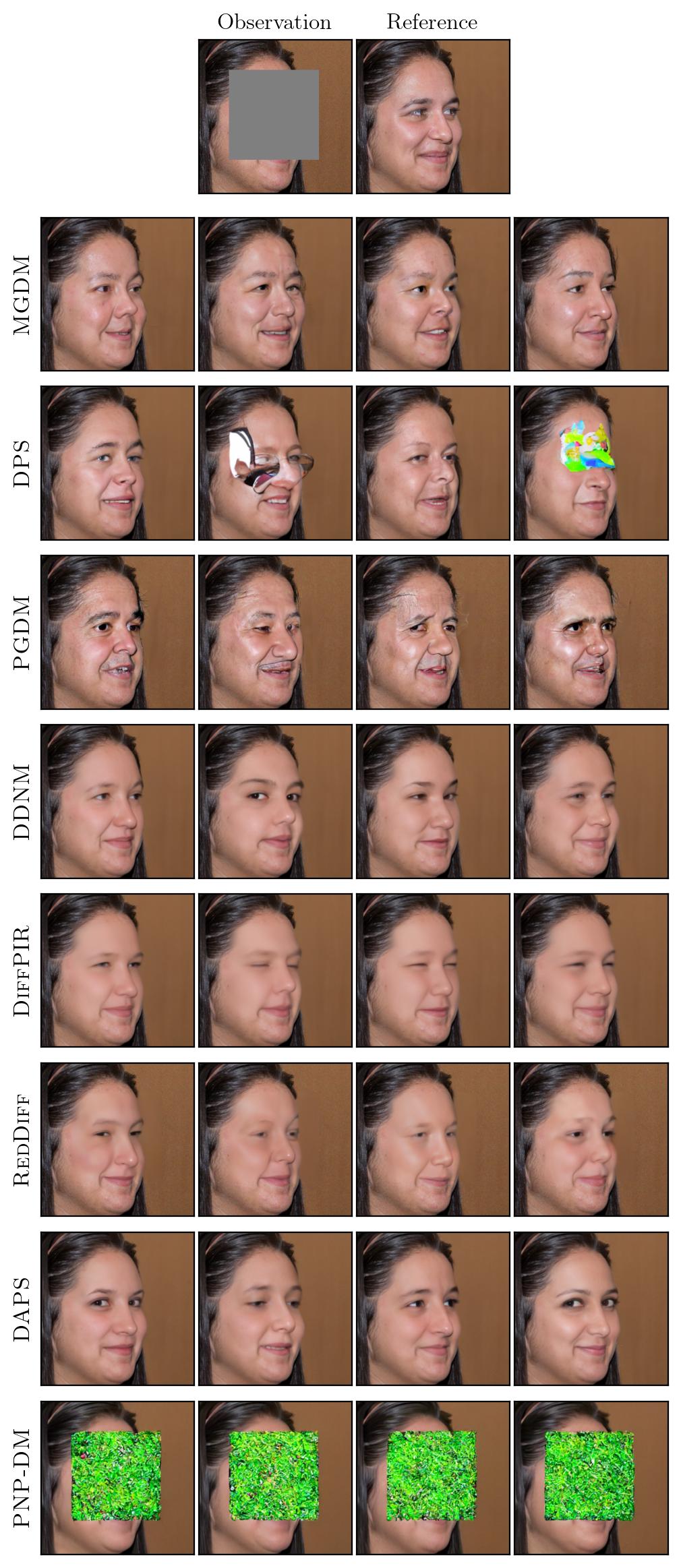}
        \hfill%
        \includegraphics[width=.49\textwidth]{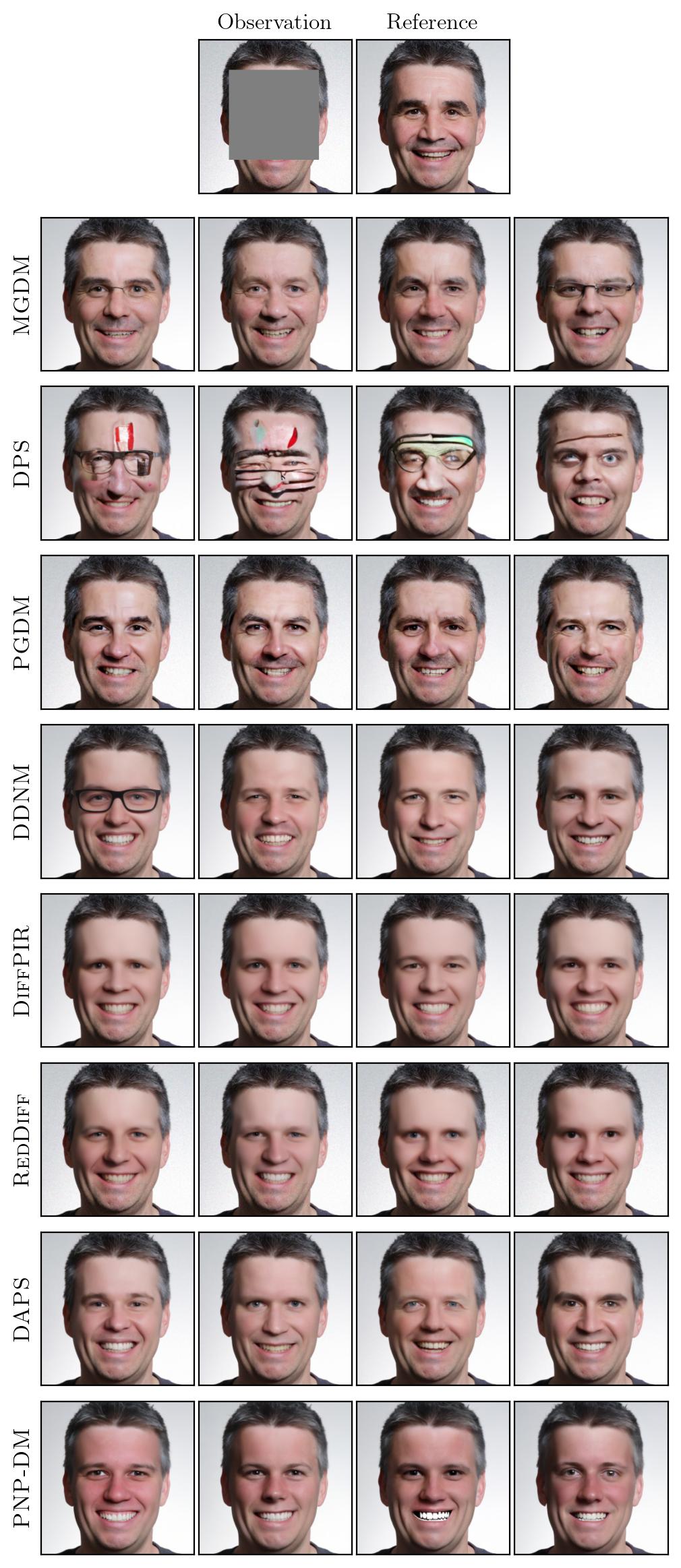}
    }
    \caption{Reconstructions for box inpainting on \ffhq\ dataset.}
    \label{fig:pnpdm-conjugacy}
\end{figure}

\paragraph*{PNP-DM.}
We adapted the implementation provided in the released code\footnote{\url{https://github.com/zihuiwu/PnP-DM-public/}}.
Specifically, we exposed the coupling parameter $\rho$ including its initial value, minimum value, and decay rate, as well as the number of Langevin steps and its step size.
The hyperparameters were set based on \citet[Table 3 and Table 4]{wu2024pnpdm}.
For inpainting tasks, while it is theoretically possible to perform the likelihood steps using Gaussian conjugacy \citep[Sec.~3.1]{wu2024pnpdm}, we found that using Langevin produced better results in practice. For example, the reconstructions in the left figure of \Cref{fig:pnpdm-conjugacy} are obtained by sampling exactly from the posterior whereas on the \rhs\ we use Langevin dynamics. 
Although the audio separation task is linear and hence the likelihood steps can be implemented exactly, we encountered similar challenges as in inpainting and therefore we used Langevin here as well.

\subsection{Experiments reproducibility}
Our code will be made available upon acceptange of the paper. In the anonymous codebase provided as companion of the paper we use $\sqrt{\a_t}$ instead of $\a_t$ to match the conventions of existing codebases.  All experiments were conducted on Nvidia Tesla V100 SXM2 GPUs. 
For the image experiments, we used $300$ images from the validation sets of \ffhq\ and \imagenet\ $256 \times 256$ that we numbered from $0$ to $299$.
The image number was used to seed the randomness of the experiments on that image.
For the audio source separation experiments, the \slakh\ test dataset has tracks named following the pattern \texttt{Track0XXXX}, where \texttt{X} represents a digit in $0-9$.
The number \texttt{XXXX} was used as the seed for the experiments conducted on each track.



\subsection{Extended results}
\label{apdx:extended-results}
We present the complete table with LPIPS, PSNR, and SSIM metrics for the image inverse problems experiment in \Cref{table:extended-ffhq-imagenet} for the \ffhq\ and \imagenet\ datasets, and in \Cref{table:extended-ffhq-ldm} for \ffhq\ LDM. Similarly, the complete results for the audio source separation experiments that include all competitors are provided in \Cref{table:extended-si-snri}.

From \Cref{table:extended-ffhq-imagenet}, one can note that \ddnm, \diffpir\ and \daps\ score better in PSNR and SSIM compared to \algo\; but score lower in LPIPS.
For most of the tasks we considered, one does not expect to recover an image very close to the reference and thus, metrics that perform pixel-wise comparisons are less relevant and favor images that are overly smooth.
We provide evidence for this in the gallery of images below where we compare qualitatively the outputs of our algorithm with those of the competitors.
It can be seen that our method provides reconstructions with ine-grained details that more coherent with the reference image.
Note for example that \ddnm, \diffpir\ and \daps\ outperform \algo\ in terms of PSNR and SSIM on the half mask task on \imagenet\ while failing to reconstruct the missing \rhs\ of the images.

\begin{table}[h]
    \centering
    \caption{Mean LPIPS/PSNR/SSIM metrics for the considered linear and nonlinear imaging tasks on the \ffhq\ and \imagenet\ $256 \times 256$ datasets with $\stdobs = 0.05$.}
    \resizebox{\textwidth}{!}{
    \begin{tabular}{l cccccccc | cccccccc}
        \toprule
        \vspace{1mm}
        & \multicolumn{8}{c}{\bf{\ffhq}} & \multicolumn{8}{c}{\bf{\imagenet}} \\
        \textbf{Task} & \algo\ & \dps & \pgdm & \ddnm & \diffpir & \reddiff & \daps & \pnpdm \ &\ \algo\ & \dps & \pgdm & \ddnm & \diffpir & \reddiff & \daps & \pnpdm \\
        \midrule
        & \multicolumn{16}{c}{LPIPS \ $\downarrow$} \\
        \midrule
        SR ($\times 4$)        & \first{0.09} & \first{0.09} & 0.30 & \third{0.15} & \second{0.10} & 0.39 & 0.16 & \second{0.10} \ &\ \second{0.26} & \first{0.25} & 0.56 & 0.34 & \third{0.31} & 0.57 & 0.37 & 0.66 \\
        SR ($\times 16$)       & \second{0.24} & \first{0.23} & 0.42 & 0.33 & \first{0.23} & 0.55 & 0.40 & \third{0.29} \ &\ \third{0.55} & \first{0.44} & 0.62 & 0.71 & \second{0.50} & 0.85 & 0.75 & 1.03 \\
        Box inpainting         & \first{0.10} & 0.17 & 0.17 & \second{0.12} & 0.14 & 0.19 & \third{0.13} & 0.18 \ &\ \first{0.23} & 0.35 & \third{0.29} & \second{0.28} & 0.30 & 0.36 & 0.30 & 0.42 \\
        Half mask              & \first{0.20} & \third{0.24} & \third{0.24} & \second{0.23} & 0.25 & 0.28 & \second{0.23} & 0.32 \ &\ \first{0.31} & 0.40 & \second{0.34} & \third{0.38} & 0.40 & 0.46 & 0.40 & 0.54 \\
        Gaussian Deblur        & \first{0.12} & \third{0.17} & 0.87 & 0.20 & \first{0.12} & 0.24 & 0.24 & \second{0.14} \ &\ \first{0.30} & \second{0.37} & 1.00 & \third{0.45} & \first{0.30} & 0.53 & 0.59 & 0.76 \\
        Motion Deblur          & \first{0.09} & \second{0.17} & $-$ & $-$ & $-$ & 0.22 & \third{0.19} & 0.21 \ &\ \first{0.22} & \third{0.40} & $-$ & $-$ & $-$ & \second{0.39} & 0.42 & 0.52 \\
        JPEG (QF = 2)          & \first{0.14} & 0.34 & 1.12 & $-$ & $-$ & 0.32 & \second{0.22} & \third{0.29} \ &\ \first{0.38} & 0.60 & 1.32 & $-$ & $-$ & \third{0.49} & \second{0.45} & 0.56 \\
        Phase retrieval        & \first{0.11} & 0.40  & $-$ & $-$ & $-$ & \third{0.26} & \second{0.14} & 0.34 \ &\ \second{0.55} & 0.62 & $-$ & $-$ & $-$ & \third{0.61} & \second{0.50} & 0.66 \\
        Nonlinear deblur       & \first{0.27} & 0.51 & $-$ & $-$ & $-$ & 0.68 & \second{0.28} & \third{0.31} \ &\ \first{0.41} & 0.82 & $-$ & $-$ & $-$ & \third{0.66} & \first{0.41} & \second{0.49} \\
        HDR    & \second{0.12} & 0.40 & $-$ & $-$ & $-$ & 0.20 & \second{0.10} & \third{0.19} \ &\ \third{0.21} & 0.84 & $-$ & $-$ & $-$ & \second{0.19} & \first{0.14} & 0.31 \\
        \midrule
        & \multicolumn{16}{c}{PSNR \ $\uparrow$} \\
        \midrule
        SR ($\times 4$)        & 27.66 & \third{28.05} & 24.57 & \first{29.45} & 27.72 & 26.75 & \second{28.44} & 27.44 \ &\ 23.88 & \third{24.37} & 18.45 & \first{24.99} & 23.43 & 23.33 & \second{24.38} & 16.4 \\
        SR ($\times 16$)       & \third{21.01} & 20.71 & 18.51 & \first{22.32} & 20.96 & \second{21.46} & 19.75 & 20.88\ &\ 18.12 & 17.66 & 15.27 & \first{19.93} & \third{18.4} & \second{19.06} & 18.18 & 14.0 \\
        Box inpainting         & \second{22.38} & 18.81 & 21.05 & \third{22.34} & \first{22.39} & 21.46 & 22.06 & 20.42 \ &\ 16.82 & 13.92 & 16.73 & \first{19.18} & \third{19.05} & 18.21 & \second{19.11} & 18.03 \\
        Half mask              & 15.39 & 14.86 & 15.29 & \first{16.38} & \third{16.04} & 15.68 & \second{16.25} & 14.35 \ &\ 13.77 & 12.15 & 14.04 & \second{15.97} & \third{15.64} & 14.84 & \first{16.00} & 14.88 \\
        Gaussian Deblur        & 25.64 & 24.03 & 13.34 & \second{26.62} & 25.78 & \first{26.68} & \third{26.12} & 25.89 \ &\ 21.57 & 20.65 & 9.92 & \first{22.89} & 21.8 & \second{22.72} & \third{22.41} & 15.85 \\
        Motion Deblur          & \first{27.82} & 24.13 & $-$ & $-$ & $-$ & \second{27.48} & \third{27.07} & 24.91 \ &\ \first{24.46} & 21.38 & $-$ & $-$ & $-$ & \second{24.06} & \third{23.64} & 22.47 \\
        JPEG (QF = 2)          & \second{25.57} & 19.56 & 12.57 & $-$ & $-$ & \third{24.53} & \first{25.72} & 22.42 \ &\ \third{21.42} & 16.33 & 5.27 & $-$ & $-$ & \second{22.07} & \first{22.68} & 20.74 \\
        Phase retrieval        & \second{27.55} & 16.56 & $-$ & $-$ & $-$ & \third{24.58} & \first{27.84} & 21.63 \ &\ \second{16.01} & 14.12 & $-$ & $-$ & $-$ & \third{15.41} & \first{18.44} & 15.02  \\
        Nonlinear deblur       & \third{23.55} & 16.08 & $-$ & $-$ & $-$ & 21.94 & \first{24.56} & \second{24.08} \ &\ \third{21.96} & 10.13 & $-$ & $-$ & $-$ & 20.57 & \first{22.68} & \second{22.20} \\
        HDR                    & \second{24.79} & 18.71 & $-$ & $-$ & $-$ & \third{21.69} & \first{26.60} & 21.59 \ &\ \second{22.90} & 9.56 & $-$ & $-$ & $-$ & 22.12 & \first{24.69} & \third{22.23} \\
        %
        \midrule
        & \multicolumn{16}{c}{SSIM \ $\uparrow$} \\
        \midrule
        SR ($\times 4$)        & 0.80 & \second{0.81} & 0.56 & \first{0.85} & 0.78 & 0.68 & \second{0.81} & 0.77\ &\ 0.65 & \second{0.68} & 0.30 & \first{0.71} & 0.60 & 0.57 & \third{0.66} & 0.25 \\ 
        SR ($\times 16$)       & \second{0.61} & 0.58 & 0.42 & \first{0.67} & 0.59 & \third{0.60} & 0.58 & 0.57\ &\ 0.31 & 0.39 & 0.21 & \first{0.49} & 0.41 & \second{0.44} & \second{0.44} & 0.10 \\
        Box inpainting         & \third{0.80} & 0.77 & 0.70 & \first{0.83} & \second{0.82} & 0.70 & \third{0.80} & 0.75\ &\ 0.71 & 0.70 & 0.62 & \first{0.77} & \second{0.76} & 0.67 & \third{0.74} & 0.64 \\
        Half mask              & 0.67 & 0.67 & 0.59 & \first{0.74} & \second{0.72} & 0.63 & \third{0.71} & 0.65\ &\ 0.59 & 0.58 & 0.52 & \first{0.68} & \second{0.67} & 0.59 & \third{0.66} & 0.57 \\
        Gaussian Deblur        & 0.73 & 0.68 & 0.14 & \first{0.77} & 0.72 & \second{0.76} & \third{0.75} & 0.72\ &\ 0.50 & 0.50 & 0.08 & \first{0.59} & 0.51 & \second{0.57} & \third{0.56} & 0.20 \\
        Motion Deblur          & \first{0.80} & 0.70 & $-$ & $-$ & $-$ & 0.71 & \second{0.78} & \third{0.75}\ &\ \first{0.67} & 0.55 & $-$ & $-$ & $-$ & \third{0.61} & \second{0.63} & 0.57 \\
        JPEG (QF = 2)          & \second{0.74} & 0.56 & 0.10 & $-$ & $-$ & \third{0.71} & \first{0.76} & 0.70\ &\ 0.51 & 0.40 & 0.02 & $-$ & $-$ & \second{0.59} & \first{0.62} & \third{0.58}  \\
        Phase retrieval        & \second{0.78} & 0.49 & $-$ & $-$ & $-$ & \third{0.61} & \first{0.81} & 0.57\ &\ \second{0.31} & \third{0.27} & $-$ & $-$ & $-$ & 0.25 & \first{0.46} & 0.23 \\
        Nonlinear deblur       & \third{0.67} & 0.44 & $-$ & $-$ & $-$ & 0.42 & \first{0.71} & \second{0.70}\ &\ \second{0.58} & 0.25 & $-$ & $-$ & $-$ & 0.41 & \first{0.61} & \second{0.58} \\
        HDR                    & \second{0.76} & 0.55 & $-$ & $-$ & $-$ & \third{0.72} & \first{0.85} & 0.69\ &\ \second{0.72} & 0.23 & $-$ & $-$ & $-$ & \second{0.72} & \first{0.82} & 0.66 \\
        \bottomrule
    \end{tabular}
    }
    \label{table:extended-ffhq-imagenet}
\end{table}

\begin{table}[h]
    \centering
    \caption{Mean LPIPS/PSNR/SSIM for linear/nonlinear imaging tasks on \ffhq\ $256 \times 256$ datasets with LDM and $\stdobs = 0.05$.}
    \resizebox{\textwidth}{!}{
    \begin{tabular}{l ccccc c ccccc c ccccc}
        \toprule
        & \algo\ & \resample & \psld & \daps & \pnpdm  && \algo\ & \resample & \psld & \daps & \pnpdm && \algo\ & \resample & \psld & \daps & \pnpdm \\
        \cmidrule(lr){2-6} \cmidrule(lr){8-12} \cmidrule(lr){14-18}
        Task & \multicolumn{5}{c}{LPIPS \ $\downarrow$} && \multicolumn{5}{c}{PSNR \ $\uparrow$} && \multicolumn{5}{c}{SSIM \ $\uparrow$} \\
        \midrule
        SR ($\times 4$)    & \first{0.14} & \third{0.22} & \second{0.21} & 0.28 & 0.40   && \second{27.39} & \third{25.85} & 25.80 & \first{27.45} & 23.81  && \first{0.79} & 0.68 & \second{0.71} & \first{0.79} & 0.70  \\
        SR ($\times 16$)   & \first{0.30} & \third{0.38} & \second{0.36} & 0.52 & 0.71   && \third{20.60} & \second{20.97} & \first{21.42} & 19.91 & 17.07  && \third{0.58} & 0.56 & \first{0.63} & \second{0.59} & 0.52  \\
        Box inpainting     & \first{0.18} & \second{0.22} & \third{0.27} & 0.37 & 0.31   && \first{21.81} & 18.56 & \second{20.01} & 11.77 & \third{19.57}  && \first{0.78} & \second{0.75} & 0.66 & 0.70 & \third{0.73}  \\
        Half mask          & \first{0.26} & \second{0.30} & \third{0.32} & 0.49 & 0.44   && \first{15.71} & \second{14.89} & \third{14.62} &  9.13 & 14.15  && \first{0.69} & \second{0.67} & 0.60 & 0.55 & \third{0.65}  \\
        Gaussian Deblur    & \second{0.18} & \first{0.16} & 0.59 & \third{0.32} & \third{0.32}   && \third{26.79} & \first{27.28} & 17.99 & \second{26.86} & 26.11  && \second{0.77} & \third{0.75} & 0.27 & \first{0.78} & \second{0.77}  \\
        Motion Deblur      & \second{0.22} & \first{0.20} & 0.70 & \third{0.36} & \third{0.36}   && \third{25.27} & \first{26.73} & 17.71 & \second{25.37} & 24.65  && \second{0.73} & \third{0.72} & 0.24 & \first{0.74} & \third{0.72}  \\
        JPEG (QF = 2)      & \first{0.23} & \second{0.26} & $-$ & \third{0.32} & 0.36   && \third{24.27} & \second{24.77} & $-$ & \first{25.22} & 23.86  && \third{0.71} & 0.66 & $-$ & \first{0.75} & \second{0.72}  \\
        Phase retrieval    & \second{0.29} & \third{0.39} & $-$ & \first{0.25} & 0.50   && \second{22.54} & \third{20.18} & $-$ & \first{27.05} & 20.03  && \second{0.62} & 0.49 & $-$ & \first{0.79} & \third{0.60}  \\
        Nonlinear deblur   & \first{0.29} & \second{0.33} & $-$ & \third{0.37} & \third{0.37}   && \second{23.71} & \first{24.10} & $-$ & 22.03 & \third{23.28}  && \second{0.69} & 0.67 & $-$ & \third{0.68} & \first{0.70}  \\
        High dynamic range & \second{0.16} & \first{0.12} & $-$ & \third{0.24} & \third{0.24}   && \second{25.59} & \first{25.91} & $-$ & \third{20.95} & 20.21  && \second{0.80} & \first{0.83} & $-$ & \third{0.74} & 0.73  \\
        \bottomrule
    \end{tabular}
    }
    \label{table:extended-ffhq-ldm}
\end{table}

\begin{table}[h]
    \centering
    \captionsetup{font=small}
    \caption{Mean \sisdri\ on \slakh\ test dataset. The last row  "All" displays the mean over the four stems. Higher metrics are better.}
    \resizebox{0.64\textwidth}{!}{
    \begin{tabular}{l ccccccccc | cc}
        \toprule
        Stems  & \algo\ & \dps & \pgdm & \ddnm & \diffpir & \reddiff & \daps & \pnpdm    & \msdm & \isdm   & \demucs \\
        \midrule
        Bass   & \first{18.49} & \third{16.50} & 16.41 & 14.94 & -2.34 & -0.40 & 11.76 & 2.90 & \second{17.12} & 19.36   & 17.16   \\
        Drums  & 18.07 & \third{18.29} & 18.14 & \first{19.05} & 9.47 & -0.98 & 15.62 & 7.89 & \second{18.68} & 20.90   & 19.61   \\
        Guitar &\first{16.68} & 9.90 & 12.84 & \third{14.38} & -1.01 & 5.68 & 11.75 & 4.51 & \second{15.38}  & 14.70   & 17.82   \\
        Piano  & \first{16.17} & 10.41 & \third{12.31} & 11.46 & 0.97 & 5.04 & 9.52 & 4.09 & \second{14.73}  & 14.13   & 16.32   \\
        \midrule
        All    & \first{17.35} & 13.77 & 14.92 & \third{14.96} & 1.77 & 2.33 & 12.16 & 4.85 & \second{16.48}   & 17.27   & 17.73   \\
        \bottomrule
    \end{tabular}
    }
    \label{table:extended-si-snri}
\end{table}

\begin{figure}[tb]
    \centering
    \subfigure{
        \includegraphics[width=.49\textwidth]{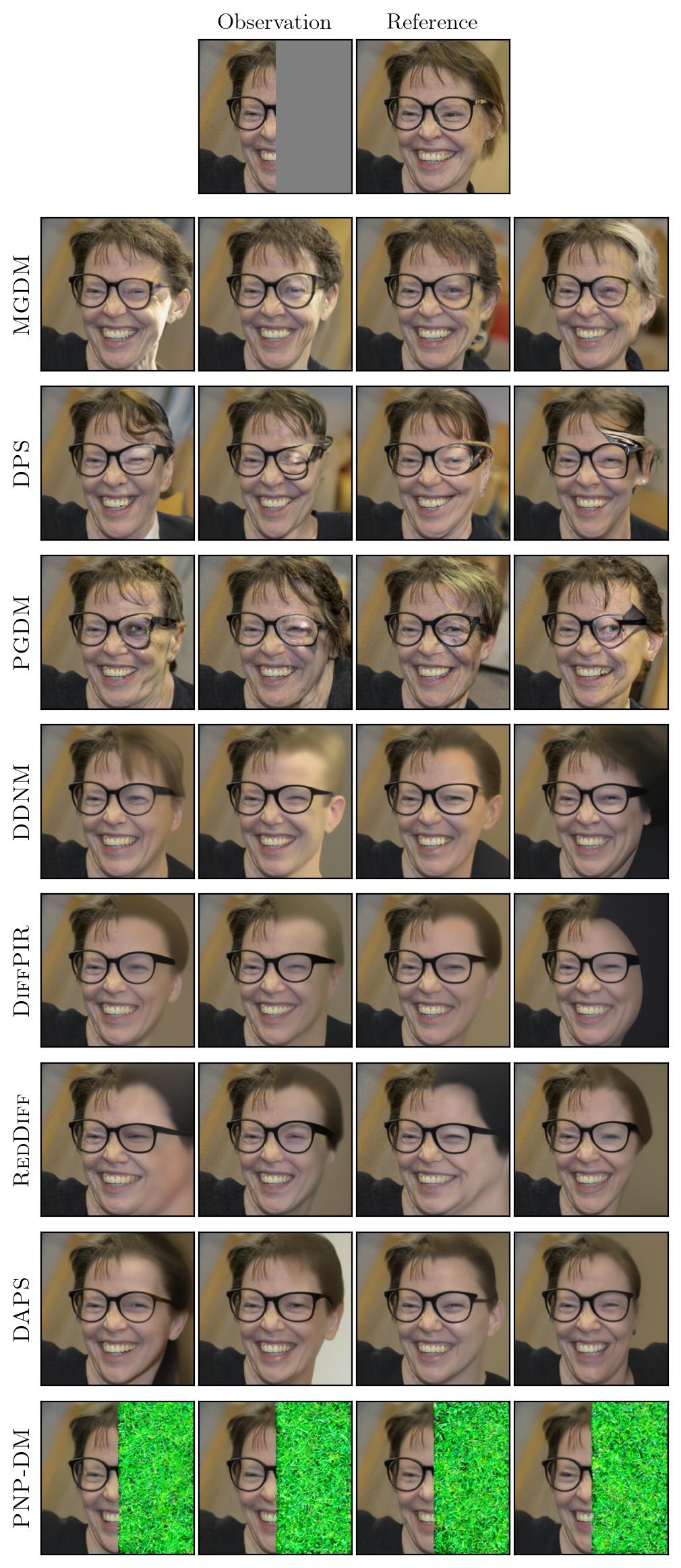}
        \includegraphics[width=.49\textwidth]{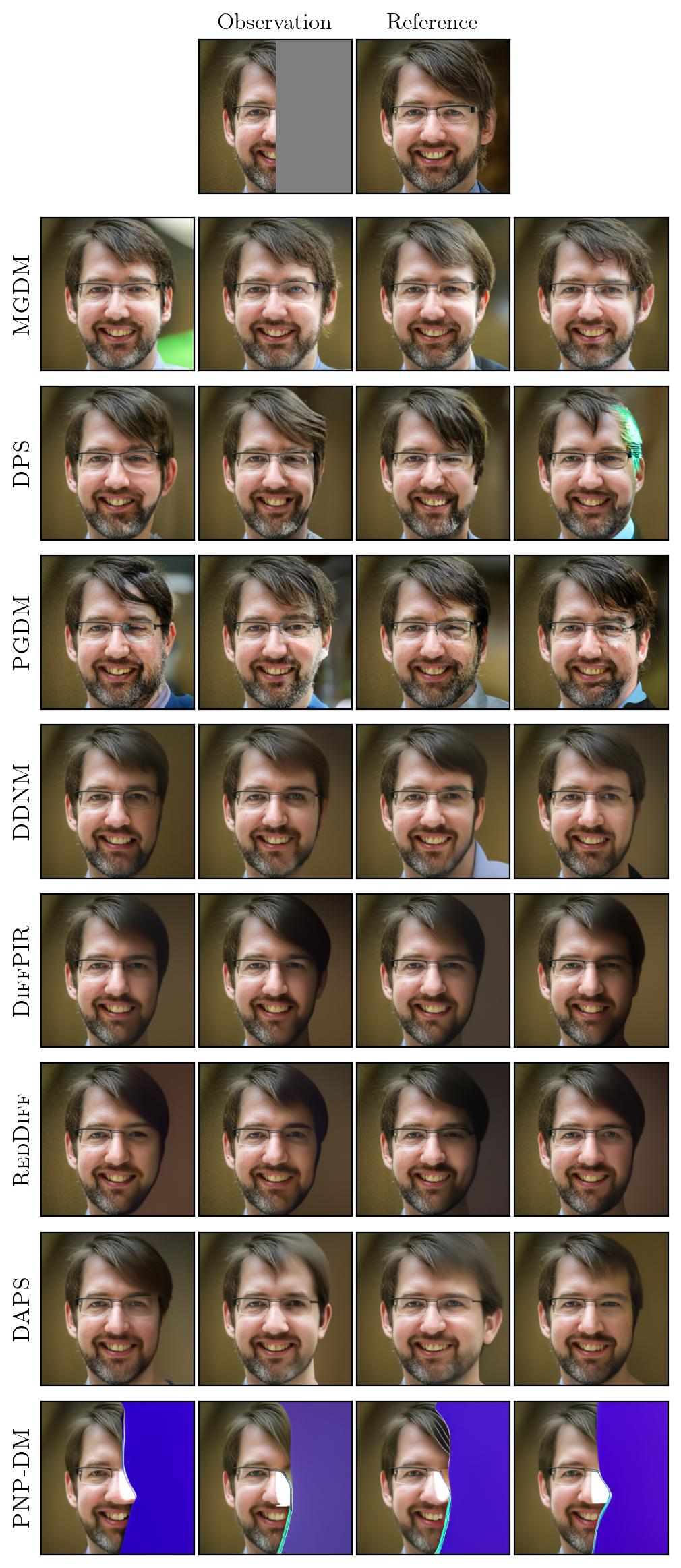}
    }
    \caption{Reconstructions for half mask inpainting on \ffhq\ dataset.}
\end{figure}

\begin{figure}[tb]
    \centering
    \subfigure{
        \includegraphics[width=.49\textwidth]{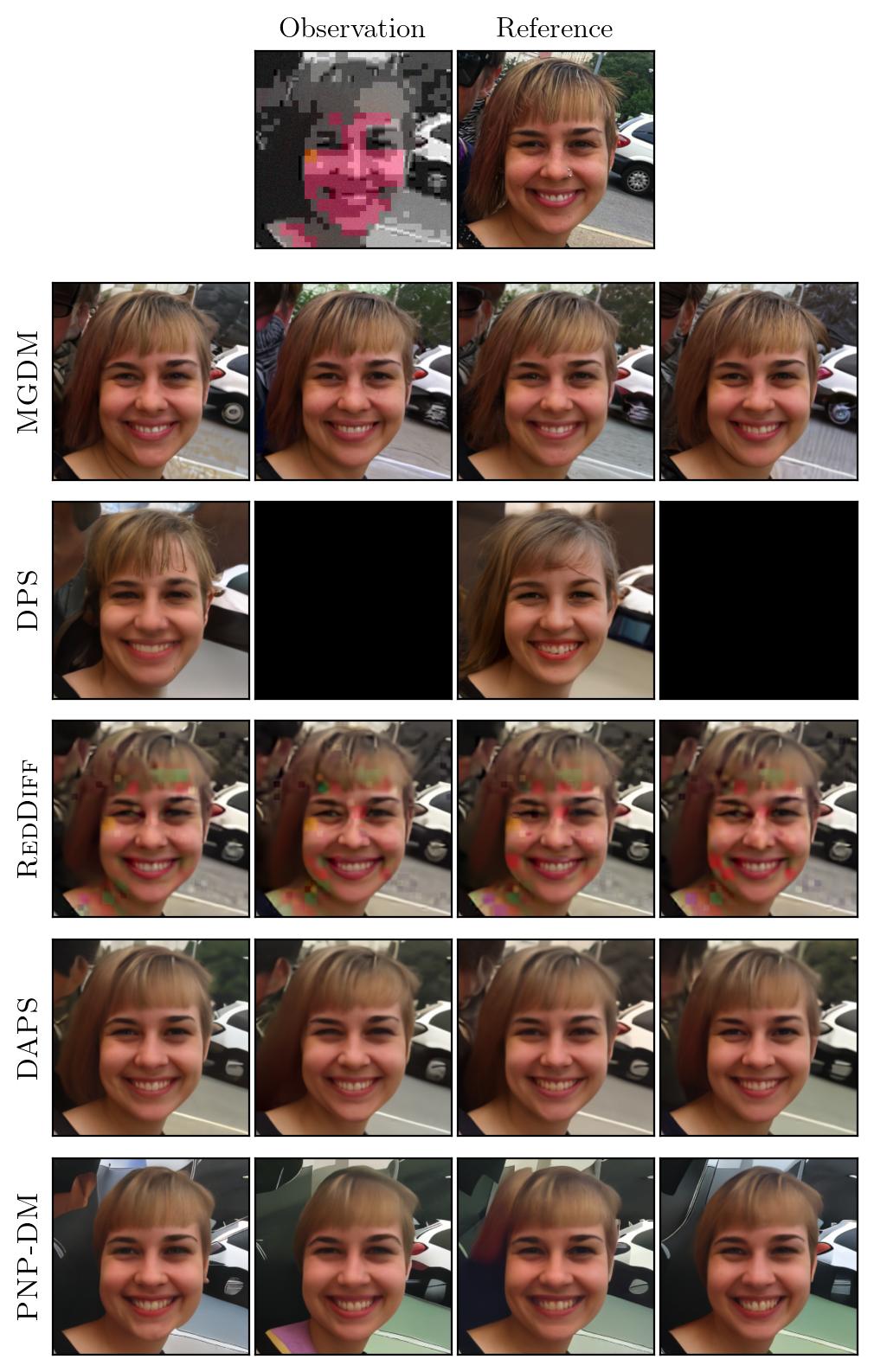}
        \includegraphics[width=.49\textwidth]{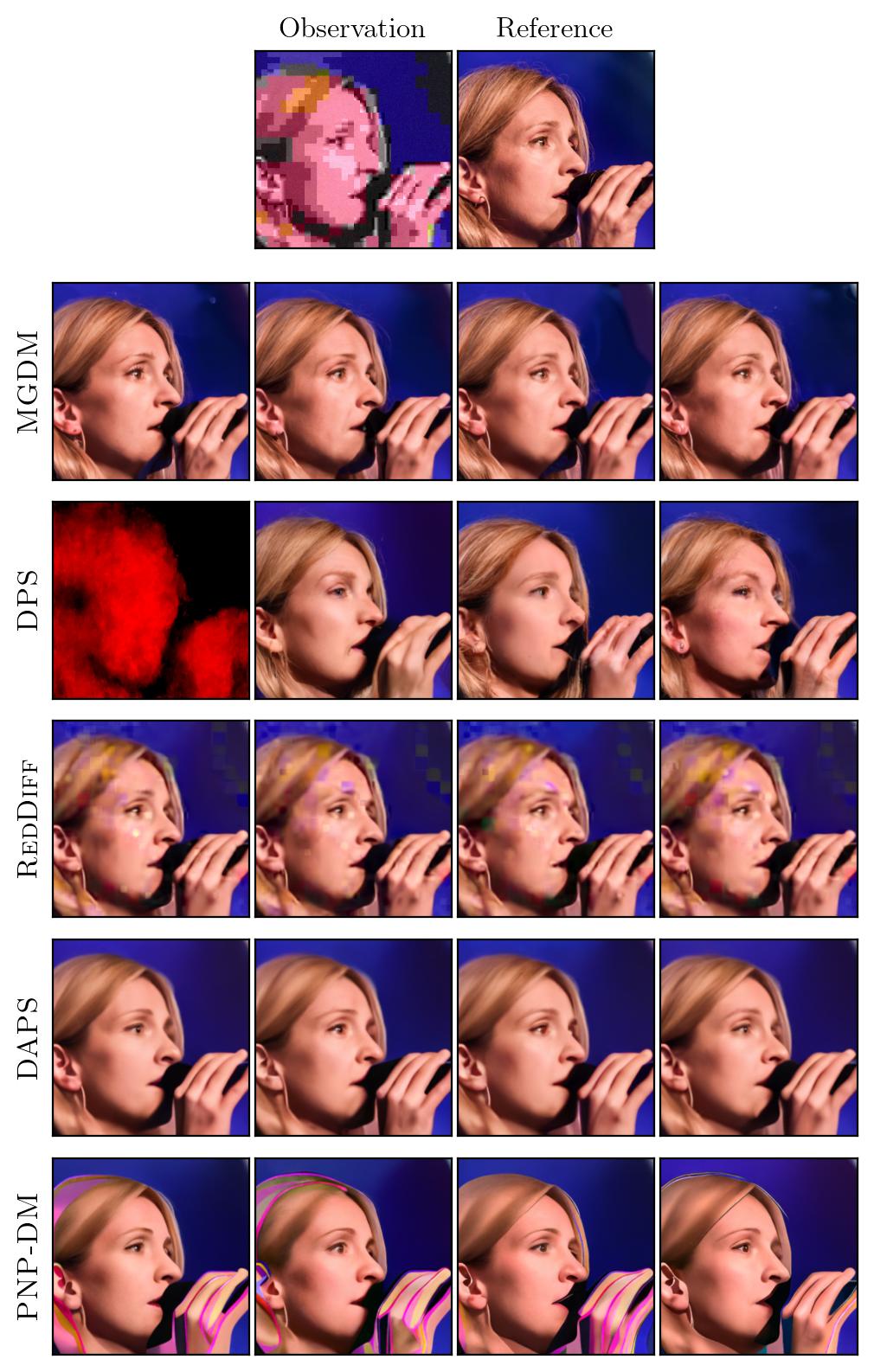}
    }
    \caption{Reconstructions for JPEG dequantization QF=2\% on \ffhq\ dataset.}
\end{figure}

\begin{figure}[tb]
    \centering
    \subfigure{
        \includegraphics[width=.49\textwidth]{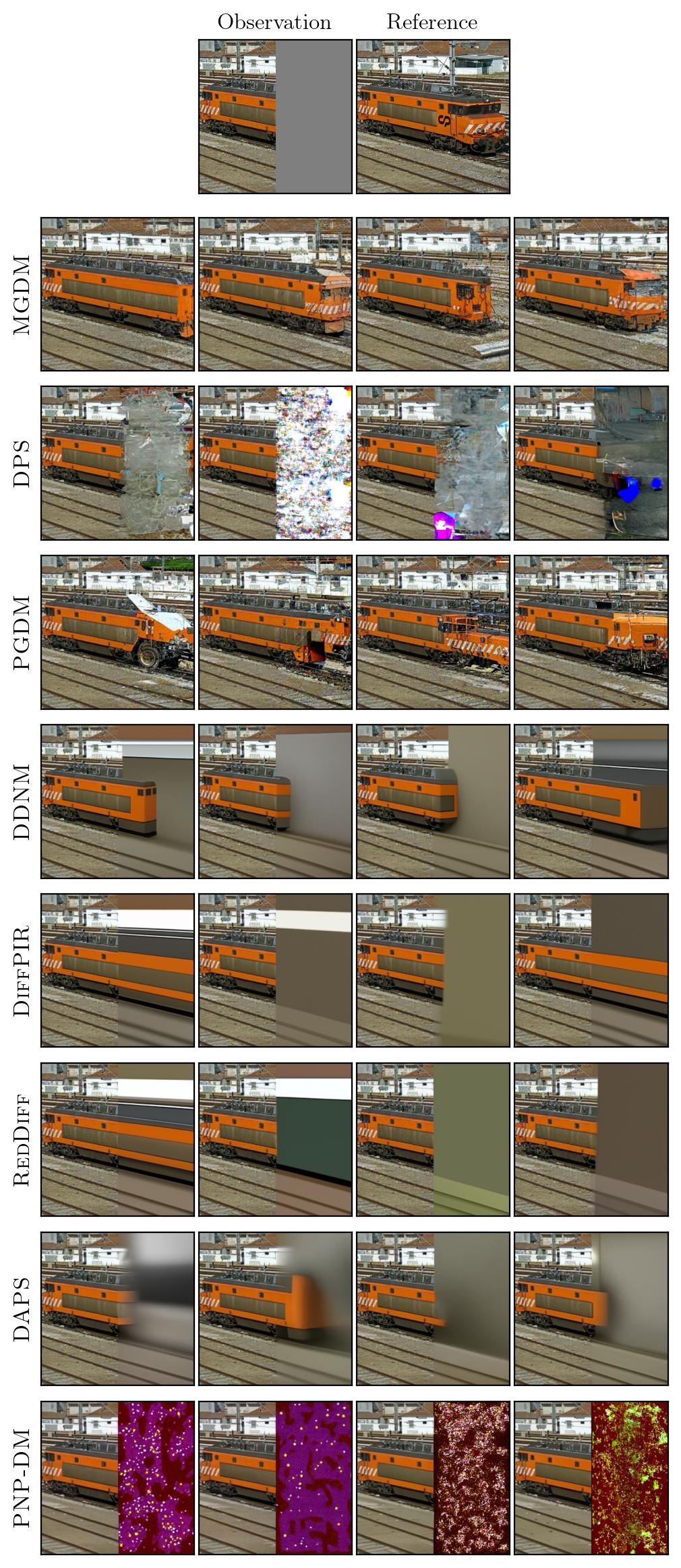}
        \includegraphics[width=.49\textwidth]{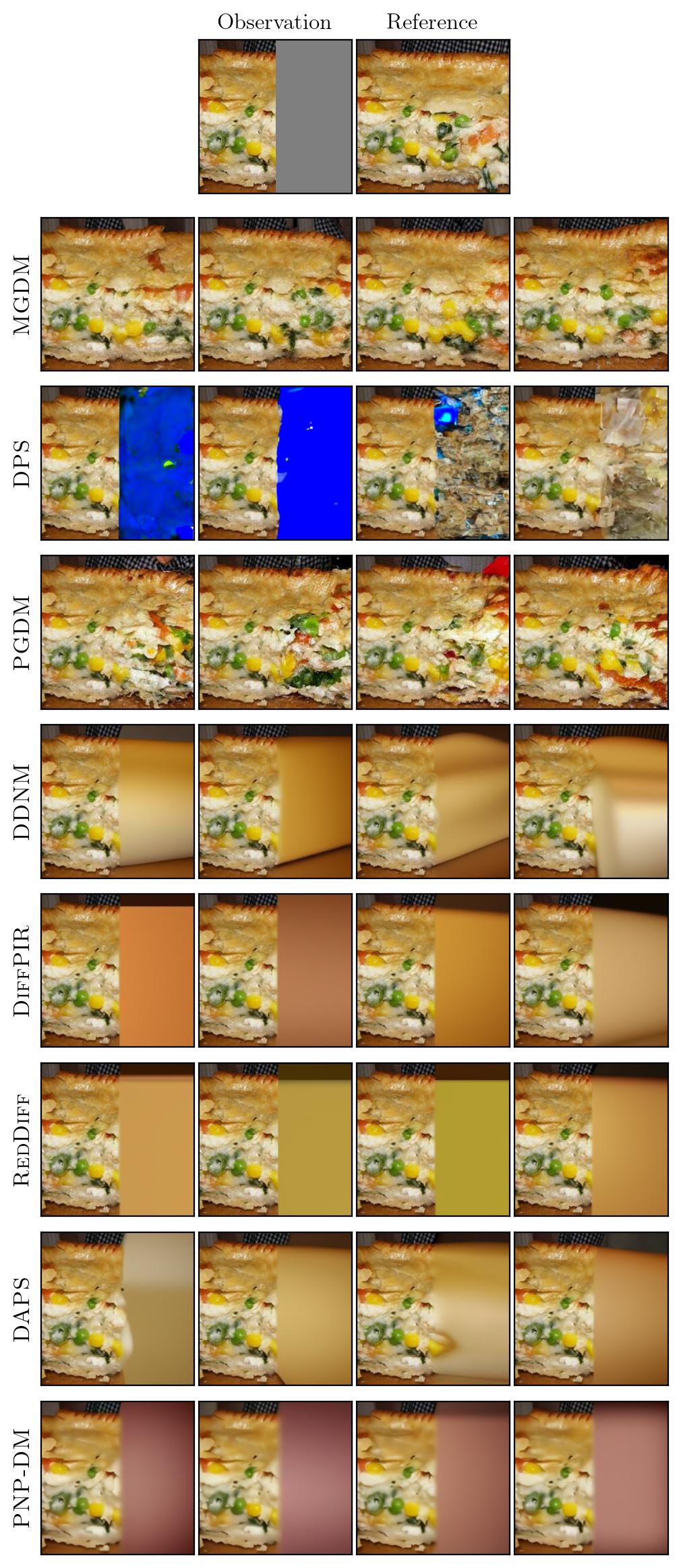}
    }
    \caption{Reconstructions Half mask inpainting on \imagenet\ dataset.}
\end{figure}

\begin{figure}[tb]
    \centering
    \subfigure{
        \includegraphics[width=.49\textwidth]{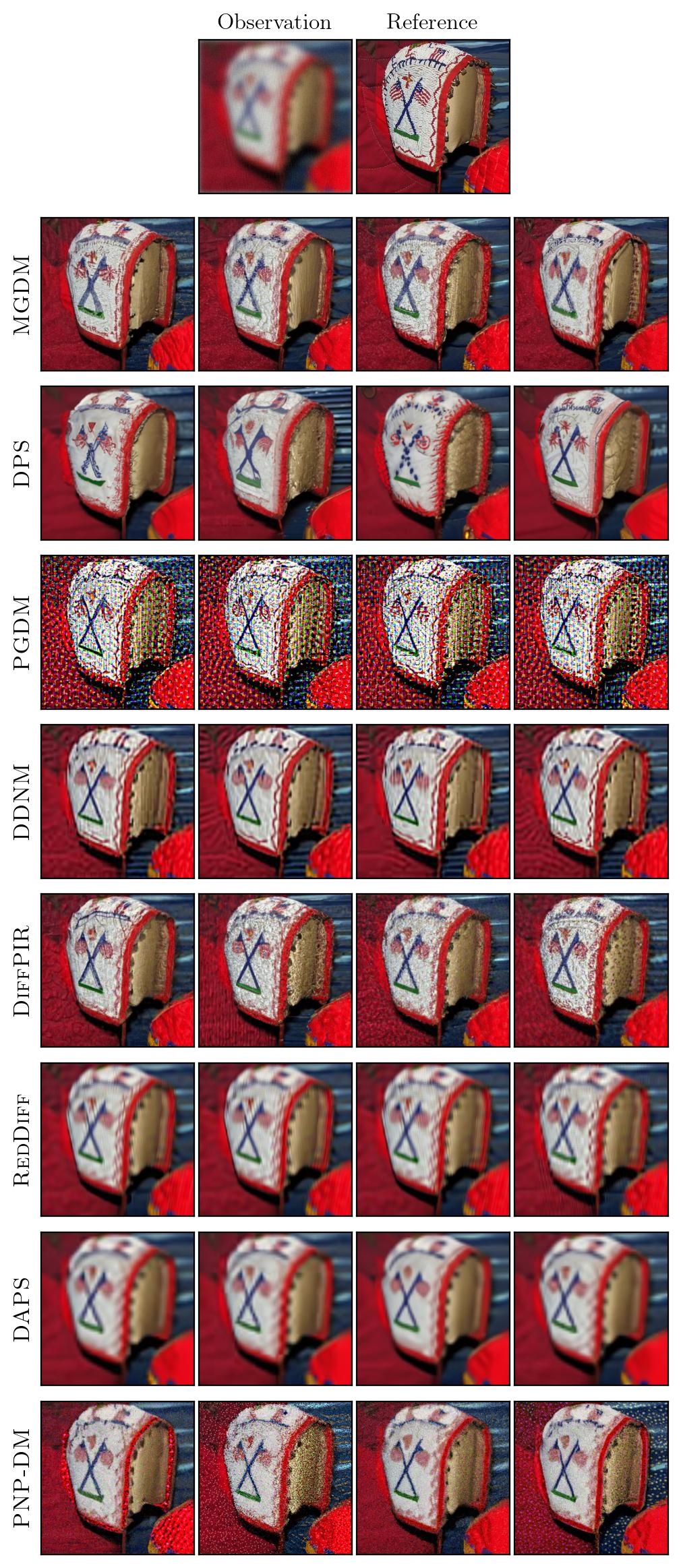}
        \includegraphics[width=.49\textwidth]{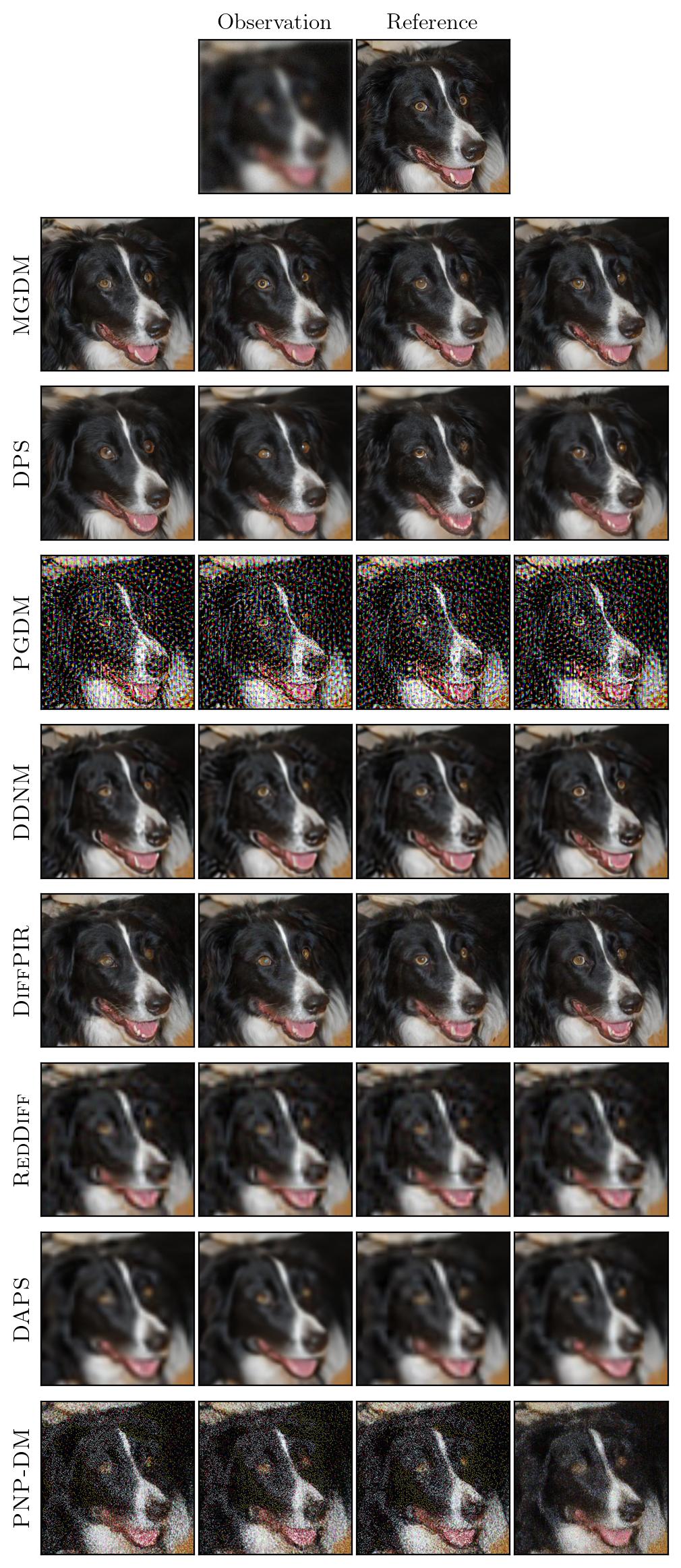}
    }
    \caption{Reconstructions for Gaussian deblurring on \imagenet\ dataset.}
\end{figure}

\begin{figure}[tb]
    \centering
    \subfigure{
        \includegraphics[width=.49\textwidth]{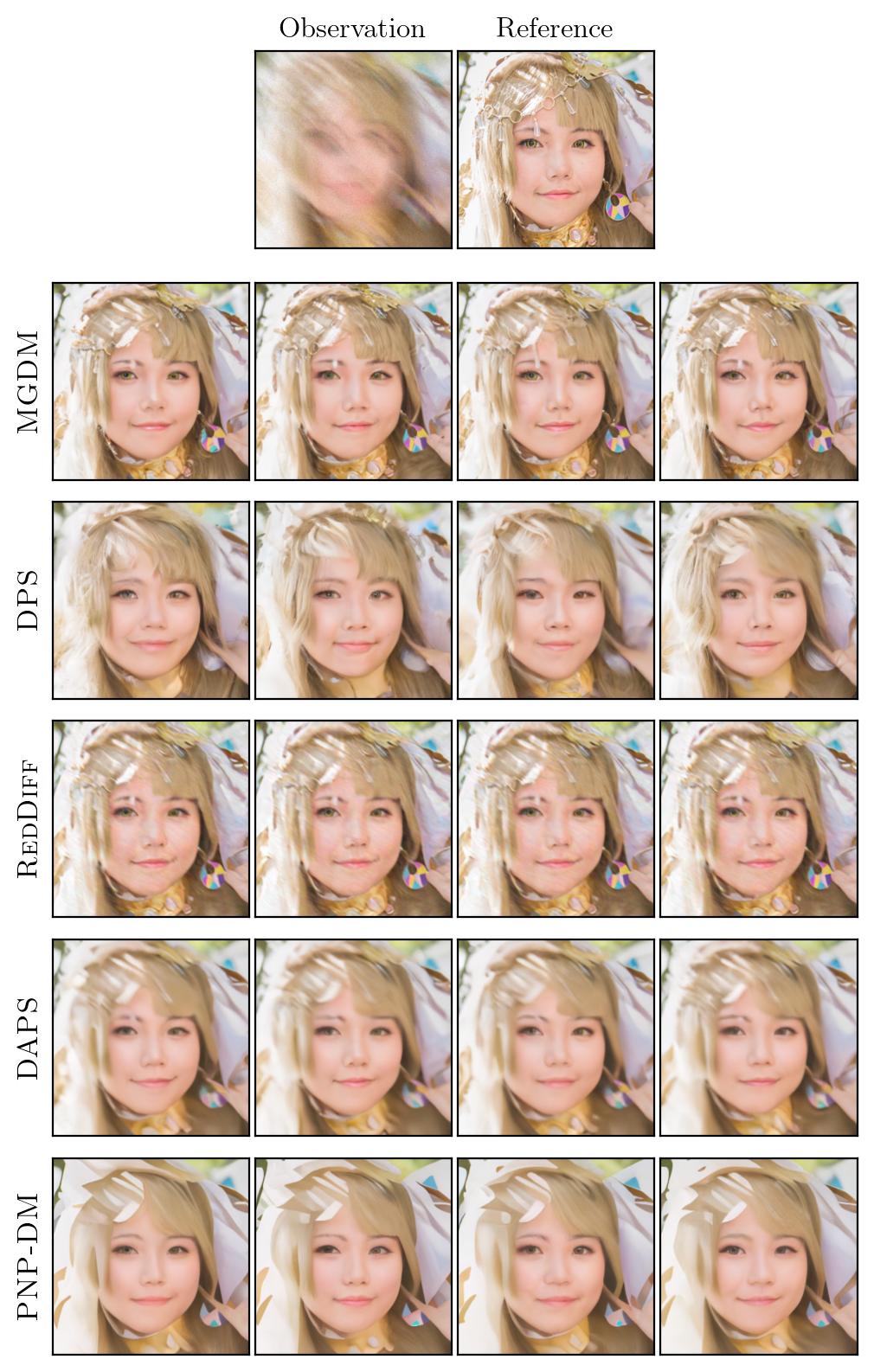}
        \includegraphics[width=.49\textwidth]{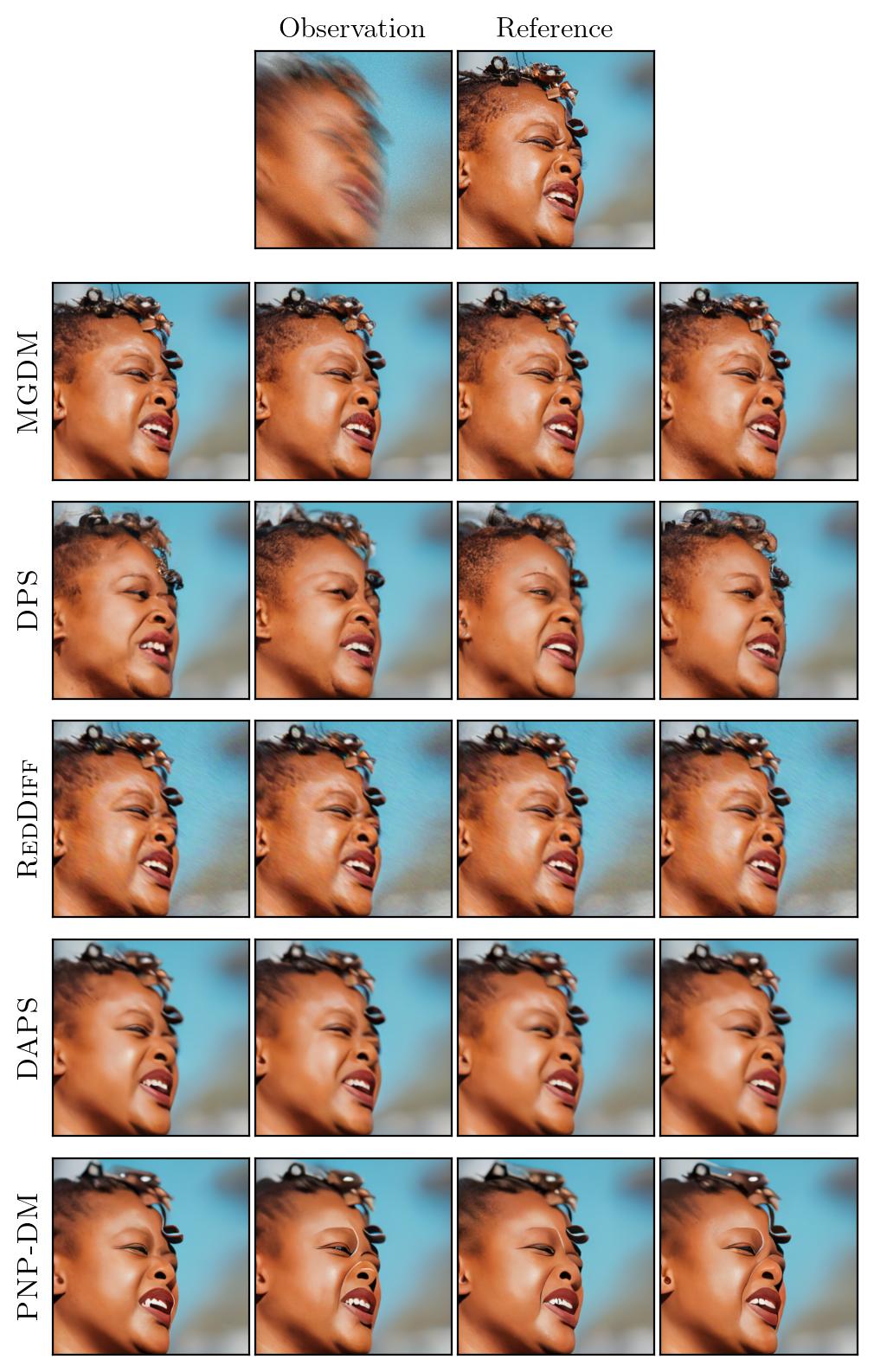}
    }
    \caption{Reconstructions for motion deblurring on \ffhq\ dataset.}
\end{figure}

\begin{figure}[tb]
    \centering
    \subfigure{
        \includegraphics[width=.49\textwidth]{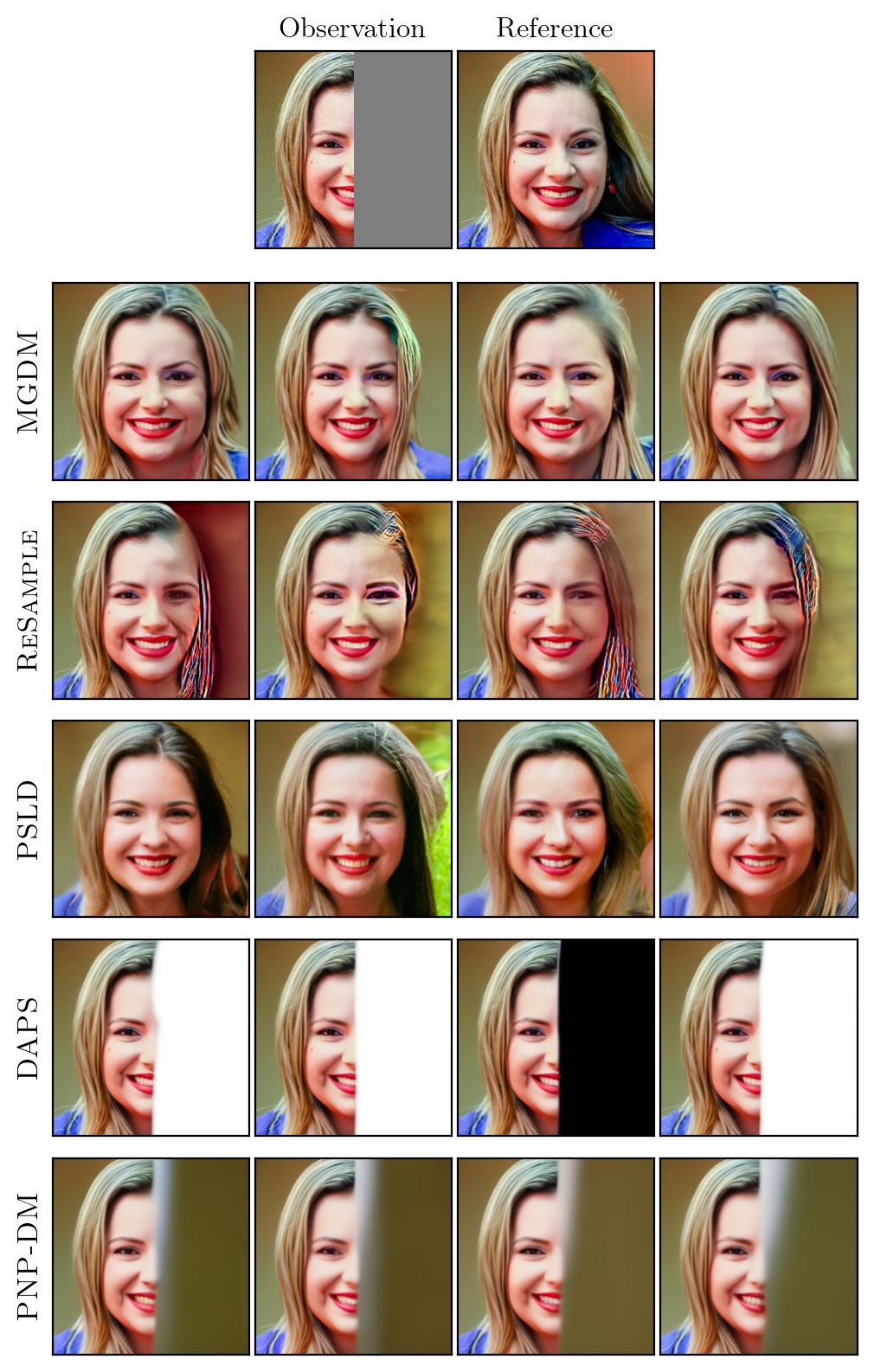}
        \includegraphics[width=.49\textwidth]{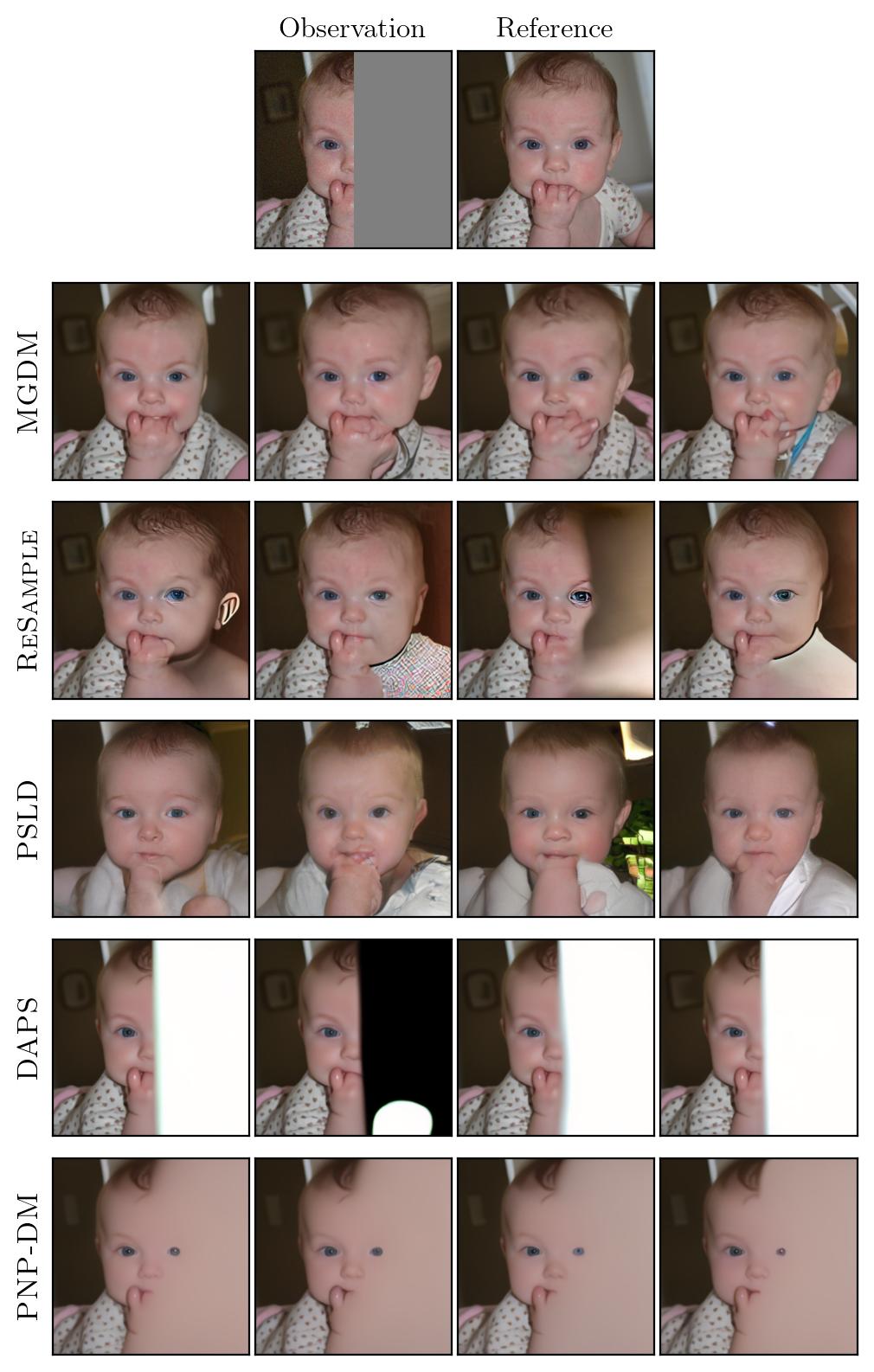}
    }
    \caption{Reconstructions for half mask inpainting on \ffhq\ dataset with LDM prior.}
\end{figure}

\begin{figure}[tb]
    \centering
    \subfigure{
        \includegraphics[width=.49\textwidth]{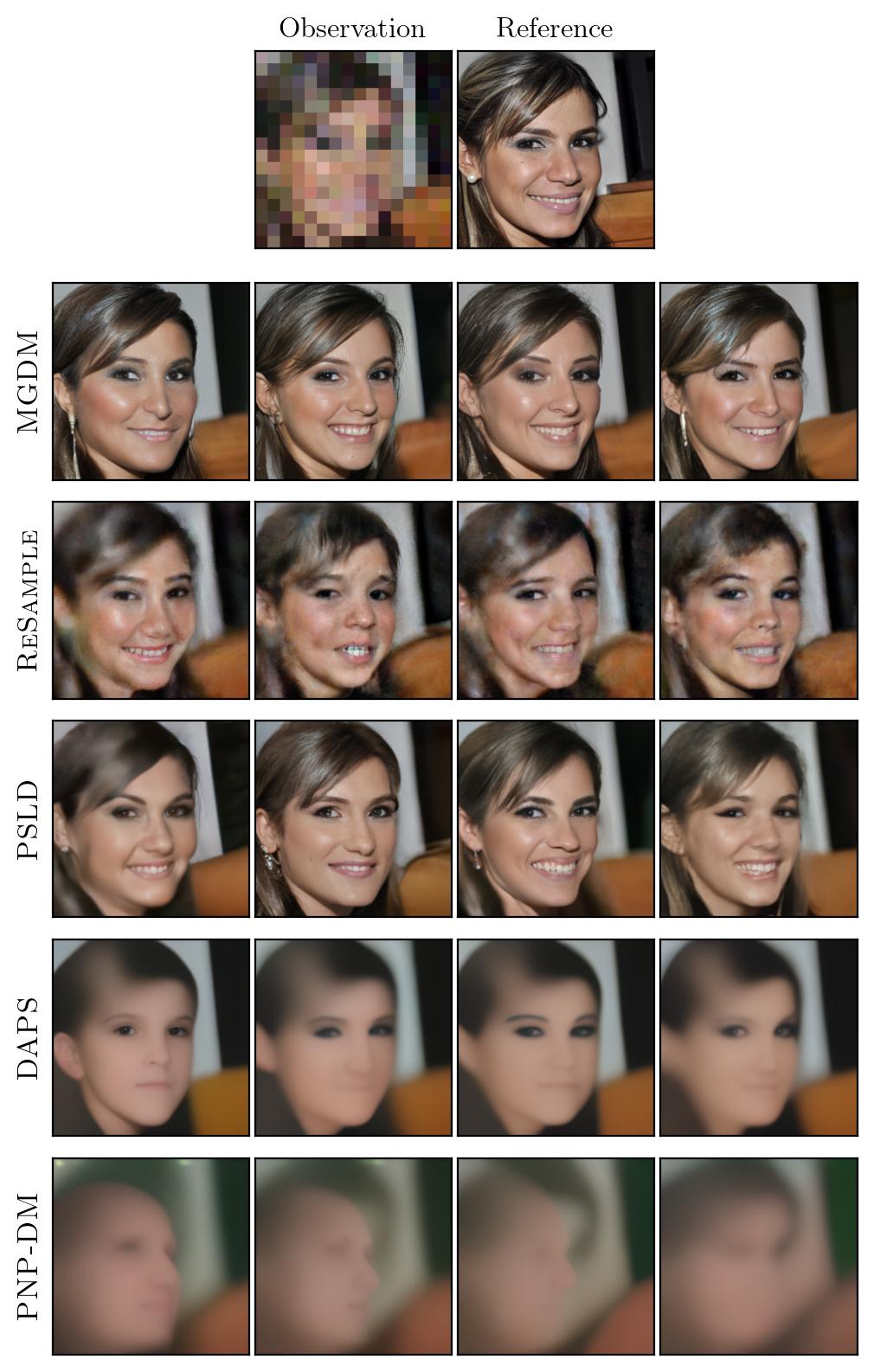}
        \includegraphics[width=.49\textwidth]{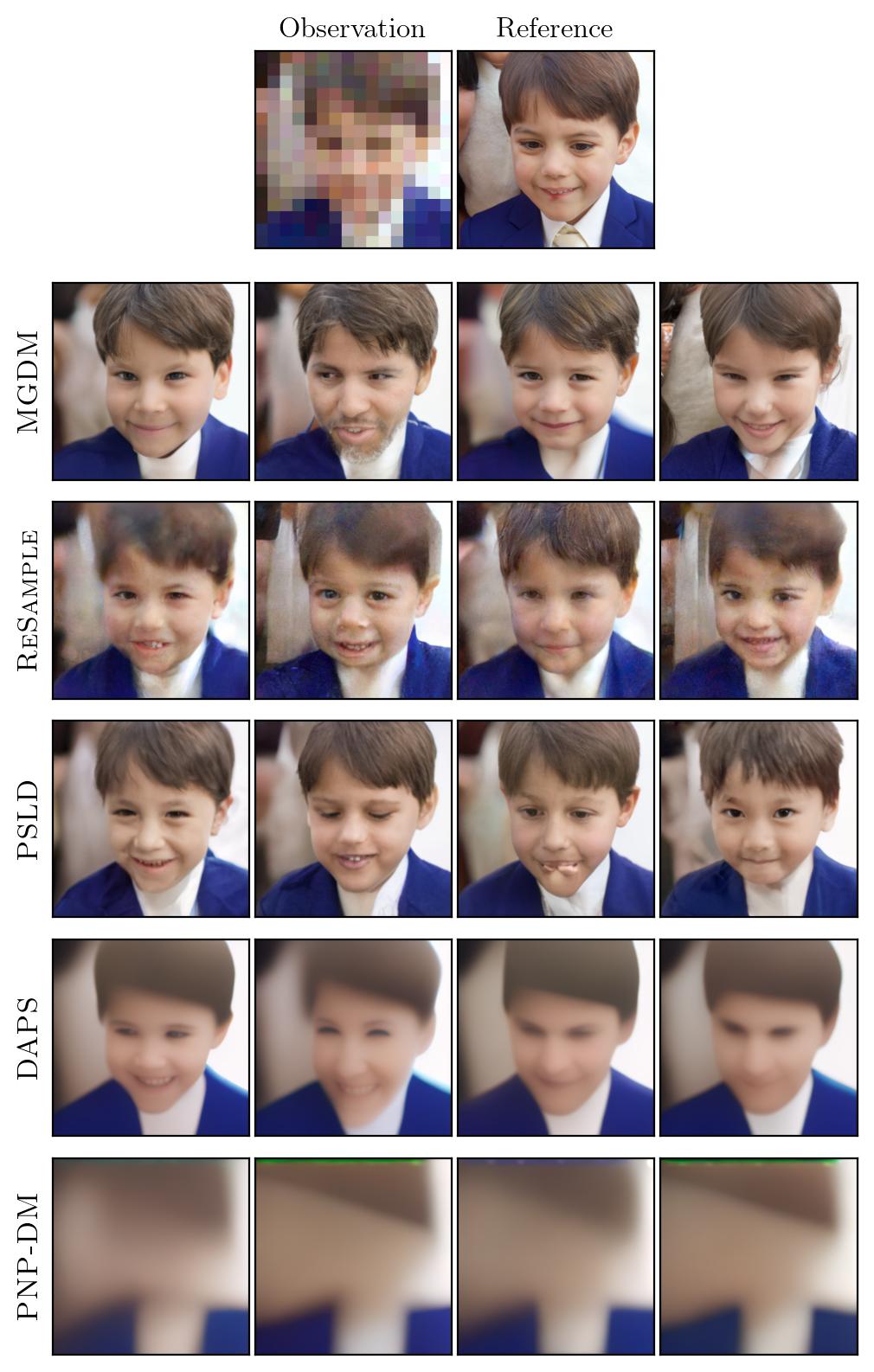}
    }
    \caption{Reconstructions for SR $\times 16$ on \ffhq\ dataset with LDM prior.}
\end{figure}

\begin{figure}
    \centering
    \includegraphics[width=.85\textwidth]{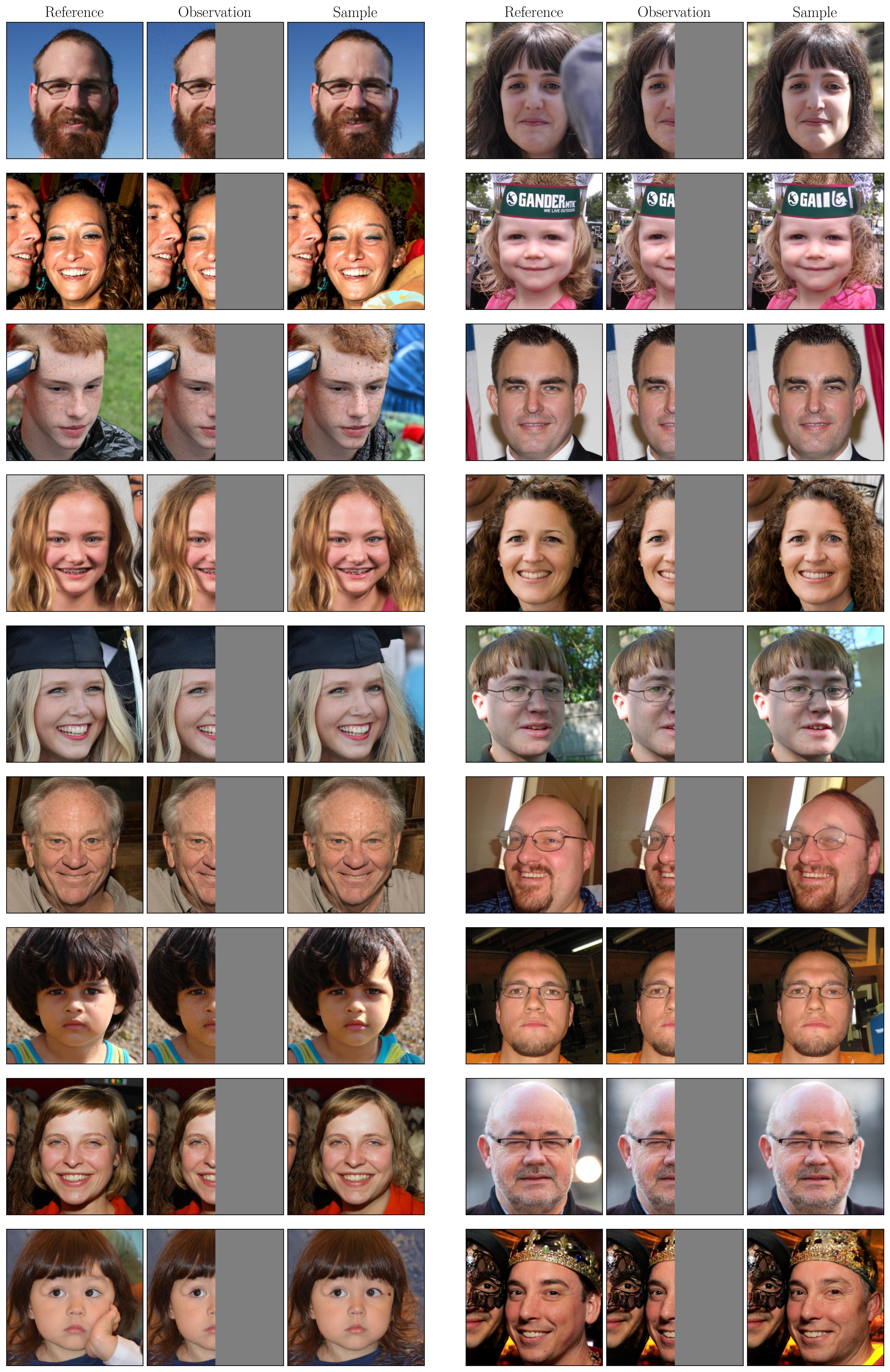}
    \caption{Half mask inpainting on \ffhq\ dataset.}
\end{figure}
\begin{figure}
    \centering
    \includegraphics[width=.85\textwidth]{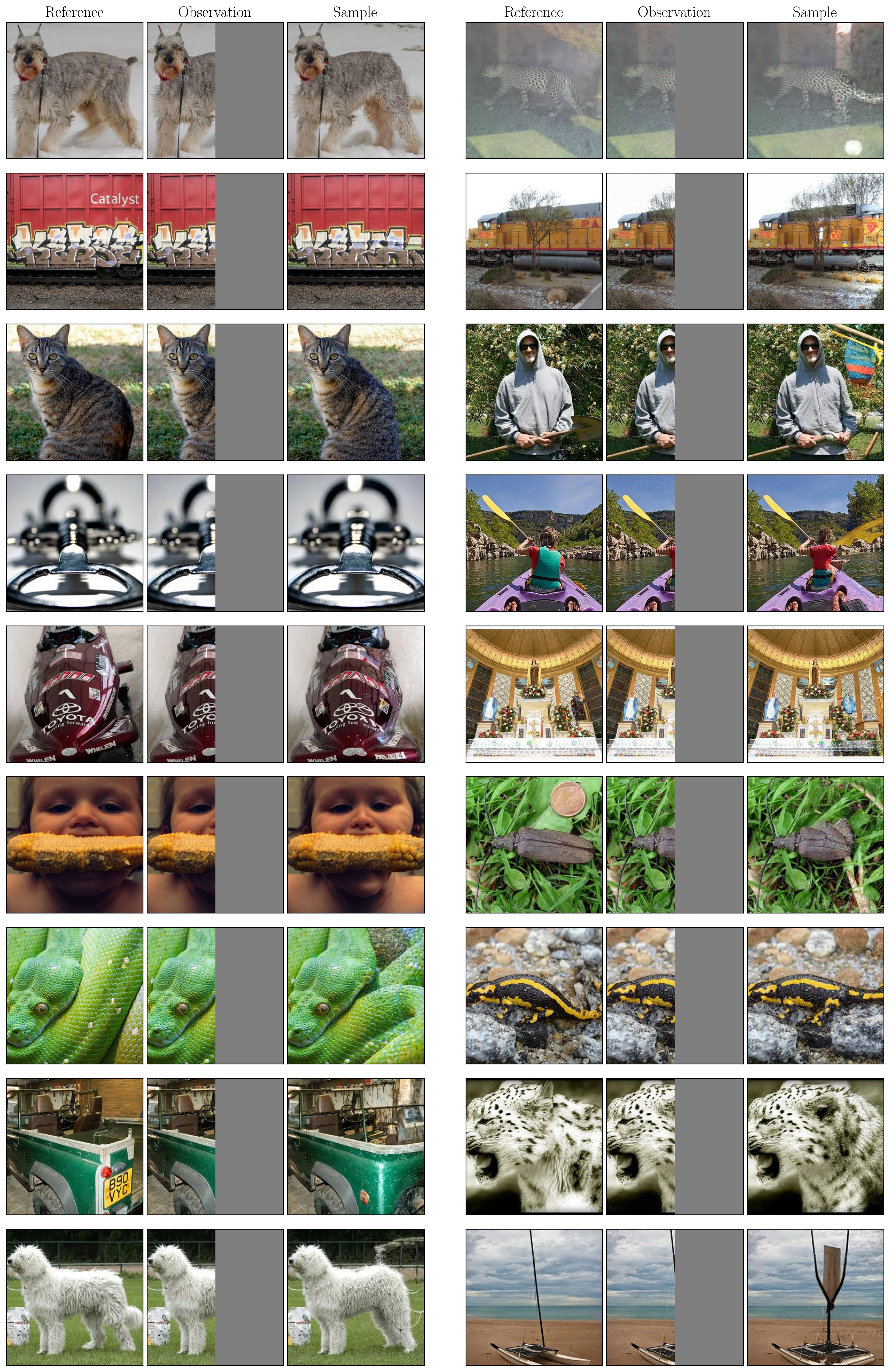}
    \caption{Half mask inpainting on \imagenet\ dataset.}
\end{figure}
\begin{figure}
    \centering
    \includegraphics[width=.85\textwidth]{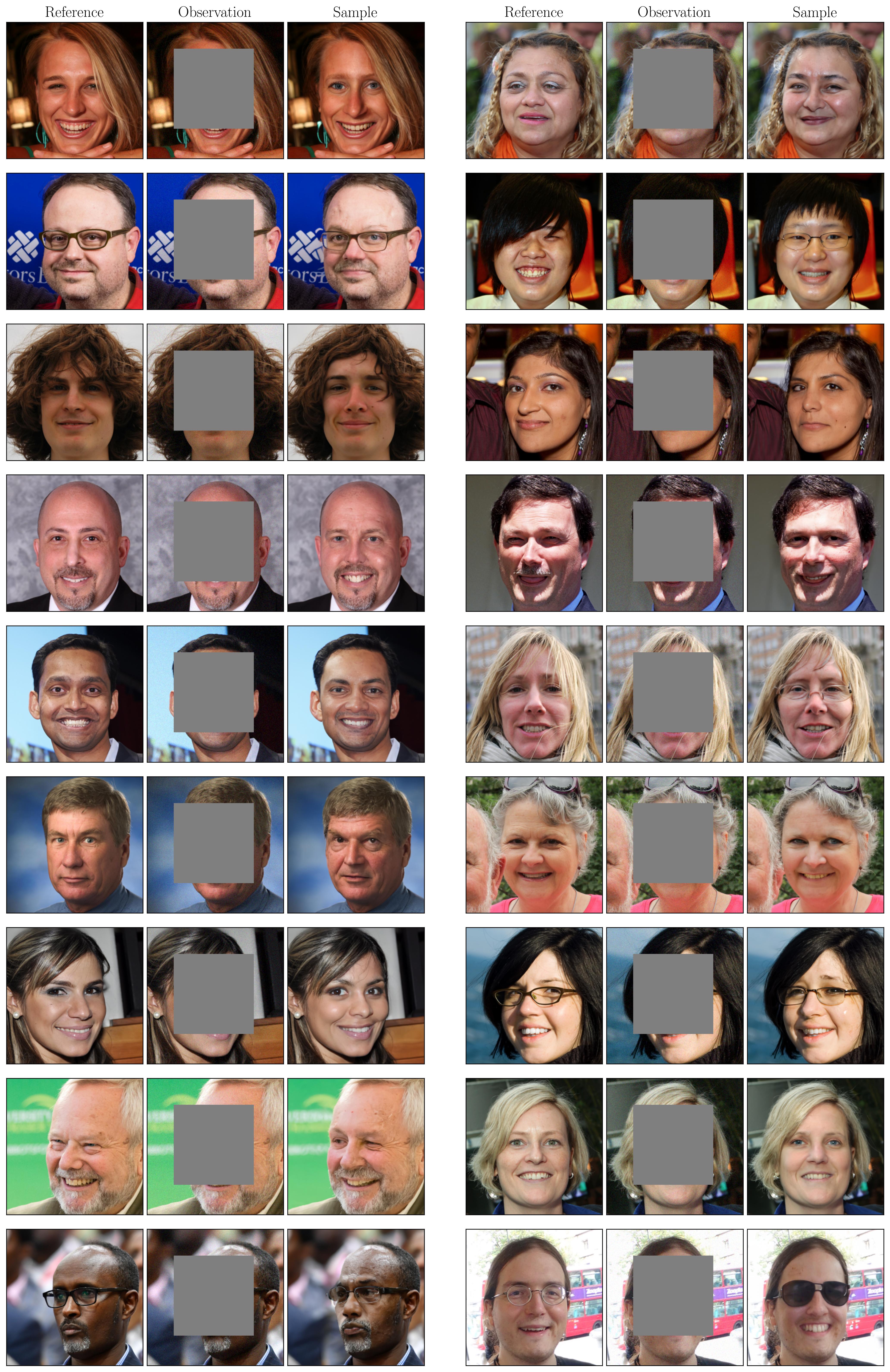}
    \caption{Box inpainting on \ffhq\ dataset.}
\end{figure}
\begin{figure}
    \centering
    \includegraphics[width=.85\textwidth]{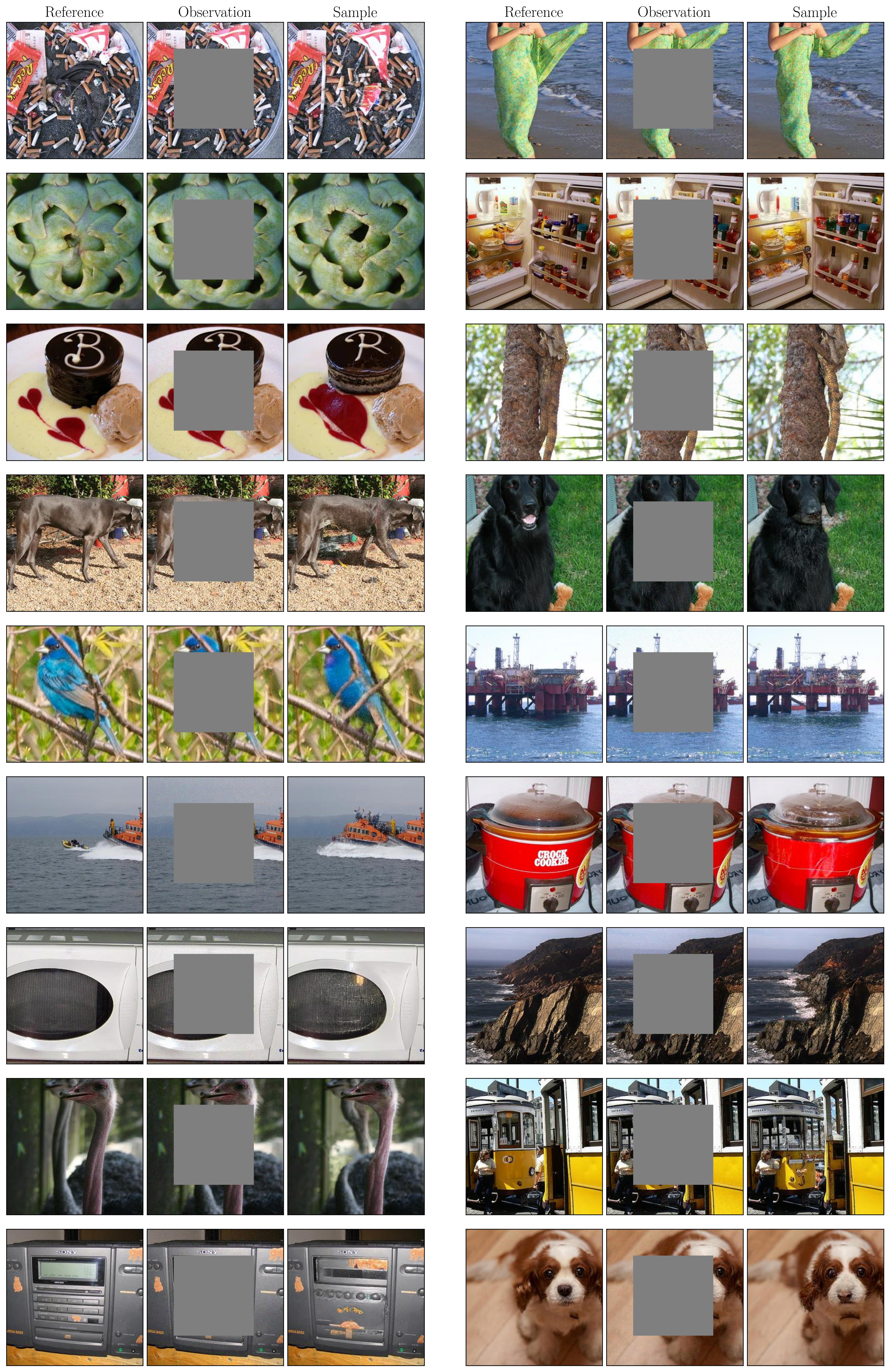}
    \caption{Box inpainting on \imagenet\ dataset.}
\end{figure}
\begin{figure}
    \centering
    \includegraphics[width=.85\textwidth]{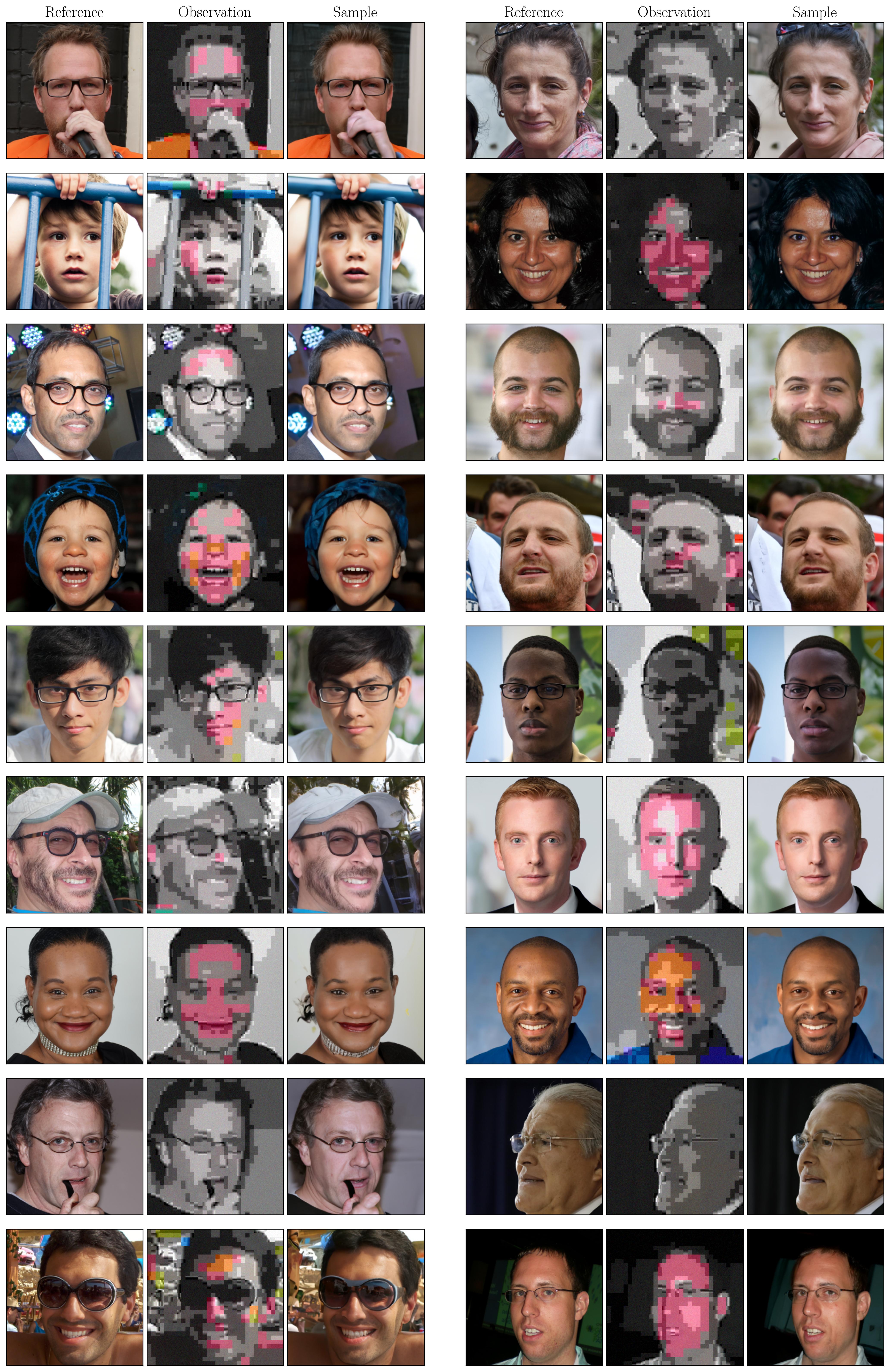}
    \caption{JPEG dequantization with $\mathrm{QF}=2$ on \ffhq\ dataset.}
\end{figure}
\begin{figure}
    \centering
    \includegraphics[width=.85\textwidth]{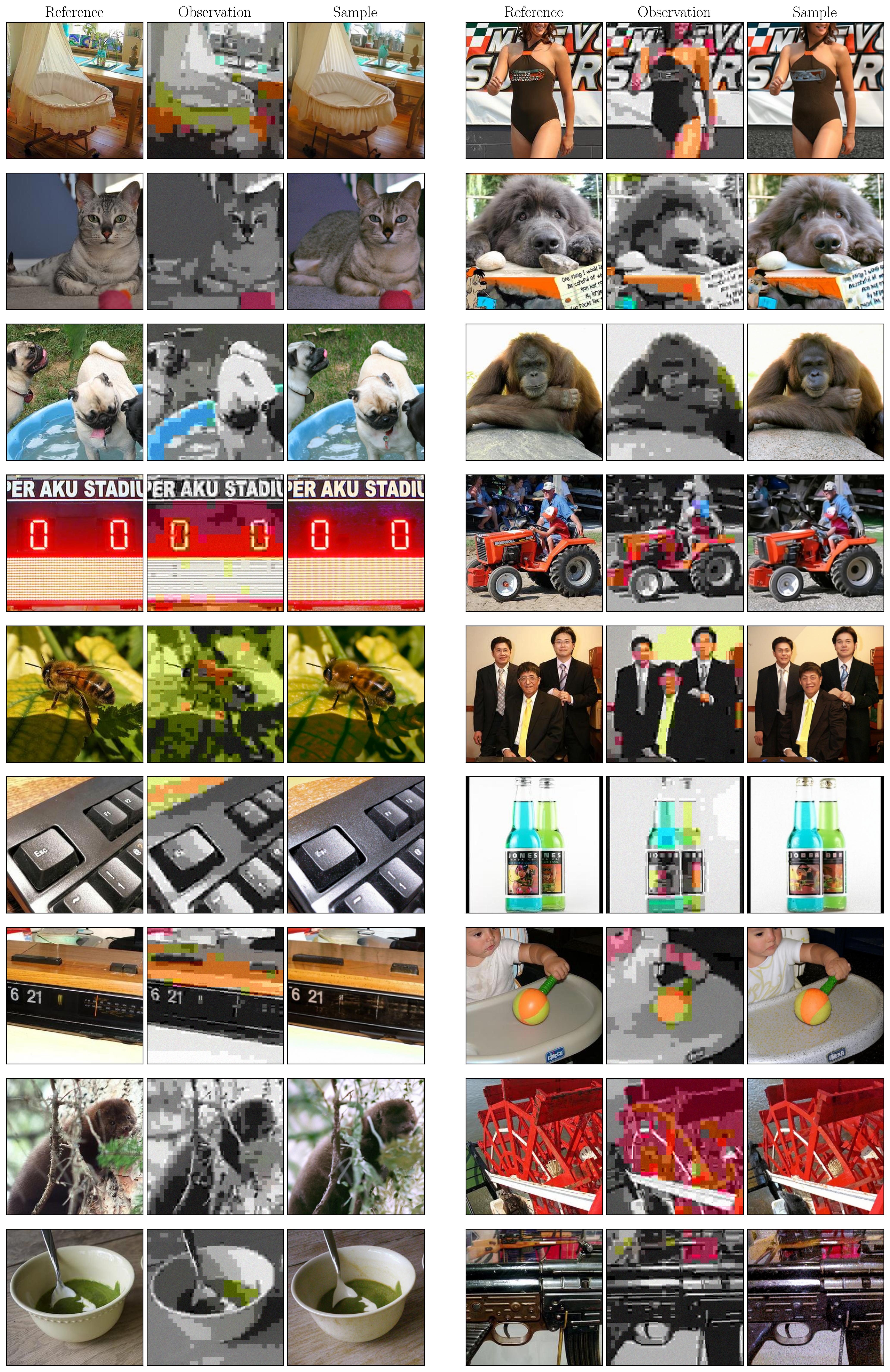}
    \caption{JPEG dequantization with $\mathrm{QF}=2$ on \imagenet\ dataset.}
\end{figure}
\begin{figure}
    \centering
    \includegraphics[width=.85\textwidth]{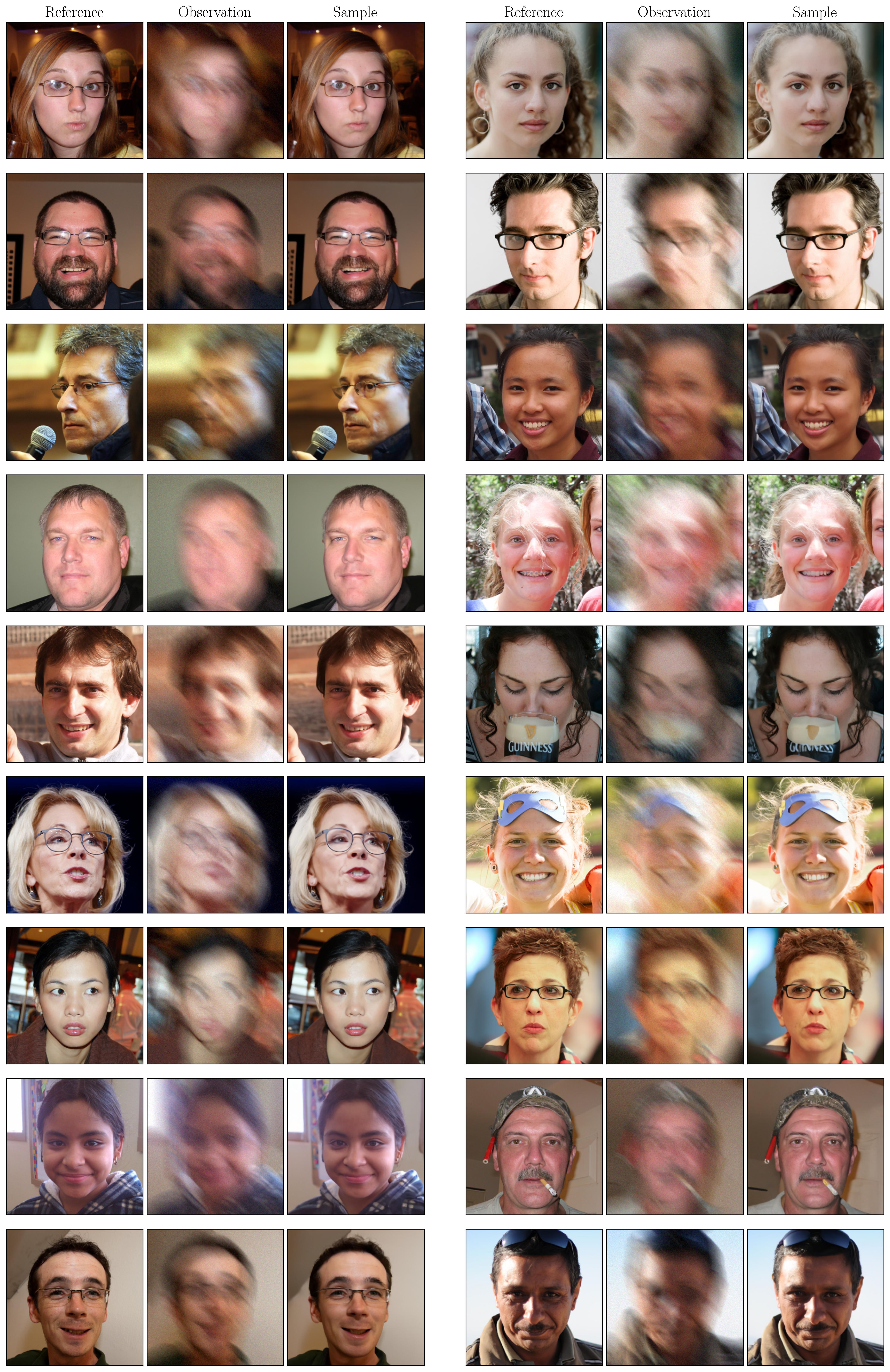}
    \caption{Motion deblurring on \ffhq\ dataset.}
\end{figure}
\begin{figure}
    \centering
    \includegraphics[width=.85\textwidth]{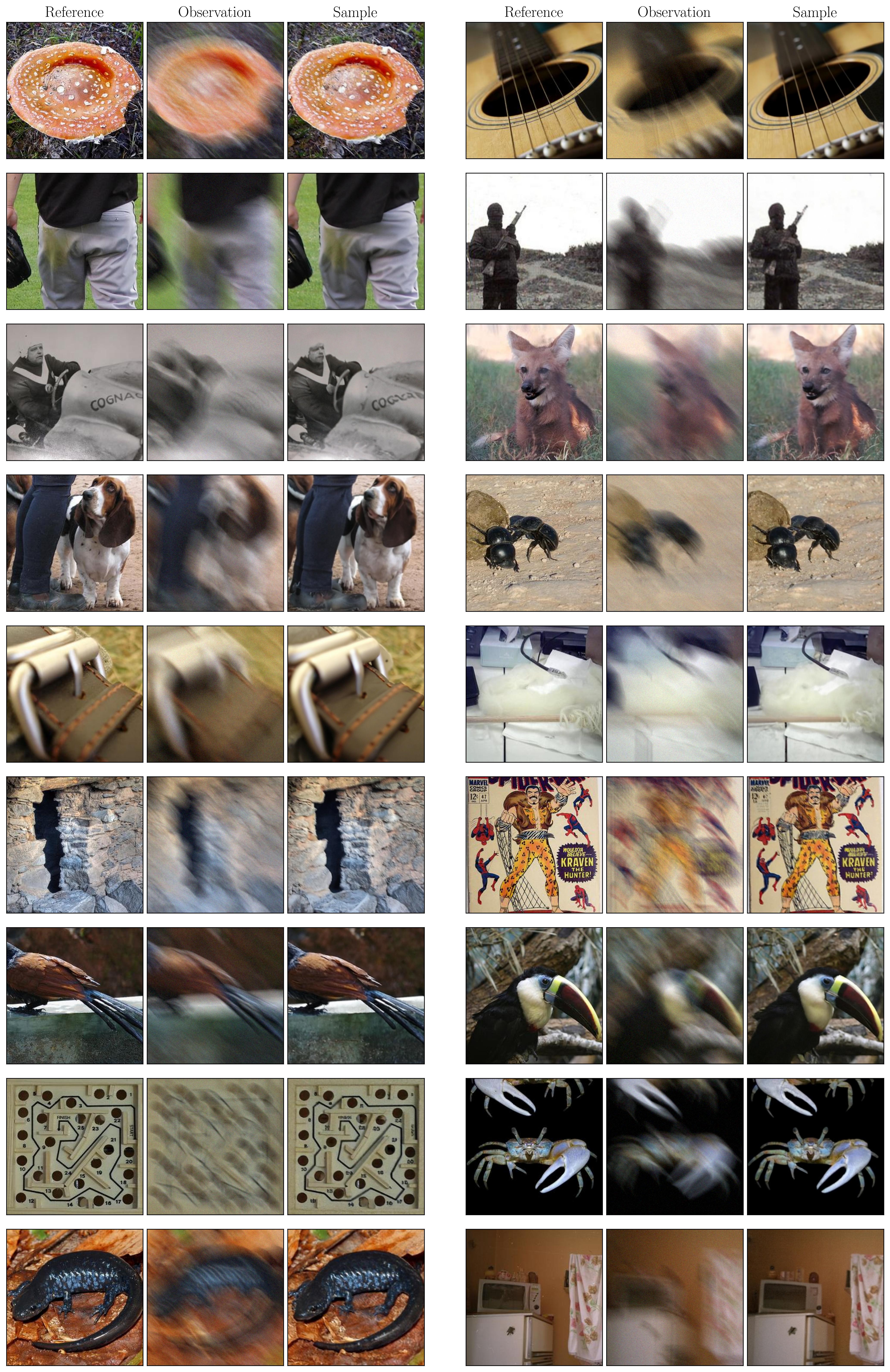}
    \caption{Motion deblurring on \imagenet\ dataset.}
\end{figure}
\begin{figure}
    \centering
    \includegraphics[width=.85\textwidth]{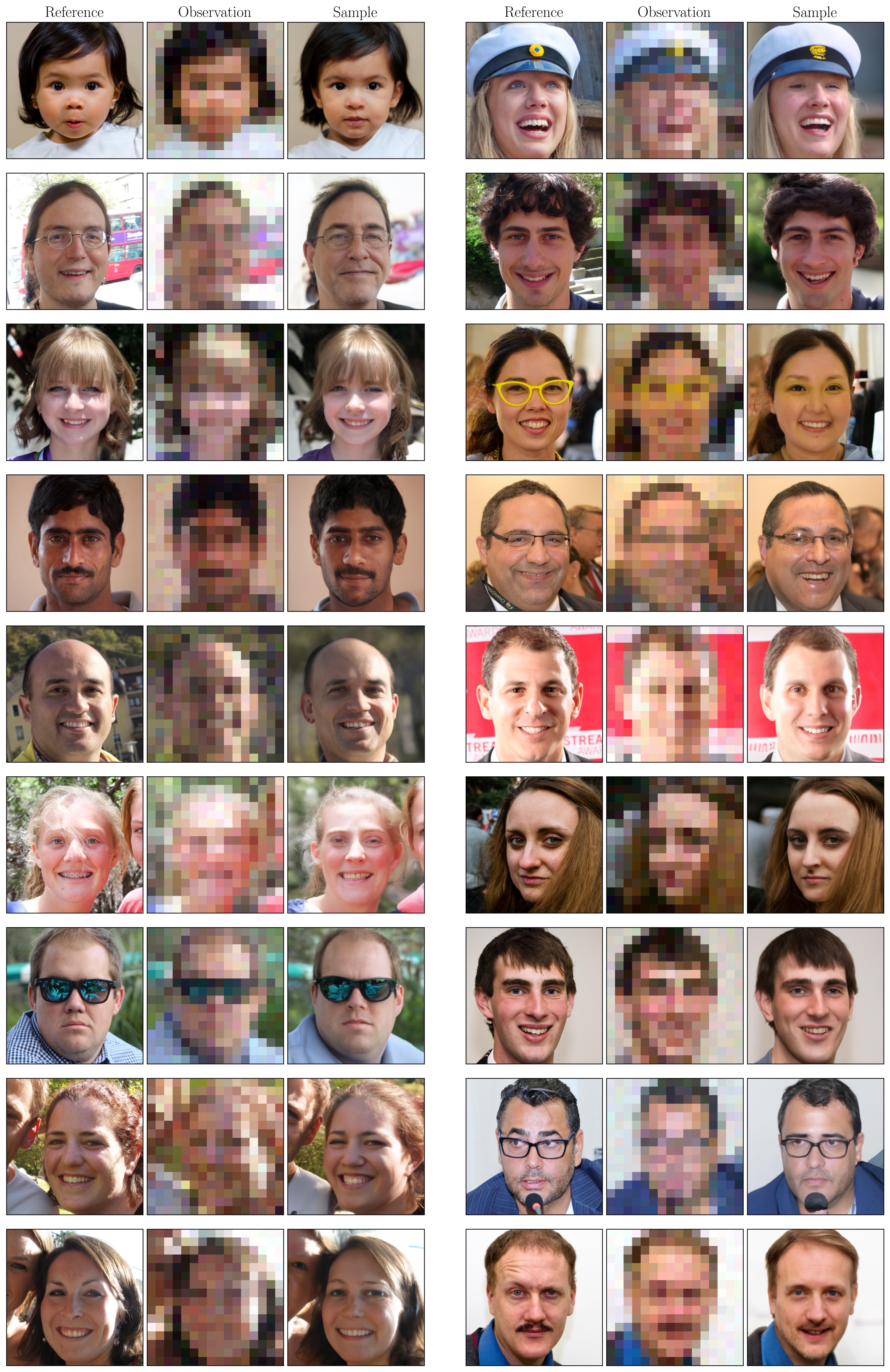}
    \caption{SR(16$\times$) on \ffhq\ dataset.}
\end{figure}
\begin{figure}
    \centering
    \includegraphics[width=.85\textwidth]{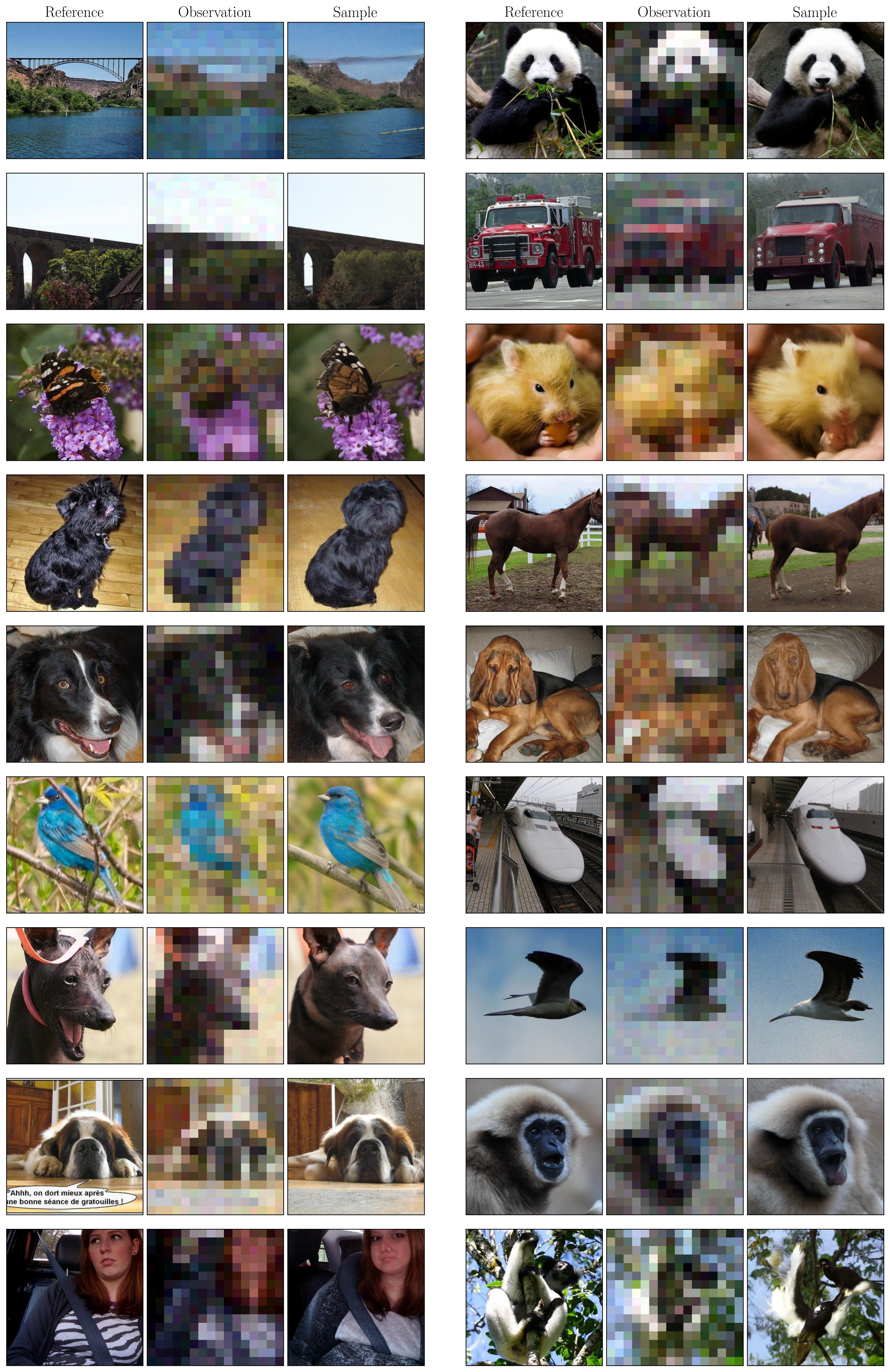}
    \caption{SR(16$\times$) on \imagenet\ dataset.}
\end{figure}
\begin{figure}
    \centering
    \includegraphics[width=.85\textwidth]{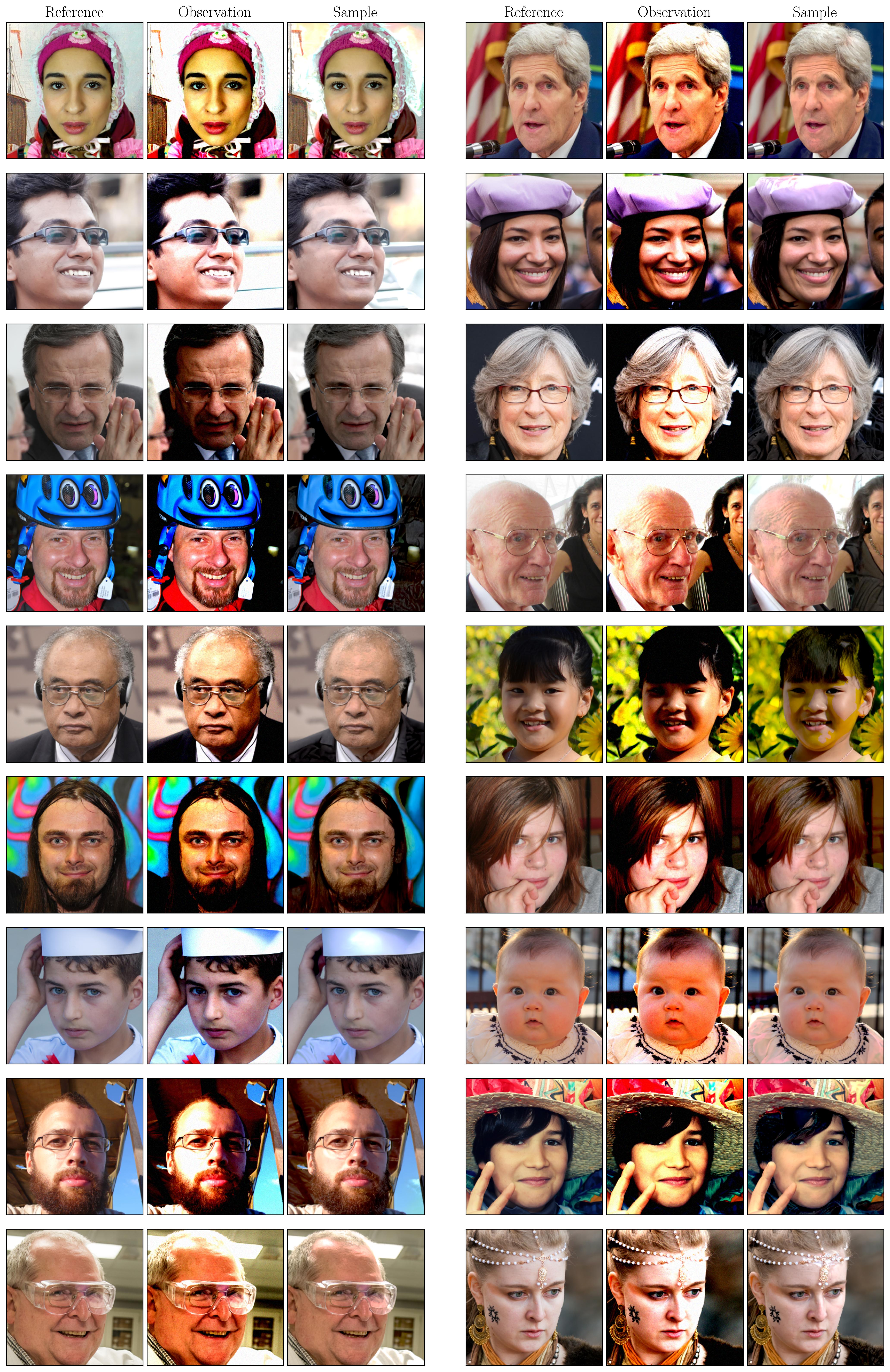}
    \caption{High dynamic range on \ffhq\ dataset.}
\end{figure}
\begin{figure}
    \centering
    \includegraphics[width=.85\textwidth]{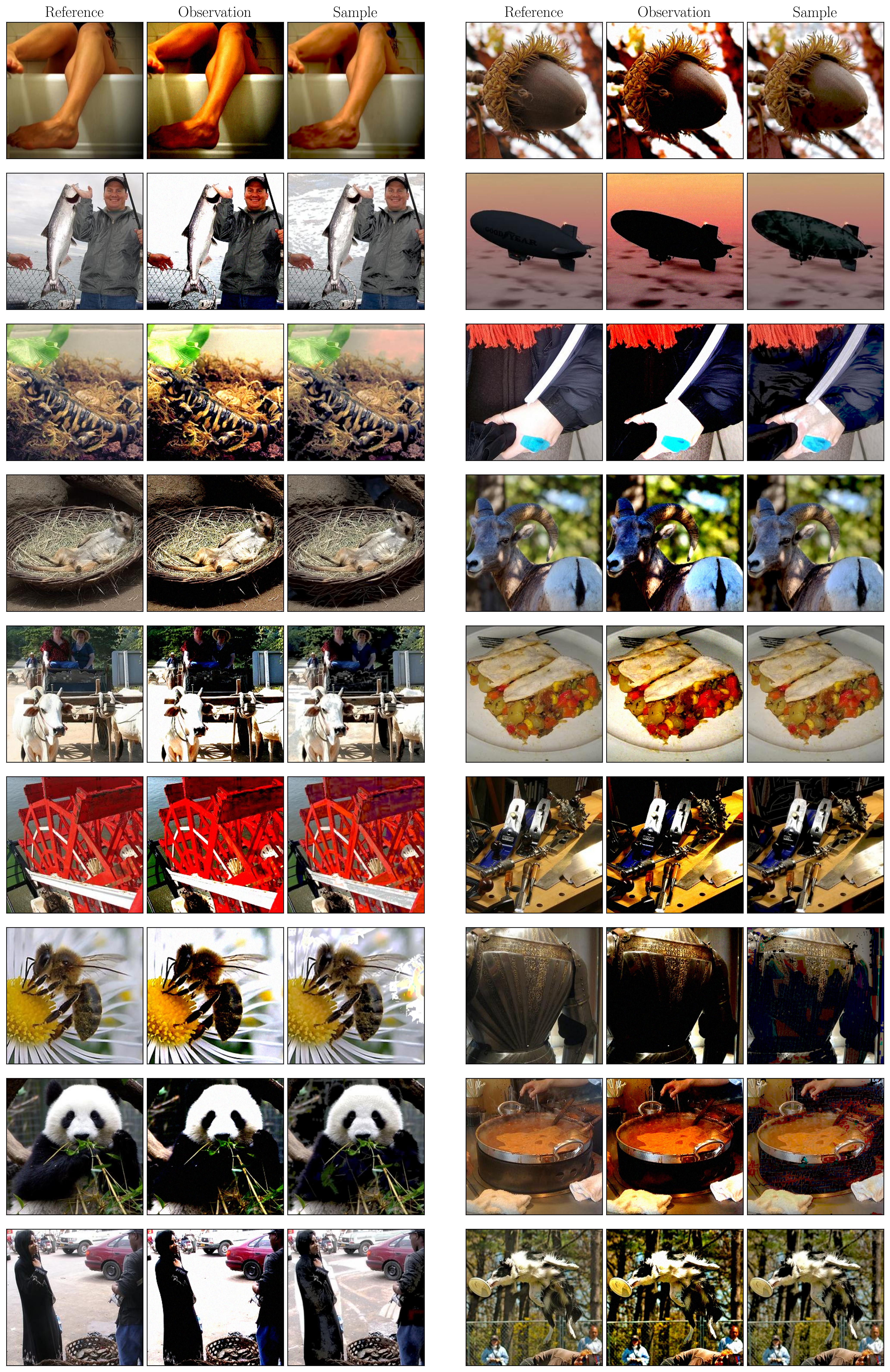}
    \caption{High dynamic range on \imagenet\ dataset.}
\end{figure}
\begin{figure}
    \centering
    \includegraphics[width=.85\textwidth]{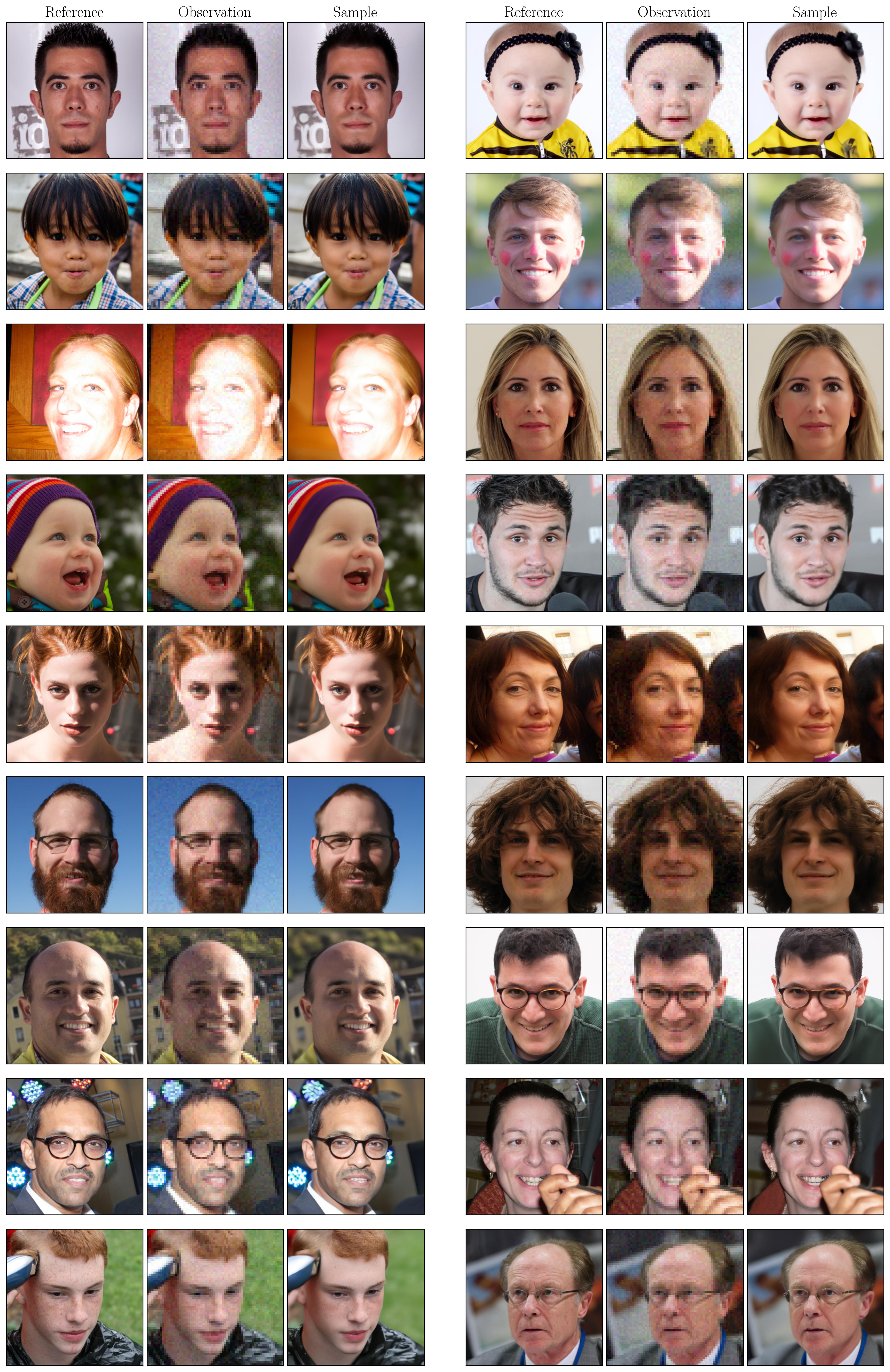}
    \caption{SR(4$\times$) on \ffhq\ dataset with latent diffusion.}
\end{figure}
\begin{figure}
    \centering
    \includegraphics[width=.85\textwidth]{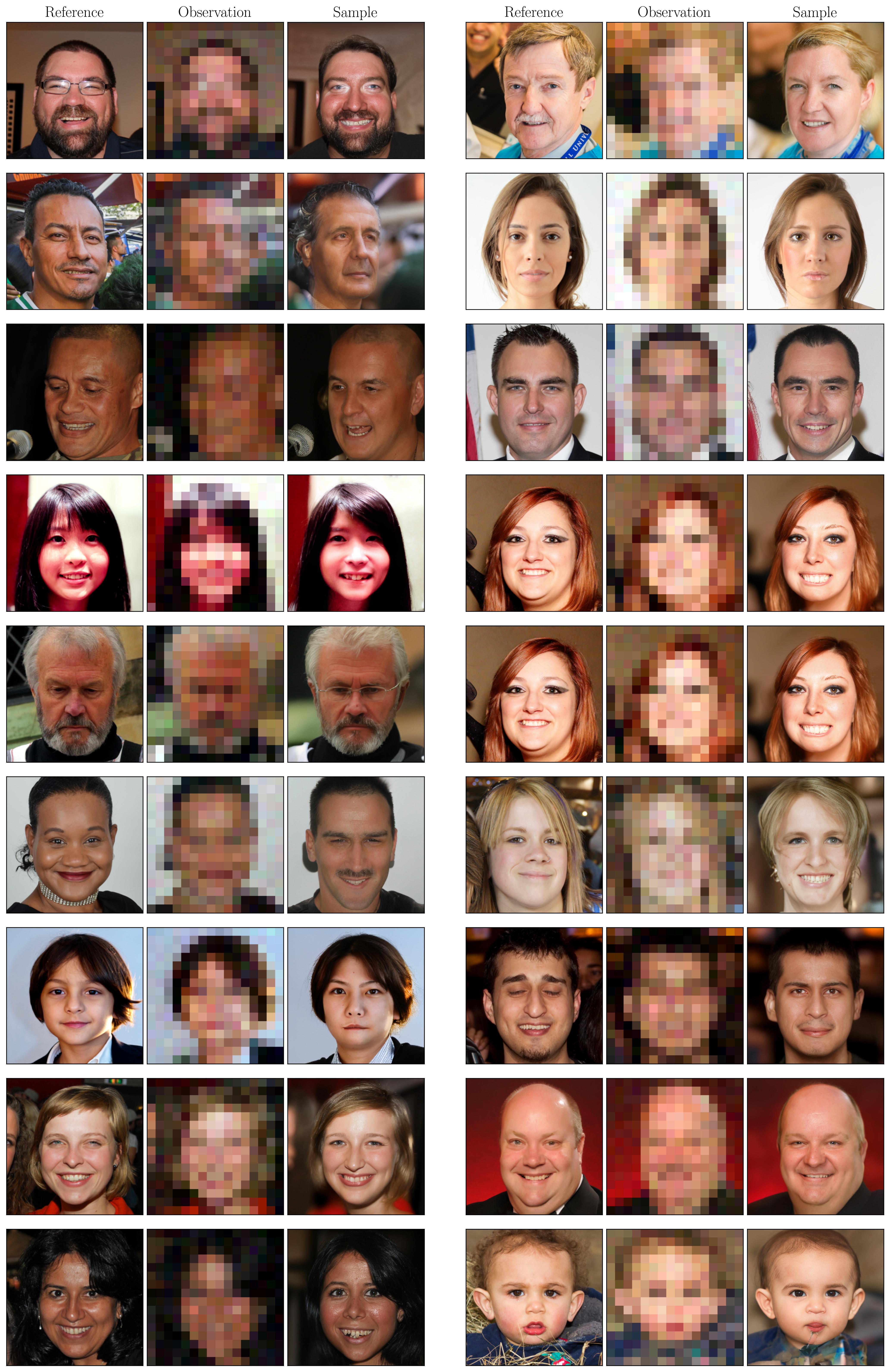}
    \caption{SR(16$\times$) on \ffhq\ dataset with latent diffusion.}
\end{figure}
\begin{figure}
    \centering
    \includegraphics[width=.85\textwidth]{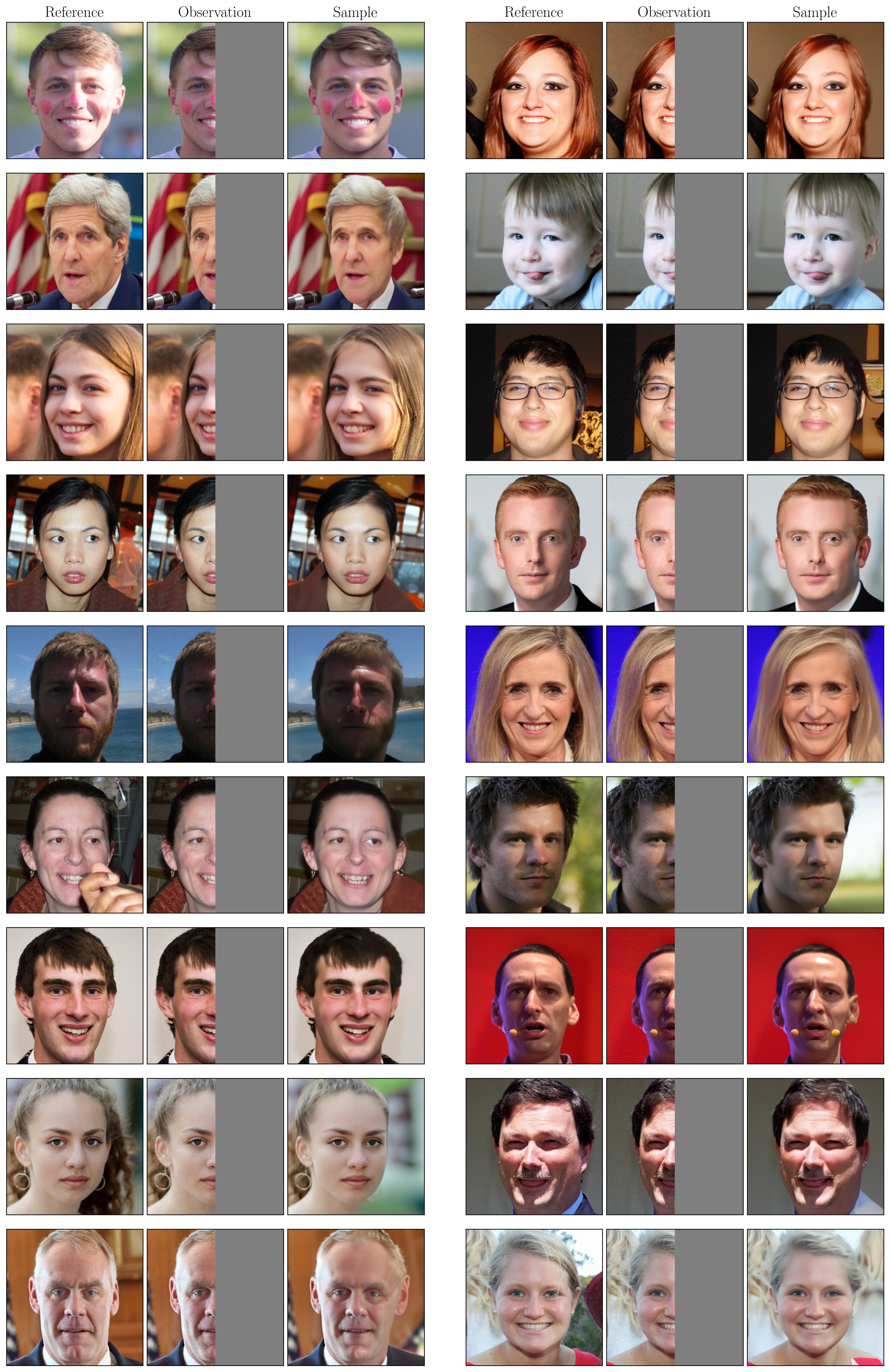}
    \caption{Half mask on \ffhq\ dataset with latent diffusion.}
\end{figure}

\label{apdx-sec:visual-reconstructions}

\end{document}